\documentclass{article}

\usepackage[preprint,nonatbib]{neurips_2025}

\newcommand{\R}{\mathbb{R}}
\newcommand{\eS}{\mathbb{S}}

\usepackage[utf8]{inputenc} 
\usepackage[T1]{fontenc}    
\usepackage{url}            
\usepackage{booktabs}       
\usepackage{amsfonts}       
\usepackage{nicefrac}       
\usepackage{microtype}      

\usepackage{wrapfig}
\usepackage{amsthm,amssymb,amsmath,hyperref,xcolor, float, graphicx,tcolorbox, tikz, tikz-cd, algorithm, algpseudocode,overpic, nicematrix}
\usepackage{enumitem}
\usepackage{verbatim}
\usepackage{tikz}
\usepackage{extarrows}
\usetikzlibrary{angles,quotes}
\DeclareMathAlphabet{\mathrnd}{U}{dutchcal}{m}{n}
\SetMathAlphabet{\mathrnd}{bold}{U}{dutchcal}{b}{n}

\hypersetup{
    colorlinks,
    linkcolor={blue},
    citecolor={blue},
    urlcolor={blue},
    filecolor = {violet}
}

\DeclareMathOperator{\sign}{sign}

\DeclareMathOperator{\Cov}{Cov}

\renewcommand{\epsilon}{\varepsilon}

\newtheorem{lemma}{Lemma}[section]
\newtheorem{proposition}{Proposition}[section]

\newtheorem{theorem}{Theorem}[section]
\newtheorem{corollary}{Corollary}[section]

\newtheorem*{corollary*}{Corollary}

\newtheorem{definition}{Definition}[section]
\theoremstyle{remark}

\newcommand{\Dh}{\mathcal{D}_{H}}
\newcommand{\Dg}{\mathcal{D}_{G}}

\usepackage{titletoc}
\usepackage{multibib}
\newcites{app}{Appendix References}

\newcommand{\EE}{\mathbb{E}}
\newcommand{\PP}{\mathbb{P}}
\newcommand{\var}{\operatorname{var}}

\usepackage{xr-hyper}
\usepackage{hyperref}

\title{G-Net: A Provably Easy Construction of High-Accuracy Random Binary Neural Networks}

\author{%
  Alireza Aghasi\thanks{Corresponding Author} \\
  Dept. Electrical Engineering and Computer Science\\
  Oregon State University\\
  \texttt{alireza.aghasi@oregonstate.edu} \\
  \And
  Nicholas F. Marshall\\
  Dept. Mathematics\\
  Oregon State University\\
  \texttt{marsnich@oregonstate.edu} \\
  \And
    Saeid Pourmand \\
  Dept. Electrical Engineering and Computer Science\\
  Oregon State University\\
  \texttt{pourmans@oregonstate.edu} \\
  \And
  Wyatt D Whiting\\
  Dept. Mathematics\\
  Oregon State University\\
  \texttt{whitinwy@oregonstate.edu} \\
}

\begin{document}

\maketitle

\begin{abstract}
  We propose a novel randomized algorithm for constructing binary neural networks with tunable accuracy. This approach is motivated by hyperdimensional computing (HDC), which is a brain-inspired paradigm that leverages high-dimensional vector representations, offering efficient hardware implementation and robustness to model corruptions. Unlike traditional low-precision methods that use quantization, we consider binary embeddings of data as points in the hypercube equipped with the Hamming distance. We propose a novel family of floating-point neural networks, G-Nets, which are general enough to mimic standard network layers. Each floating-point G-Net has a randomized binary embedding, an embedded hyperdimensional (EHD) G-Net, that retains the accuracy of its floating-point counterparts, with theoretical guarantees, due to the concentration of measure. Empirically, our binary models match convolutional neural network accuracies and outperform prior HDC models by large margins, for example, we achieve almost 30\% higher accuracy on CIFAR-10 compared to prior HDC models. G-Nets are a theoretically justified bridge between neural networks and randomized binary neural networks, opening a new direction for constructing robust binary/quantized deep learning models. Our implementation is available at \url{https://github.com/GNet2025/GNet}.
\end{abstract}

\section{Introduction}
We are motivated by hyperdimensional computing (HDC), a brain-inspired computational paradigm that uses high-dimensional vectors as the basic elements of computation 
\cite{reviewpaper}. Applications of HDC include
classification and regression 
\cite{ Duan2022, onlineHD2021,
Geethan2021, Wang2023, Zou2021},
reinforcement learning 
\cite{Chen2022,Ni2022}, and dimensionality reduction
\cite{Imani2019b, Morris2019}. 
Recent works have also explored online learning approaches \cite{onlineHD2021, Imani2019a} and advanced the theoretical understanding of HDC \cite{LaplaceHDC, Thomas2021, yu2022understanding}.
HDC methods can be efficiently implemented on hardware, 
making these models attractive for edge and low-energy computing \cite{basklar2021,Chuang2020,Imani2019,Karunaratne2020,Khalegi2021,kleyko2023survey}. For readers less familiar with HDC, Section~\ref{sec:HDCTax} of the Appendix offers a detailed taxonomy and broader overview of HDC methods.

For predictive tasks, an HDC pipeline typically consists of two core components: a hyperdimensional embedding module, which often carries some component of randomness, and an inference module \cite{verges2025classification}. The embedding transforms input data into high-dimensional representations---often binary hypervectors---that retain key structural features while simplifying the data space. The inference module, commonly a lightweight classifier, operates on these embeddings and often resembles a support vector machine in decision making \cite{kleyko2022survey,kleyko2023survey}. These components are generally decoupled, and due to limited representational capacity, lack of task-specific learning---especially in the embedding stage---and the simplicity of the overall architecture, HDC models struggle to match the performance of state-of-the-art predictors. Moreover, the high-dimensional and highly structured (e.g., binary) nature of the embedding space hinders the training of more sophisticated models in that domain.

In this paper, we introduce a new class of (random) binary neural networks that operate in the hyperdimensional Hamming space without requiring training within that space. Typical approaches to training binary neural networks include using a straight-through estimator (STE)
\cite{courbariaux2016binarized, RastegariMohammad2016XICU}, a continuous piecewise-defined approximation \cite{LiuZechun2018BNEt}, a thermometer encoder \cite{zhang2021fracbnn}, or quantization 
\cite{gholami2022survey,hubara2018quantized} (see Section \ref{sec:BNNTax} of the Appendix for a more detailed overview of binary neural networks).
In contrast, our approach leverages the geometry of the hypercube and random matrix theory. We propose a class of floating-point networks, \emph{G-Nets}, that can be converted into randomized binary networks with accuracy guarantees derived from the concentration of measure. Central to our method is the random binary encoding $\phi:\R^p \to \{-1,1\}^N$, defined as $\phi(x) = \texttt{sign}(\mathrnd{G}x)$, where $\mathrnd{G} \in \R^{N \times p}$ has i.i.d. standard normal entries---a common HDC encoding scheme \cite{rachkovskij2015formation, Imani2019}. For large $N$, Grothendieck's identity (see Lemma~\ref{grothendieck}) ensures that $\frac{1}{N} \left\langle \phi(x),\phi(y)\right\rangle \approx \frac{2}{\pi}\arcsin(\langle x,y\rangle )$ when $x,y\in\eS^{p-1}$.
Thus, inner products in $\eS^{p-1}$ can be approximated by those in the Hamming space, up to a known nonlinearity. By absorbing this nonlinearity into the activation functions, we construct a new class of multi-layer neural networks that can be trained on source real-valued data and directly mapped along with the data to the binary hyperdimensional space, maintaining a comparable predictive performance.

Empirically, our approach circumvents the need to train expressive models directly in the binary domain, which is both high-dimensional ($N \gg p$) and structurally constrained---posing significant challenges for fitting predictive models. Theoretically, the proposed framework for constructing high-performing binary neural networks addresses key open questions in HDC research. First, we address questions about how far HDC models can be pushed to achieve accuracies on par with neural networks (asymptotically, the accuracies of the proposed hyperdimensional models converge to that of a G-Net as $N$ increases). Second, we provide non-asymptotic results on how large the hyperdimension $N$ needs to be for a desirably close performance to that of a G-Net. At a more abstract level, the proposed framework can be interpreted as an inexpensive access to a \emph{meta-distribution} over binary neural networks tailored to a prediction task, where G-Net controls its mean accuracy and the hyperdimension controls the level of concentration. Sampling from this distribution without training may benefit applications such as privacy, robustness, and ensemble learning. Moreover, individual binary networks drawn from this distribution can be further refined through fine-tuning techniques.

The remainder of the paper is organized as follows. Section~\ref{sec:BunEm} introduces the bundle embedding problem, which concerns embedding both data and predictive models coherently, with minimal loss in performance within the embedded space. Section~\ref{sec:GNetIdea} presents the core G-Net framework as a solution to this problem and outlines architectural variations designed for varying levels of hardware compatibility. In Section~\ref{Cons:sec:GNet}, we provide a detailed consistency analysis between the performance of a G-Net and its hyperdimensional counterpart (EHD G-Net), including non-asymptotic, layer-wise, and network-level results showing how the choice of hyperdimension governs the discrepancy between the two. Section~\ref{sec:rademacher} demonstrates, under realistic conditions, how the encoding can be significantly simplified by replacing Gaussian matrices with Rademacher matrices. Finally, Section~\ref{sec:experiments} presents numerical experiments that showcase the strong performance of the proposed method, followed by discussion and concluding remarks. Proofs of all theoretical results, along with additional experiments and implementation details, are provided in the Appendix.

\paragraph{Notations.}
Let \(\mathbb{S}^{n-1}\) and \(\mathbb{B}_2^n\) denote the unit sphere and unit Euclidean ball in \(\mathbb{R}^n\), respectively. We use lowercase letters for vectors and uppercase letters for matrices. For two binary vectors $x, y\in\{-1,1\}^n$, $\Dh(x,y)$ represents the normalized Hamming distance: $\Dh(x,y)=n^{-1}|\{i:x_i\neq y_i\}|$. For two vectors $x, y\in\eS^{n-1}$, $\Dg(x,y)$ represents the normalized geodesic distance between the two vectors obtained by dividing the smaller angle between the two vectors by $\pi$. A random vector/matrix is called Gaussian (or Rademacher) if the elements are independently drawn from a standard normal distribution (or a Rademacher distribution taking values 1 and $-1$ with equal probability $1/2$). We use script font for random quantities (e.g., $\mathrnd{g}$, $\mathrnd{G}$). We use teletype font for functions that operate on matrices/vectors in an element-wise fashion, e.g., $\texttt{sign}(x)$ is a vector with $\sign(x_i)$ as its $i$-th element.  

\section{Bundle Embedding Problem}\label{sec:BunEm}
Quantization simplifies data representation and computation to better suit hardware constraints. Its effectiveness is maximized when the geometric structure of the data is preserved, motivating the use of (nearly) distance-preserving mappings, often referred to as (near) isometries.

\begin{figure}[t]
\hspace{.25cm}
\begin{overpic}[trim={4cm 4cm  3cm 4cm},clip,height=1in]{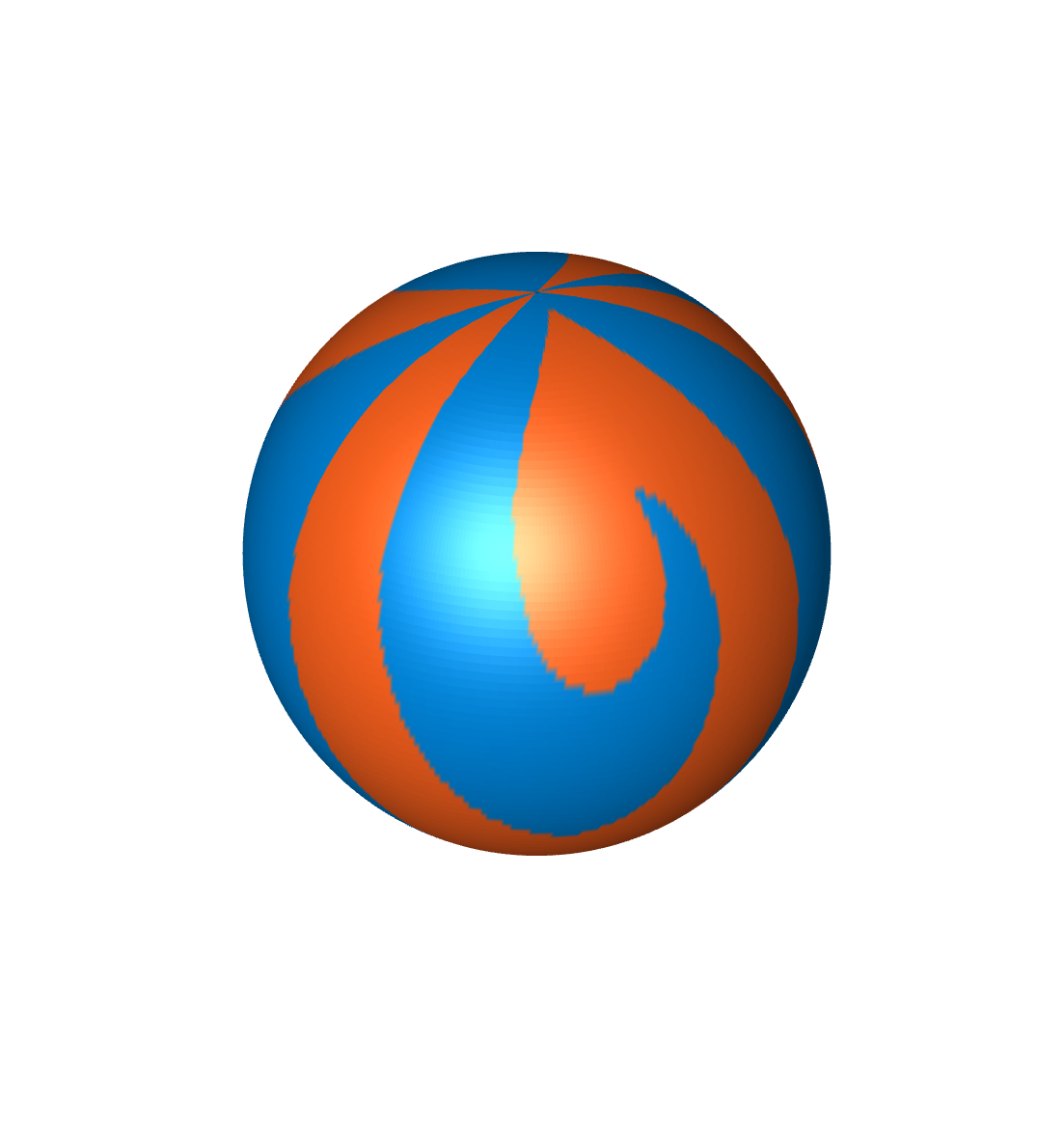}
\end{overpic}
\put (0,10) {\scalebox{1}{\rotatebox{0}{$\xlongrightarrow[\text{    }]{~~ \text{  $\hat{\mathrnd{x}}=\sign(\hat{\mathrnd{G}}x)$   
    } ~~ }$}}}
\put (-53,-5) {\scalebox{.8}{\rotatebox{0}{$x\in\mathcal{X}\subseteq\mathbb{S}^2$}}}
\put (-49,-17) {\scalebox{.8}{\rotatebox{0}{$p_\theta(y\mid x)$}}}
\hspace{2.5cm}
\begin{overpic}[trim={3.5cm 5cm  3cm 4cm},clip,height=1.05in]{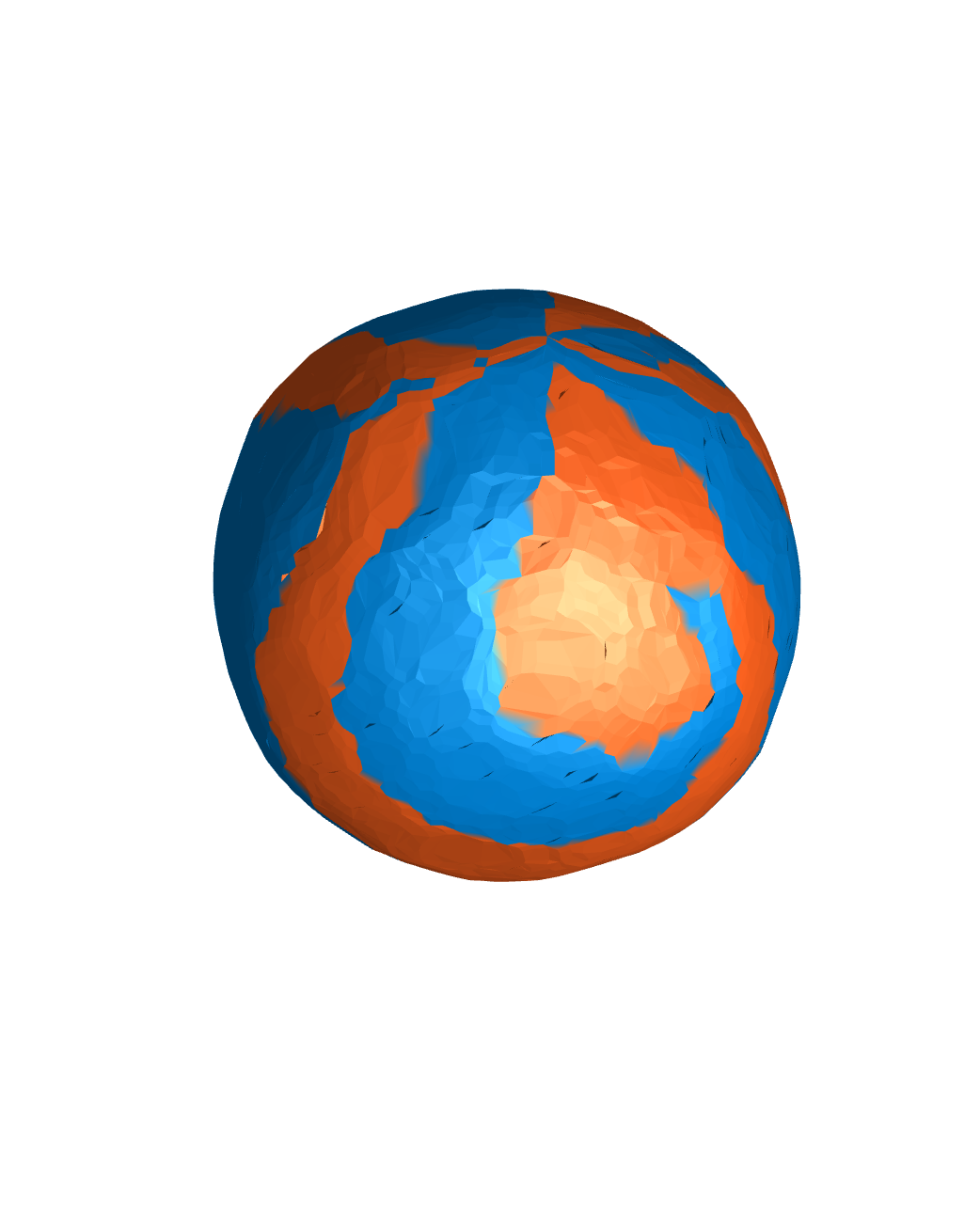}
\end{overpic}
\put (14,18) {\scalebox{.85}{\rotatebox{0}{$\longrightarrow$}}}
\put (-5,11) {\scalebox{.55}{\rotatebox{0}{$PC_1$}}}
\put (14,19) {\scalebox{.85}{\rotatebox{90}{$\longrightarrow$}}}
\put (24,13) {\scalebox{.55}{\rotatebox{0}{$PC_2$}}}
\put (18.5,31) {\scalebox{.55}{\rotatebox{0}{$PC_3$}}}
\put (5.5,22) {\scalebox{.75}{\rotatebox{-135}{$\longrightarrow$}}}
\put (-72,-5) {\scalebox{.8}{\rotatebox{0}{$\mathcal{P}_3(\hat{\mathrnd{x}}), ~\hat{\mathrnd{x}}\in\hat{\mathcal{X}}\subseteq\{-1,1\}^{100}$}}}
\put (-58,-17) {\scalebox{.8}{\rotatebox{0}{$\hat{p}_\theta(y\mid \hat{\mathrnd{x}})\approx p_\theta(y\mid x)$}}}
\hspace{2.5cm}
\begin{overpic}[trim={3.5cm 4.75cm  3cm 4cm},clip,height=1.0in]{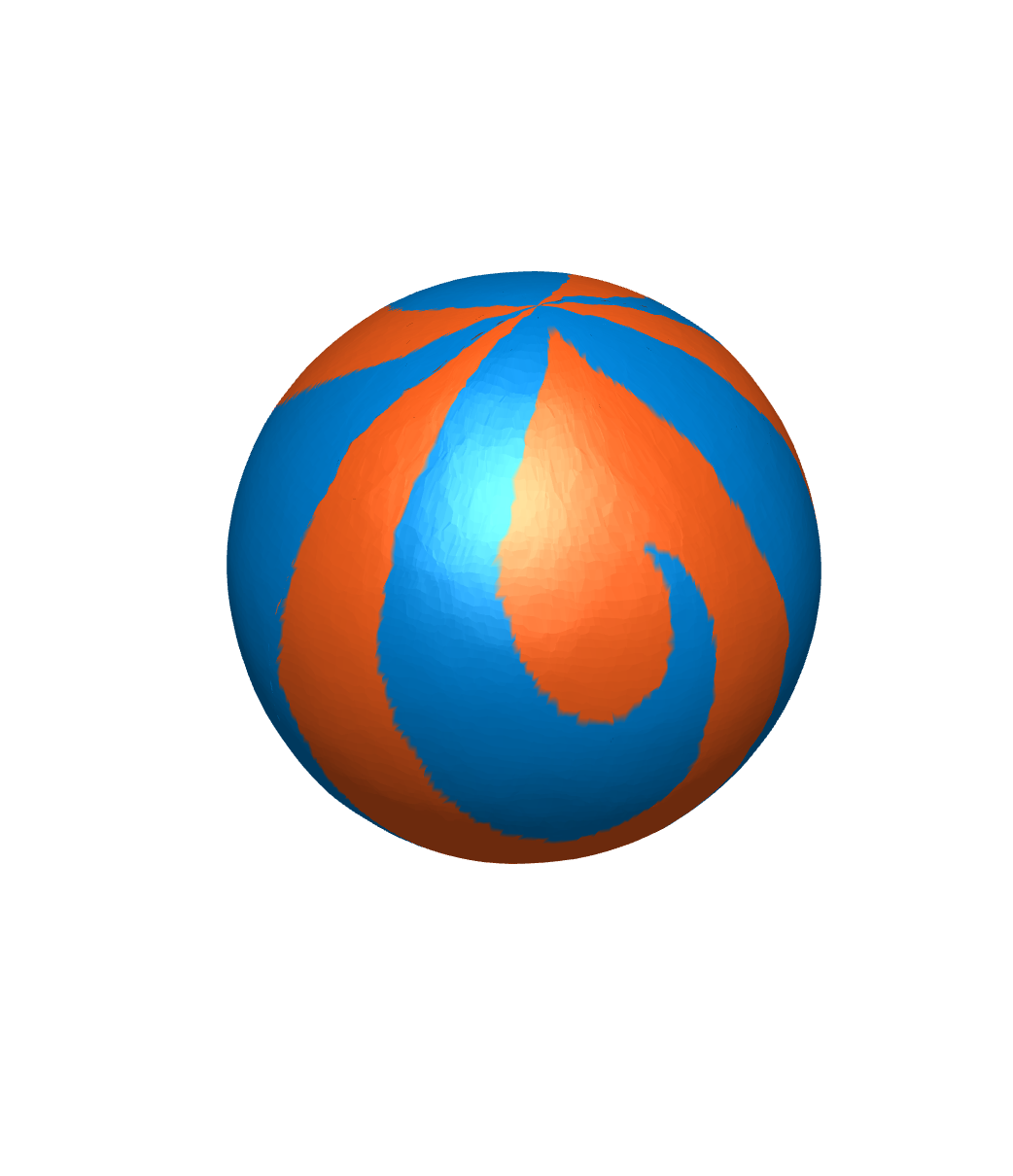}
\end{overpic}
\put (-285,70) {\scalebox{1}{\rotatebox{0}{ $\xlongrightarrow[\text{    }]{ \quad \qquad \qquad \qquad  \text{  $\tilde{\mathrnd{x}}=\sign( \tilde{\mathrnd{G}}x)$   
    }\qquad \qquad \qquad \qquad   }$}}}
\put (-72,-6) {\scalebox{.8}{\rotatebox{0}{$\mathcal{P}_3(\tilde{\mathrnd{x}}), ~\tilde{\mathrnd{x}}\in\tilde{\mathcal{X}}\subseteq\{-1,1\}^{1000}$}}}
\put (-58,-17) {\scalebox{.8}{\rotatebox{0}{$\tilde{p}_\theta(y\mid \tilde{\mathrnd{x}})\approx p_\theta(y\mid x)$}}}
\put (14,15) {\scalebox{.85}{\rotatebox{0}{$\longrightarrow$}}}
\put (-5,8) {\scalebox{.55}{\rotatebox{0}{$PC_1$}}}
\put (14,16) {\scalebox{.85}{\rotatebox{90}{$\longrightarrow$}}}
\put (24,10) {\scalebox{.55}{\rotatebox{0}{$PC_2$}}}
\put (18.5,28) {\scalebox{.55}{\rotatebox{0}{$PC_3$}}}
\put (5.5,18.5) {\scalebox{.75}{\rotatebox{-135}{$\longrightarrow$}}}\vspace{0cm}
\caption{A graphical demonstration of data embedding, and bundle embedding}\label{encoding:fig}\vspace{-.4cm}
\end{figure}
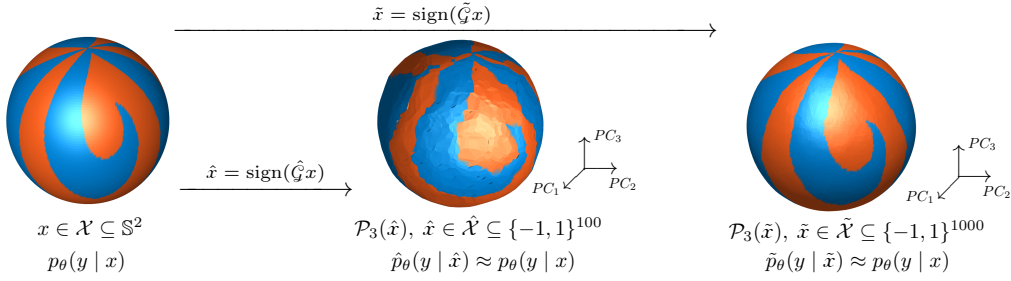

\begin{definition}
    Suppose that $\varepsilon>0$, and let $\mathcal{X}$ and $\mathcal{Y}$ be metric spaces equipped with distances $\mathcal{D}_{\mathcal{X}}$ and $\mathcal{D}_{\mathcal{Y}}$, respectively. A mapping $\mathcal{A}:\mathcal{X}\to\mathcal{Y}$ is called an $\varepsilon$-near-isometry if for all $x_1,x_2\in \mathcal{X}$: $
   \left| \mathcal{D}_{\mathcal{Y}}\left(\mathcal{A}(x_1),\mathcal{A}(x_2)\right) - \mathcal{D}_{\mathcal{X}}(x_1,x_2)\right|\leq \epsilon.$ 
\end{definition}
A prominent example of such mappings is the \emph{random binary embedding}, where real-valued vectors are converted to high-dimensional binary codes via a random linear projection followed by a quantization operator (such as a $\sign$ function). Typically, such embeddings takes the form $\mathcal{A}(x) = \texttt{sign}(\mathrnd{G}x)$, where $\mathrnd{G}$ is a Gaussian matrix. It is well-understood that by increasing the height of $\mathrnd{G}$, the mapping can be made arbitrarily close to an isometry, as stated below. 
\begin{proposition}\label{G:Isometry}
    For a Gaussian matrix $\mathrnd{G}\in\R^{N\times n}$, consider the mapping  $\mathcal{A}(x) = \texttt{sign}(\mathrnd{G}x)$ from $\mathcal{X} = \eS^{n-1}$ (equipped with a geodesic distance $\Dg$) to $\mathcal{Y}=\{-1,1\}^N$ (equipped with a normalized Hamming distance $\Dh$). Fix $\varepsilon\in(0,1)$, then there exist absolute constants $c_1$ and $c_2$ such that if $N\geq c_1n/\varepsilon^2$, with probability exceeding $1-\exp(-c_2 n)$, $\mathcal{A}$ is an $\varepsilon$-near-isometry. That is
    \begin{equation}\label{near:iso}
    \forall x,y\in \eS^{n-1}: \qquad \left | \Dh\left( \texttt{sign}\left( \mathrnd{G}x\right), \texttt{sign}\left( \mathrnd{G}y\right)\right) -    \Dg(x,y)  \right| \leq \varepsilon. 
    \end{equation}
\end{proposition}
Throughout this paper, we refer to \(\mathcal{X}\) as the \emph{primal} space and \(\mathcal{Y}\) as the \emph{embedded} space. For sufficiently large $N$, Proposition \ref{G:Isometry} guarantees, with high probability, an $\varepsilon$-consistency between the pairwise distances measured in the primal space $\mathcal{X}=\eS^{n-1}$ and the embedded space $\mathcal{Y}=\{-1,1\}^N$. To be able to use a similar metric on both spaces, for the same setting as Proposition \ref{G:Isometry}, one may consider the normalized mapping $\mathcal{A}(x) =N^{-1/2} \texttt{sign}(\mathrnd{G}x)$ which will map unit vectors in $\eS^{n-1}$ to unit vectors in $\eS^{N-1}$. As detailed in the proof discussion of Proposition \ref{G:Isometry}, in this case, the consistency result \eqref{near:iso} can be reformulated as
\begin{equation}\label{near:iso:2}
    \forall x,y\in \eS^{n-1}: \qquad \left | \sin^2\left(\frac{\pi}{2}\Dg\left( N^{-1/2}\texttt{sign}\left( \mathrnd{G}x\right), N^{-1/2}\texttt{sign}\left( \mathrnd{G}y\right)\right)\right) -    \Dg(x,y)  \right| \leq \varepsilon, 
\end{equation}
where a geodesic metric is used on both spaces. Equation \eqref{near:iso:2} reveals that no matter how small $\varepsilon$ is, there is an intrinsic $\sin^2(\frac{\pi}{2}\cdot )$ deformation of the geodesic distance when applying $\mathcal{A}$ to the source data. To visually illustrate the effect of this embedding, Figure~\ref{encoding:fig} shows the result of applying \(\mathcal{A}\) to a dense set of points on \(\mathbb{S}^2\), using \(N=100\) and \(N=1000\). The binary codes produced in \(\{-1,1\}^N\) are then visualized in \(\mathbb{R}^{3}\) via their first three principal components to illustrate how the representation improves as $N$ increases. Although the spherical geometry is nearly intact for $N=1000$, the deformation is noticeable when comparing the color labels—the coloring pattern on the right deviates from that of the original on the left.


In this paper, the \emph{bundle embedding problem} broadly refers to the problem of coherently embedding both data and a predictive model defined on it, such that the model remains functional in the embedded space without requiring retraining. As a motivating example, suppose that a model  $p_\theta(y\!\mid\! x)$ is trained on the source data $\mathcal{X}$ illustrated in Figure~\ref{encoding:fig}. After binary embedding, such a model is unusable due to two main issues: first, the data format has changed from continuous to binary, and, second, there is a nonlinear distortion introduced by the embedding, which standard predictors are not equipped to correct easily. On the other hand, training a high-accuracy classifier in the embedded space that operates in binary mode is itself difficult, owing to both the discreteness of the problem and the often-large dimension $N$. In this case our goal is to construct predictive models that can be embedded along with the data to operate in binary mode while maintaining accuracy.

To this end, we introduce a flexible class of neural networks, whose performance remains close after the data are quantized via a random sign embedding, ensuring consistent accuracies in both the primal and embedded domains. Rewiring these models to their binary neural network counterparts is as simple as going through a random sign embedding of the primal model weights. In a hyperdimensional-computing pipeline, this approach supplies high-accuracy predictors that operate directly on the binary-represented data, obviating any training in the embedded domain.
\vspace{-.2cm}
\section{G-Net Main Idea}\vspace{-.1cm}\label{sec:GNetIdea}
At a high level, many standard feed-forward neural networks can be viewed as a cascade of $L$ layers given by
$y_\ell = \sigma_\ell(W_\ell y_{\ell-1})$,
where $W_\ell$ are learnable weight matrices, $y_\ell$ is the output of the $\ell$-th layer, and $\sigma_\ell$ is a nonlinear, typically element-wise, activation function. In this section, we discuss the bundle embedding of a class of neural networks originally operating on real-valued data, demonstrating how they can be systematically transformed to operate on binary data while maintaining accuracy controllably close to the original real-valued model. The construction is grounded in Grothendieck’s identity (cf.\ \cite[page~63]{vershynin2018high}); hence, we refer to the proposed architectures as \emph{G-Nets}.
\begin{lemma}[Grothendieck's Identity]  \label{grothendieck}
Let $\mathrnd{g}$ be a standard Gaussian vector in $\mathbb{R}^p$.  Then, for  fixed vectors $u,v \in \mathbb{R}^{p}$:
$
\mathbb{E}\left[ \sign(\mathrnd{g}^\top {u}) \sign(\mathrnd{g}^\top {v}) \right] = \frac{2}{\pi} \arcsin\left( \frac{{u}^\top {v}}{\|u\|_2\|v\|_2} \right).
$
\end{lemma}
If $\mathrnd{G}\in\R^{N\times p}$ is a standard Gaussian matrix with rows $\mathrnd{g_i^\top}$, then by the law of large numbers\vspace{-.2cm}
\[u,v\in\eS^{p-1}: ~~~
\frac{1}{N} \left\langle \texttt{sign}(\mathrnd{G}u), \texttt{sign}(\mathrnd{G}v)\right\rangle = \frac{1}{N}\sum_{i=1}^N\sign(\mathrnd{g}_i^\top u)\sign(\mathrnd{g}_i^\top v)\approx \frac{2}{\pi}\arcsin(\langle u, v\rangle ). 
\]
This observation offers a new perspective on Grothendieck's Identity: through the random sign embedding, inner products in $\eS^{p-1}$ are approximately mapped to inner products in $\{-1,1\}^N$ up to some scaling and sine distortion.
More generally, when $W \in \mathbb{R}^{n \times p}$ has normalized rows (that is, $w_i^\top \in \mathbb{S}^{p-1}$, $i\in\{1,\ldots,n\}$), and $y \in \mathbb{S}^{p-1}$, then $\frac{1}{N} \texttt{sign}\left(W \mathrnd{G}^\top\right) \texttt{sign}\left(\mathrnd{G} y\right) \approx \frac{2}{\pi} \texttt{arcsin} (W y)$,
where $\texttt{sign}$ and $\texttt{arcsin}$ functions act element-wise. Applying a suitable activation $\tau$ (later to be specified) can preserve this approximation, and for a general input $y \in \mathbb{R}^p$ deliver a neural network layer of the form
\begin{equation} \label{networkwithG}
\sigma\left(W \frac{y}{\|y\|_2} \right)\approx \tau \left(    \texttt{sign}\left(W\mathrnd{G}^\top\right) \texttt{sign}\left( \mathrnd{G} y\right) \right),
\end{equation}
where $\sigma = \tau \circ \frac{2N}{\pi}\texttt{arcsin}$. Observe that the left-hand side of \eqref{networkwithG} operates on inner products of real-valued vectors, while the main right-hand side operation is inner products between their binary embeddings. The activations differ between the two sides, since $\sigma$ further carries the nonlinear distortion introduced by Grothendieck's identity. If the discrepancy between the two sides of \eqref{networkwithG} is controllably small (specifically, by picking a large $N$), then cascading multiple layers---embedded via independent Gaussian matrices $\mathrnd{G}_\ell$---forms a neural network operating in the embedded regime, with performance that closely matches its primal counterpart. The above construction and conversion can be extended to $L$ layers as follows.

\begin{definition}[G-Net] \label{gnet}
Let $n_0,\ldots,n_L$ be a sequence of positive integers, and $\tau$ be an activation function. Suppose that $W_\ell$ is an $n_\ell \times n_{\ell-1}$ matrix whose rows are $\ell_2$-normalized for $\ell=1,\ldots,L$, and let an input vector $y_0=x \in \mathbb{R}^{n_0}$ be given. We define the output $y_L$ of an $L$-layer G-Net iteratively by $y_\ell = \sigma \left( W_\ell \frac{y_{\ell-1}}{\|y_{\ell-1}\|_2} \right)$, for $\ell = 1,\ldots,L$, where $\sigma = \tau \circ \frac{2}{\pi} \arcsin$.
\end{definition}

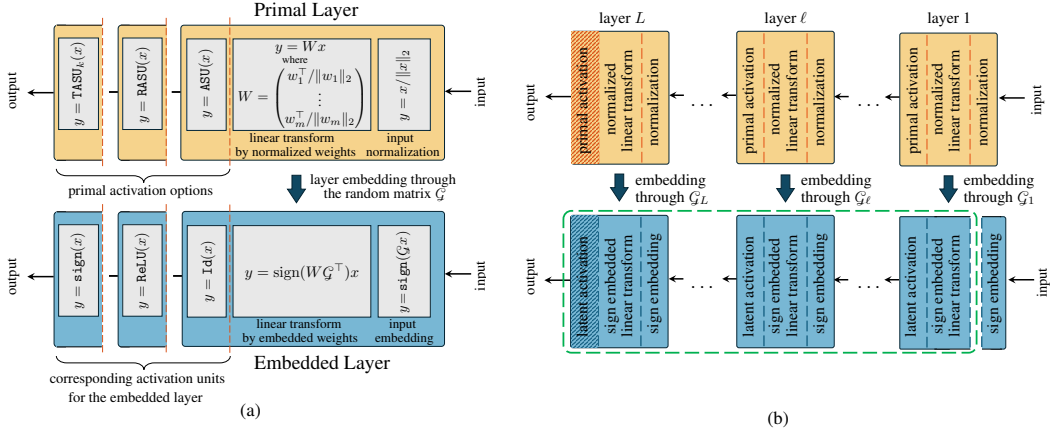
\begin{figure}[t]
\begin{center}
\begin{overpic}[trim={4cm 0cm  3cm -1cm},clip,height=2in]{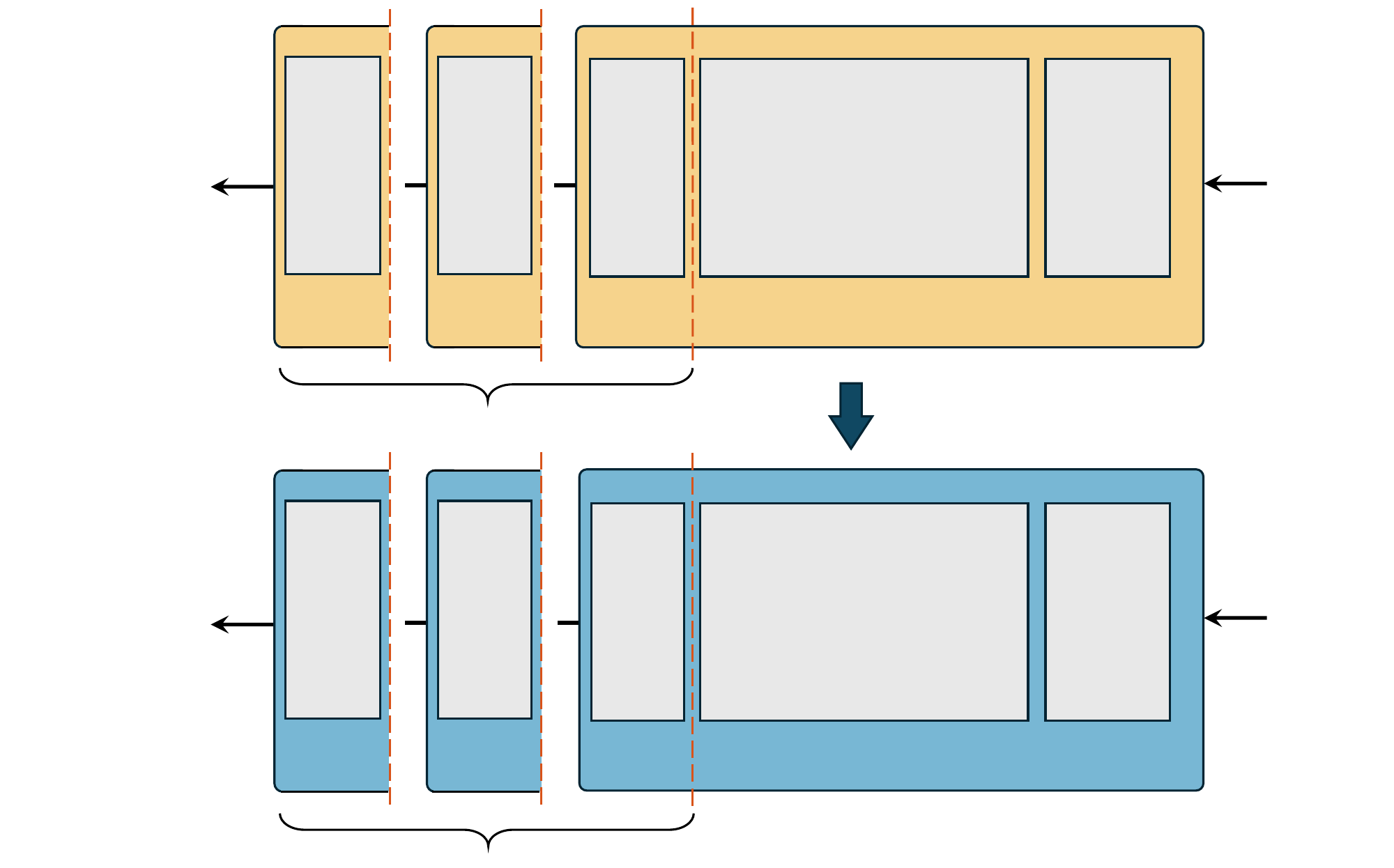}
\put (83.5,55.5) {\scalebox{.55}{\rotatebox{90}{$y = x/\|x\|_2$  }}}
\put (81.5,51) {\scalebox{.5}{\rotatebox{0}{input}}}
\put (77.1,48) {\scalebox{.5}{\rotatebox{0}{normalization}}}
\put (57.5,70.6) {\scalebox{.55}{\rotatebox{0}{$y = Wx$  }}}
\put (59,68) {\scalebox{.4}{\rotatebox{0}{where  }}}
\put (49.5,59.7) {\scalebox{.54}{\rotatebox{0}{$W =  \begin{pmatrix}w_1^\top/\|w_1\|_2 \\ \vdots \\ w_m^\top/\|w_m\|_2 \end{pmatrix} $ }}}
\put (52,51) {\scalebox{.5}{\rotatebox{0}{linear transform}}}
\put (49,48) {\scalebox{.5}{\rotatebox{0}{by normalized weights}}}
\put (27.5,55) {\scalebox{.55}{\rotatebox{90}{$y = \texttt{RASU}(x)$  }}}
\put (41,56.1) {\scalebox{.55}{\rotatebox{90}{$y = \texttt{ASU}(x)$  }}}
\put (13,39.5) {\scalebox{.55}{\rotatebox{0}{primal activation options}}}

%
\put (83.5,14.2) {\scalebox{.54}{\rotatebox{90}{$y \!=\! \texttt{sign}(\mathrnd{G}x)$  }}}
\put (82,11) {\scalebox{.5}{\rotatebox{0}{input}}}
\put (79,8) {\scalebox{.5}{\rotatebox{0}{embedding}}}
\put (51,22) {\scalebox{.58}{\rotatebox{0}{$y = \sign(W\mathrnd{G}^\top)x$  }}}
\put (53.5,11) {\scalebox{.5}{\rotatebox{0}{linear transform}}}
\put (50.5,8) {\scalebox{.5}{\rotatebox{0}{by embedded weights}}}
\put (27.3,15) {\scalebox{.55}{\rotatebox{90}{$y = \texttt{ReLU}(x)$  }}}
\put (41.5,16.5) {\scalebox{.55}{\rotatebox{90}{$y = \texttt{Id}(x)$  }}}
\put (13.5,54.5) {\scalebox{.55}{\rotatebox{90}{$y = \texttt{TASU}_k(x)$  }}}
\put (13.5,15) {\scalebox{.55}{\rotatebox{90}{$y = \texttt{sign}(x)$  }}}
\put (65,42) {\scalebox{.55}{\rotatebox{0}{layer embedding through}}}
\put (68,39) {\scalebox{.55}{\rotatebox{0}{the random matrix $\mathrnd{G}$}}}
\put (100.5,19) {\scalebox{.55}{\rotatebox{90}{input}}}
\put (100.5,58.5) {\scalebox{.55}{\rotatebox{90}{input}}}
\put (0,18) {\scalebox{.55}{\rotatebox{90}{output}}}
\put (0,58) {\scalebox{.55}{\rotatebox{90}{output}}}
\put (53,78) {\scalebox{.75}{\rotatebox{0}{Primal Layer}}}
\put (53,2) {\scalebox{.75}{\rotatebox{0}{Embedded Layer}}}
\put (9,-1.) {\scalebox{.55}{\rotatebox{0}{corresponding activation units}}}
\put (13,-5.) {\scalebox{.55}{\rotatebox{0}{for the embedded layer}}}
\put (50,-8) {\scalebox{.7}{\rotatebox{0}{(a)}}}
\end{overpic}\hspace{.1cm}
\begin{overpic}[trim={.7cm -0.3cm  .1cm 0cm},clip,height=1.9in]{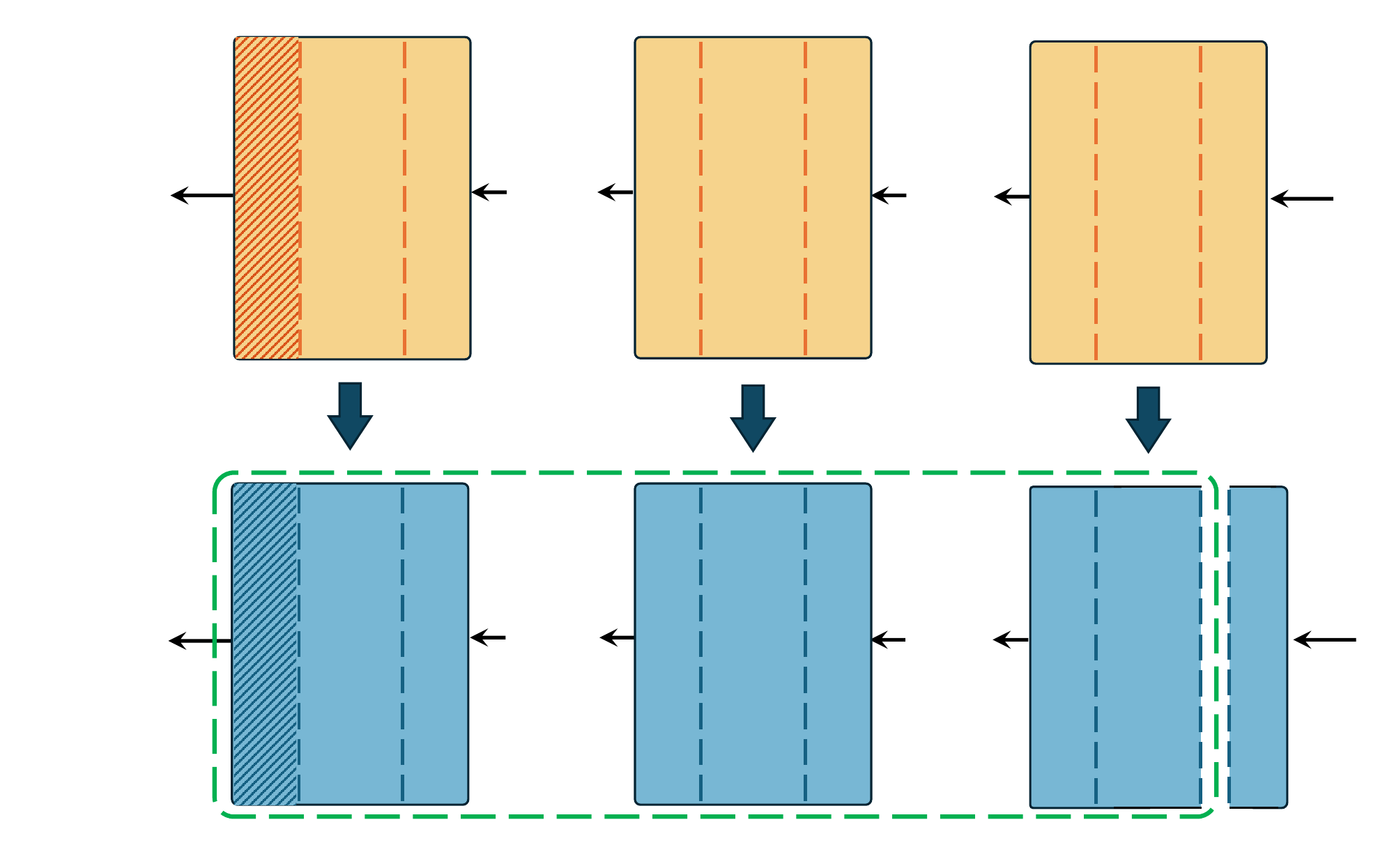}
\put (98,14.5) {\scalebox{.55}{\rotatebox{90}{input}}}
\put (97,47) {\scalebox{.55}{\rotatebox{90}{input}}}
\put (7,15) {\scalebox{.55}{\rotatebox{90}{output}}}
\put (7,46) {\scalebox{.55}{\rotatebox{90}{output}}}
\put (50,-8) {\scalebox{.7}{\rotatebox{0}{(b)}}}
\put (36.3,15.5) {\scalebox{.75}{\rotatebox{0}{$\cdots$}}}
\put (36.3,48) {\scalebox{.75}{\rotatebox{0}{$\cdots$}}}
\put (66,15.5) {\scalebox{.75}{\rotatebox{0}{$\cdots$}}}
\put (66,48) {\scalebox{.75}{\rotatebox{0}{$\cdots$}}}
%
\put (87.3,41.5) {\scalebox{.6}{\rotatebox{90}{normalization}}}
\put (79.3,43.5) {\scalebox{.6}{\rotatebox{90}{normalized}}}
\put (82.3,40.5) {\scalebox{.6}{\rotatebox{90}{linear transform}}}
\put (74.8,40) {\scalebox{.6}{\rotatebox{90}{primal activation}}}
%
\put (58.2,41.5) {\scalebox{.6}{\rotatebox{90}{normalization}}}
\put (50.2,43.5) {\scalebox{.6}{\rotatebox{90}{normalized}}}
\put (53.4,40.5) {\scalebox{.6}{\rotatebox{90}{linear transform}}}
\put (45.6,40) {\scalebox{.6}{\rotatebox{90}{primal activation}}}
\put (28.6,41.5) {\scalebox{.6}{\rotatebox{90}{normalization}}}
\put (21,43.5) {\scalebox{.6}{\rotatebox{90}{normalized}}}
\put (24,40.5) {\scalebox{.6}{\rotatebox{90}{linear transform}}}
\put (16,40) {\scalebox{.6}{\rotatebox{90}{primal activation}}}
%
\put (89.2,8) {\scalebox{.6}{\rotatebox{90}{sign embedding}}}
\put (79.3,8) {\scalebox{.6}{\rotatebox{90}{sign embedded}}}
\put (82.3,7.5) {\scalebox{.6}{\rotatebox{90}{linear transform}}}
\put (74.5,7.5) {\scalebox{.6}{\rotatebox{90}{latent activation}}}
%
\put (58.2,8) {\scalebox{.6}{\rotatebox{90}{sign embedding}}}
\put (53.4,7.5) {\scalebox{.6}{\rotatebox{90}{linear transform}}}
\put (50.2,8) {\scalebox{.6}{\rotatebox{90}{sign embedded}}}
\put (45.6,7.5) {\scalebox{.6}{\rotatebox{90}{latent activation}}}
\put (28.6,8) {\scalebox{.6}{\rotatebox{90}{sign embedding}}}
\put (21,8) {\scalebox{.6}{\rotatebox{90}{sign embedded}}}
\put (24,7.5) {\scalebox{.6}{\rotatebox{90}{linear transform}}}
\put (16,7.5) {\scalebox{.6}{\rotatebox{90}{latent activation}}}
\put (85,34.5) {\scalebox{.6}{\rotatebox{0}{embedding}}}
\put (85,31.5) {\scalebox{.6}{\rotatebox{0}{through $\mathrnd{G}_1$}}}
\put (56,34.5) {\scalebox{.6}{\rotatebox{0}{embedding}}}
\put (56,31.5) {\scalebox{.6}{\rotatebox{0}{through $\mathrnd{G}_\ell$}}}
\put (26.5,34.5) {\scalebox{.6}{\rotatebox{0}{embedding}}}
\put (26.5,31.5) {\scalebox{.6}{\rotatebox{0}{through $\mathrnd{G}_L$}}}
\put (78.5,63.5) {\scalebox{.6}{\rotatebox{0}{layer 1}}}
\put (49.5,63.5) {\scalebox{.6}{\rotatebox{0}{layer $\ell$}}}
\put (20,63.5) {\scalebox{.6}{\rotatebox{0}{layer $L$}}}
\end{overpic}
\end{center}\vspace{.2cm}
\caption{(a) A G-Net layer (top), and its corresponding EHD G-Net layer (bottom); the input to each block is denoted by $x$ and the output is represented by $y$; (b) A G-Net and corresponding EHD G-Net constructed by cascading the proposed layer blocks}
\label{FigGNet:LayNet}\vspace{-.4cm}
\end{figure}

\begin{definition}[EHD G-Net: Embedded Hyperdimensional G-Net] \label{EHDGNet}Suppose a trained G-Net is given as in  Definition \ref{gnet}. Fix a sequence of hyperdimensions $N_1,\ldots,N_L$, and let $\mathrnd{G}_\ell$ be independent Gaussian matrices of size $N_\ell \times n_{\ell-1}$.  We define the output $\tilde{\mathrnd{y}}_L$ of the $L$-layer Hyperdimensional embedding of the given G-Net iteratively by
$\tilde{\mathrnd{y}}_\ell = \tau(  \sign(W_\ell G_\ell^\top) \sign(G_\ell \tilde{\mathrnd{y}}_{\ell-1})),$ for $\ell=1,\ldots,L$. \vspace{-.2cm}
\end{definition}
We propose three element-wise choices for \(\tau\). The simplest choice is the identity map, \(\tau(x) = \texttt{Id}(x)\), primarily used for analysis purposes, which corresponds to the arc-sine activation unit $\sigma(x) = \texttt{ASU}(x)\triangleq \frac{2}{\pi}\texttt{arcsin}(x)$ in the G-Net architecture. Additionally, we use $\tau(x) = \texttt{ReLU}(x)$, and $\tau(x) = \texttt{sign}(x)$, which lead respectively to the following G-Net activations $\sigma$:
\[
\texttt{RASU}(x) \triangleq \begin{cases}
\frac{2}{\pi}\arcsin(x), & x\geq 0,\\[5pt]
0, & x<0,
\end{cases}
\quad\text{and}\quad
\texttt{TASU}_\kappa(x) \triangleq \tanh\left(\frac{2\kappa}{\pi}\arcsin(x)\right).
\]
In the last activation, we employed a smooth approximation of the \(\texttt{sign}\) function, specifically \(\tau(x) = \tanh(\kappa x)\) for sufficiently large \(\kappa\), to ensure that the associated G-Net activation remains smooth. The top panel of Figure \ref{FigGNet:LayNet}(a) presents the architecture of a G-Net layer, and the bottom panel depicts the corresponding EHD G-Net layer. 
An $L$-layer G-Net and correspondingly a similar-depth EHD G-Net can be constructed through the composition of the proposed layers. Note that in Definition~\ref{EHDGNet} and Figure~\ref{FigGNet:LayNet}, we omitted the normalization factors \(1/N_\ell\). In intermediate layers, this factor can be absorbed by the subsequent $\texttt{sign}$ operation. For the final layer, neglecting this normalization merely affects the output's scale, which is typically inconsequential.

The core bundle embedding process involves first training a G-Net on the original real-valued data to learn the weights \(W_\ell\), and subsequently converting it into an EHD G-Net by selecting sufficiently tall Gaussian matrices \(\mathrnd{G}_\ell\). The resulting EHD G-Net is a sequential neural network featuring binary weights \(\texttt{sign}(W_\ell \mathrnd{G}_\ell^\top)\) and appropriately adjusted activations, which are acquired \emph{without training on the embedded binary data}. Once the EHD G-Net is constructed, the first sign embedding block of the first layer (i.e., $\texttt{sign}(\mathrnd{G}_1 x)$) can be interpreted as the random sign embedding of the source data, and the rest of the pipeline (shown as the dashed box in the bottom panel of Figure~\ref{FigGNet:LayNet}(b)) can be viewed as an embedding of the primal inference module. Since HDC models typically employ a simple inference stage (see Section \ref{sec:HDCTax} of the Appendix), the framework can equivalently be viewed as an HDC pipeline with multi-stage embedding: the first $L-1$ layers constitute the embedding stack, and the final classification layer serves as the inference block. Under either interpretation, unlike many current HDC frameworks, the designs of our embedding and inference modules are tightly coupled by construction.

The primary distinctions between a G-Net layer (top panel of Figure \ref{FigGNet:LayNet}(a)) and a standard feed-forward network layer are the \(\ell_2\)-normalization applied at the input, and the enforced row-wise normalization on the learnable weight matrices \(W\). While these constraints may initially seem to limit the flexibility and expressiveness of G-Nets, they have been shown to accelerate training and enhance the generalization accuracy \cite{salimans2016weight, luo2018cosine, wu2022exploring}, making them advantageous design choices rather than limitations. Furthermore, the choice of activation function---being the sole distinction between RASU and TASU G-Nets---directly influences the operating mode of the resulting EHD G-Nets. In RASU networks, each layer output (i.e., the right-hand side of \eqref{networkwithG}) is a non-negative integer vector, whereas in TASU layers, the output is binary, allowing their EHD G-Nets to operate entirely in binary mode.

The rest of the paper delves more deeply into a theoretical understanding of the proposed framework.  
We first examine how the hyperdimension \(N\) controls the approximation error between a G-Net layer and its EHD counterpart, and how these errors accumulate when multiple layers are cascaded.  Although Grothendieck’s identity is naturally linked to Gaussian distributions, Section~\ref{sec:rademacher} shows that much simpler Rademacher embeddings offer comparable guarantees.

From a practical standpoint, the framework extends well beyond fully connected layers: convolution, batch-normalization, pooling, and similar modules can all be incorporated into a G-Net and subsequently embedded into a hyperdimensional model. In addition to addressing such details in Section \ref{sec:GNetImp} of the Appendix, we will also discuss details such as handling the bias term for all such layers, and the activation chosen for the last layer, which determines whether the resulting network behaves as a classifier or a regressor.
\vspace{-.2cm}
\section{Consistency of G-Net and EHD G-Net}\label{Cons:sec:GNet}\vspace{-.1cm}
This section presents a non-asymptotic analysis of the consistency between a G-Net layer and its hyperdimensional embedding, providing high-probability bounds on the required value of $N$ to achieve a target discrepancy between the primal and embedded layers. First, we analyze the RASU layer, which yields bounds analogous to those of an ASU layer.
\begin{theorem}\label{RASU:layer}
Consider a RASU (or ASU) layer with input $x \in \mathbb{S}^{p-1}$ and a weight matrix $W \in \mathbb{R}^{n \times p}$ whose rows are $\ell_2$-normalized. Let the layer output be given by $y = \texttt{RASU}(Wx)$ (or $y = \texttt{ASU}(Wx)$). Define the embedded output as $\tilde{\mathrnd{y}} = \texttt{ReLU}\left(\texttt{sign}(W \mathrnd{G}^\top)\, \texttt{sign}(\mathrnd{G} x)\right)$ (or $\tilde{\mathrnd{y}} =\texttt{Id}\left( \texttt{sign}(W \mathrnd{G}^\top)\, \texttt{sign}(\mathrnd{G} x)\right)$), where $\mathrnd{G} \in \mathbb{R}^{N \times p}$ is a standard Gaussian matrix. Then, for any $c > 0$, with probability at least $1 - \exp(-c)$:
$\left\| \frac{1}{N} \tilde{\mathrnd{y}} - y \right\|_2 \leq \sqrt{2N^{-1}(c + \log 2n )n}.$
\end{theorem}

\begin{corollary}
To ensure that the discrepancy between a RASU/ASU layer output and its hyperdimensional embedding satisfies $\left\| \frac{1}{N} \tilde{\mathrnd{y}} - y \right\|_2 \leq \varepsilon$, it is possible to choose the embedding dimension $N$ such that
\begin{equation}\label{RASU:N}
N = \mathcal{O}\left(\varepsilon^{-2}{n \log n}\right).
\end{equation}
\end{corollary}

A few remarks are in order. First, the $\log n$ term in \eqref{RASU:N} arises from reusing the same embedding matrix $\mathrnd{G}$ across all rows of $W$, which induces dependencies among components of the discrepancy vector. In the discussion following the proof of Theorem~\ref{RASU:layer} in the Appendix, we argue that if independence were enforced---e.g., by using a distinct embedding matrix $\mathrnd{G}_i$ for each row $w_i$ and computing $\tilde{\mathrnd{y}}_i = \texttt{ReLU}\left(\texttt{sign}(w_i^\top \mathrnd{G}_i^\top) \texttt{sign}(\mathrnd{G}_i x)\right)$---this would remove the $\log n$ factor but at a considerable cost in both computation and memory. Second, in our discrepancy analysis, we compare $y$ to the scaled output $\frac{1}{N} \tilde{\mathrnd{y}}$ purely for normalization purposes, since $\tilde{\mathrnd{y}}$ is integer-valued by construction. As stated before, this scaling does not affect downstream computations, as $\tilde{\mathrnd{y}}$ is passed through another sign-based embedding layer in subsequent processing, rendering the $1/N$ scaling irrelevant in practice.
Next, we analyze a TASU layer, where in addition to the standard discrepancy, one also needs to account for the error introduced by approximating the $\sign(\cdot)$ function with 
$\tanh(\kappa\cdot)$. 
\begin{theorem}\label{TASU:layer}
    Consider a TASU layer with output $y = \texttt{TASU}_\kappa( Wx)$, where $x \in \mathbb{S}^{p-1}$ and $W \in \mathbb{R}^{n \times p}$ has normalized rows $w_i^\top$ ($\|w_i\|_2=1$ for $i=1,\ldots,n$). Define the embedded layer output as $\tilde{\mathrnd{y}} = \texttt{sign}\left(\texttt{sign}(W \mathrnd{G}^\top)\, \texttt{sign}(\mathrnd{G} x)\right)$, where $\mathrnd{G} \in \mathbb{R}^{N \times p}$ is a standard Gaussian matrix. Assume 
    \begin{equation}\label{lmin:assump}
        \left|w_i^\top x\right|\geq \ell_{\min}>0, \qquad i=1,\ldots,n.
    \end{equation}
    Pick a target discrepancy $\varepsilon\leq \sqrt{n}$ and set
$\kappa \geq \frac{\pi}{2\ell_{\min}}\log \frac{4\sqrt{n}}{\varepsilon}.$
Then, for any scalar $c>0$, selecting $N \geq 8(c+\log 2n)n\kappa^2/\varepsilon^2$ guarantees that with probability exceeding $1-3\exp(-c)$: $\|y-\tilde{\mathrnd{y}}\|_2\leq \varepsilon$.
\end{theorem}
We emphasize that assumption~\eqref{lmin:assump} is essential for the analysis, to handle the discontinuity of the $\sign$ function. Specifically, when $w_i^\top x = 0$ for some row $i$, the TASU output yields $y_i = 0$, whereas the corresponding embedded output $\tilde{\mathrnd{y}}_i$ takes a binary $\pm 1$ value as the output of a $\sign$ function. Practically, a common approach to handle such zero inputs to a sign quantizer is to assign $\pm1$ randomly with equal probabilities. In our framework, this behavior emerges naturally from the symmetry of the Gaussian matrix $\mathrnd{G}$. Substituting the required $\kappa$ into the hyperdimension bound, we find that achieving an $\varepsilon$-discrepancy necessitates a hyperdimension of $N = \mathcal{O}({n \log^3 n}/(\varepsilon^2\ell_{\min}^2))$, which is larger compared to that required for a RASU layer. The increased scale of $\log^2 n/\ell_{\min}^2$ reflects the cost of constructing an embedded layer that, unlike the integer-valued outputs of RASU layers, produces binary codes at the layer's output.
\vspace{-.2cm}
\subsection{Network Consistency}\vspace{-.1cm}\label{sec:NetConsistency}
The preceding layer-wise result motivates us to analyze the overall discrepancy of the EHD G-Net as a cascade of multiple layers. For conventional feed-forward architectures, tracking the error accumulation across the network layers is straightforward (e.g., see \cite{aghasi2017net,aghasi2020fast}). In G-Net, however, the normalization operation complicates this tracking, requiring a more careful analysis. To streamline the discussions, while still covering the key technical steps, we focus on the base ASU G-Net case. Specifically, we consider a cascade of $L$ sequential layers, where
\begin{equation}\label{eq:seqNet}
y_\ell = \texttt{ASU}\left(W_\ell\frac{y_{\ell-1}}{\|y_{\ell-1}\|_2}\right),\qquad \tilde{\mathrnd{y}}_\ell = \texttt{Id}\left(\texttt{sign}\left(W_\ell \mathrnd{G}_{\ell}^\top\right)\texttt{sign}\left(\mathrnd{G}_\ell\tilde{\mathrnd{y}}_{\ell-1}\right)\right),\qquad \ell = 1,\ldots,L.
\end{equation}
Here $W_\ell\in\R^{n_\ell\times n_{\ell-1}}$ are the G-Net weights with normalized rows, and $\mathrnd{G_\ell}\in\R^{N_\ell\times n_{\ell-1}}$ are independent Gaussian matrices used for the embedding of each layer.  Since G-Net layers operate on normalized inputs, we define $\delta_\ell =  \frac{\tilde{\mathrnd{y}}_\ell}{\|\tilde{\mathrnd{y}}_\ell\|_2} - \frac{y_\ell}{\|y_\ell\|_2}$ to quantify the accumulated discrepancy up to the 
$\ell$-th layer. Clearly, since both networks are fed with identical inputs, $\delta_0 = 0$. 

While normalizing the layer inputs is a common practice to improve the training and generalization of neural networks \cite{salimans2016weight}, to avoid edge cases, the cascade analysis demands some assumptions about the degree of nonlinear distortion each layer introduces, as will be detailed below.
\begin{definition} A given matrix $W\in\R^{n\times p}$ with normalized rows is called consistent with near isometry of $\texttt{ASU}$ in set $\mathcal{D}\subseteq \mathbb{B}^p_2$, if for all $x,y\in\mathcal{D}$: 
\begin{equation}\label{near:isomx-y}
     \|x-y\|_2^2 - \varepsilon\leq \beta^{-1} \left\| \texttt{ASU}\left(Wx\right)-\texttt{ASU}\left(Wy\right) \right\|_2^2 \leq  \|x-y\|_2^2 + \varepsilon,
\end{equation}
where $\beta>0$ and $\varepsilon\in[0,1)$ are constants independent of $x$ and $y$ (possibly dependent on $n$ and $p$).
\end{definition}
When the layers of an ASU G-Net maintain the property in \eqref{near:isomx-y}, we can show that the network discrepancy scales linearly with the number of layers:
\begin{theorem} \label{th:netConsistency}Consider the cascade of $L$ G-Net layers and the corresponding hyperdimensional embedding as \eqref{eq:seqNet}. Assume that each layer $\ell$ of the network is consistent with near isometry of $\texttt{ASU}$ in $\eS^{n_{\ell-1}-1}$ with parameters $\beta_\ell$ and $\varepsilon_\ell$, as stated in \eqref{near:isomx-y}. Then with probability exceeding $1-L\exp(-c)$:
\begin{equation*}
  \|\delta_L\|_2 \leq   c' \sum_{\ell=1}^L \left(\sqrt{\frac{(c+\log n_\ell)n_\ell}{N_\ell\beta_\ell\left(1-\varepsilon_\ell\right)}} + \sqrt{\frac{\varepsilon_\ell}{1+\varepsilon_\ell}}\right),
\end{equation*}
where $\delta_L =  \frac{\tilde{\mathrnd{y}}_L}{\|\tilde{\mathrnd{y}}_L\|_2} - \frac{y_L}{\|y_L\|_2}$, and $c'$ is an absolute constant. \vspace{-.2cm}
\end{theorem}
A natural question is: what conditions on the set $\mathcal{D}$ or on the matrix 
$W$ are sufficient to ensure that \eqref{near:isomx-y} holds? One simple scenario occurs when 
$\mathcal{D}$ has a sufficiently small radius, so that \eqref{near:isomx-y} follows directly from the continuity of 
$\texttt{ASU}$ (the case with more restricted datasets). Beyond this rather trivial case, broadly speaking, one can show that when $W$ is picked to be a tall generic matrix with normalized rows, \eqref{near:isomx-y} is likely to hold within the entire unit ball $\mathbb{B}^p_2$. To state this rigorously, let's consider $\mathrnd{W}\in\R^{n\times p}$ of the form
\begin{equation}\label{eq:W}
\mathrnd{W} = \begin{bmatrix} \mathrnd{g}_1/\|\mathrnd{g}_1\|\\\vdots \\\mathrnd{g}_n/\|\mathrnd{g}_n\|\end{bmatrix},
\end{equation}
where $\mathrnd{g}_i\sim\mathcal{N}(0,I_p)$ are independent standard normal vectors.
    \begin{theorem}\label{near:ISO:Th}
        Consider $\mathrnd{W}\in\R^{n\times p}$,  following the construction format in \eqref{eq:W}, where $n\gtrsim p\geq 27$. Let $\mathrnd{g}_i\sim\mathcal{N}(0,I_p)$ be independent standard normal vectors. Then, for all $x, y\in \mathbb{B}_2^p$, with probability exceeding $1-c\exp(-c'p)$: $\left\|x-y \right\|_2^2 - \varepsilon_{n,p}^{l}\leq \beta_{n,p}^{-1}\left\|\texttt{ASU}(\mathrnd{W} x) - \texttt{ASU}(\mathrnd{W} y)\right\|^2  \leq \left\|x-y \right\|_2^2 + \varepsilon_{n,p}^{u}$, 
where \vspace{-.3cm}
\begin{equation*}
\beta_{n,p}^{-1} = \frac{\pi^2 p}{4\left(\sqrt{n}+\sqrt{p}\right)^2}, \qquad \varepsilon_{n,p}^l = c_l\frac{\sqrt{p}}{ \sqrt{n}+\sqrt{p}}, \qquad  \varepsilon_{n,p}^u = c_u\left( \frac{\sqrt{p}}{ \sqrt{n}+\sqrt{p}} + \frac{n+p^2}{p\left( \sqrt{n}+\sqrt{p}\right)^2}\right),\vspace{-.1cm}
\end{equation*}
and $c, c', c_l$ and $c_u$ are absolute numerical constants. 
\end{theorem}
Observe that by choosing $n$ to be a sufficiently large multiple of $p$, $\varepsilon_{n,p}^l$ and $\varepsilon_{n,p}^u$  in Theorem \ref{near:ISO:Th} can be made desirably small. The proof of the theorem precludes the use of concentration results for Lipschitz functions, due to the infinitely steep slope of the $\arcsin$ function around $\pm 1$. A technical contribution and key part of the proof is bounding a fourth-order induced norm of $\mathrnd{W}$ (which does not offer independence along the columns). Clearly, $W_\ell$ are determined by the G-Net training algorithm, however, by covering a large class of matrices, Theorem \ref{near:ISO:Th} showcases that when the trained weight matrices are tall and ``spread-out'', there is a good chance that \eqref{near:isomx-y} holds. It is also noteworthy that wide neural networks experience only minor changes in their weight distributions during training \cite{jacot2018neural} (see Section \ref{sec:MinorSh} of Appendix for more details). This provides an opportunity to control the weight distribution of the trained models. Empirically, in all the experiments performed in this paper, we randomly initialized the G-Net weights according to \eqref{eq:W}, and never observed an instance of poor consistency between a G-Net and its corresponding hyperdimensional embedding.
\vspace{-.2cm}    
\section{Rademacher Embedding}\vspace{-.2cm}\label{sec:rademacher}
One promising approach to enhance the efficiency of the proposed framework for hardware implementation is to utilize random vectors that are simpler to generate and apply  than Gaussian vectors. In this section, we demonstrate that even the most basic choice---a Rademacher vector---can be employed as the embedding matrix, provided that the G-Net weights are sufficiently ``spread out''---a condition that often holds in practice. The following result extends Grothendieck's lemma to Rademacher vectors.
\begin{theorem}[Approximate Grothendieck identity]\label{thm1} 
Suppose that $u,v \in \eS^{p-1}$ are unit vectors. Let $\mathrnd{r}$ be a Rademacher random vector in $\mathbb{R}^p$. Then,
$$
\left| \mathbb{E} \left[ \sign\left( \mathrnd{r}^\top u \right) \sign \left( \mathrnd{r}^\top v \right) \right]
- \frac{2}{\pi} \arcsin(u^\top v) \right| \le c g\left(u,\frac{v - \langle u,v\rangle u}{\|v - \langle u,v\rangle u\|_2}\right),
$$
where $c = 264$ is an absolute constant, and for $w$ and $w'$ in $\eS^{p-1}$, $g(w,w') = \sum_{i=1}^p (w_i^2+{w_i'}^2)^{3/2}$.
\end{theorem}
\begin{corollary}\label{corr:rad}
    Consider unit vectors $u, v\in\eS^{p-1}$ such that $\|u\|_\infty=\mathcal{O}(p^{-1/2})$,  $\|v\|_\infty=\mathcal{O}(p^{-1/2})$, and there exists some constant $c>0$ such that $|\langle u,v\rangle|\leq 1-c$. Then
    \[
 \left|\mathbb{E} \left[ \sign\left( \mathrnd{r}^\top u \right) \sign \left( \mathrnd{r}^\top v \right) \right]
- \frac{2}{\pi} \arcsin(u^\top v)\right|= \mathcal{O}\left(p^{-1/2}\right).
\]
\end{corollary}
Informally speaking, Corollary \ref{corr:rad} says that Grothendieck's identity approximately holds for Rademacher vectors when the inner product factors are dispersed. The layer consistency results of Section \ref{Cons:sec:GNet} can be readily extended to this substantially simpler encoding scheme. 
\begin{proposition}\label{RASU:layerRad}
Consider a similar RASU (or ASU) layer as Theorem \ref{RASU:layer}, where  for fixed constants $C,\delta > 0$, $\|x\|_\infty \le C p^{-1/2}$, $\|w_i\|_\infty \leq  C p^{-1/2}$, and $|w_i^\top x| \le 1 -\delta$ for all $i \in \{1,\ldots,n\}$. Define the embedded output as $\tilde{\mathrnd{y}} = \texttt{ReLU}\left(\texttt{sign}(W \mathrnd{R}^\top)\, \texttt{sign}(\mathrnd{R} x)\right)$ (or $\tilde{\mathrnd{y}} =\texttt{Id}\left( \texttt{sign}(W \mathrnd{R}^\top)\, \texttt{sign}(\mathrnd{R} x)\right)$), where $\mathrnd{R} \in \mathbb{R}^{N \times p}$ is a Rademacher matrix. Then, with high probability:
$
\| \frac{1}{N} \tilde{\mathrnd{y}} - y \|_2 \leq \mathcal{O}(\sqrt{N^{-1}n \log n} + \sqrt{n/p}).
$
\end{proposition}

\begin{proposition}\label{TASU:layerRad}
  Consider a similar TASU layer as Theorem \ref{TASU:layer},  where additionally for fixed constants $C,\delta > 0$, $\|x\|_\infty \le C p^{-1/2})$, $\|w_i\|_\infty \leq  C p^{-1/2}$, and $0<\ell_{\min}\leq |w_i^\top x| \le 1 -\delta$ for all $i \in \{1,\ldots,n\}$.  Define the embedded layer output as $\tilde{\mathrnd{y}} = \texttt{sign}\left(\texttt{sign}(W \mathrnd{R}^\top)\, \texttt{sign}(\mathrnd{R} x)\right)$, where $\mathrnd{R} \in \mathbb{R}^{N \times p}$ is a Rademacher matrix. Assume $C p^{-1/2} \le \ell_\text{min}/\pi$. Fix $\varepsilon\leq \sqrt{n}$ and set
$\kappa \geq \frac{\pi}{\ell_{\min}}\log \frac{4\sqrt{n}}{\varepsilon}.$
Then, picking $N = \mathcal{O}(n\kappa^2 \log n/\varepsilon^2)$ guarantees that $\|y-\tilde{\mathrnd{y}}\|_2 = \mathcal{O}(\varepsilon + \kappa \sqrt{n/p})$ with high probability.
\end{proposition}
In both cases, we observe an additional $\sqrt{n/p}$ discrepancy term which, unlike the Gaussian case, does not vanish by increasing $N$. However, taking into account the fact that $\|y\|_2$ scales with $\sqrt{n}$, the above results can be interpreted as the \emph{relative} discrepancy diminishing with increasing $N$ and $p$. While Gaussian embedding offers the best theoretical results, our numerical experiments show that Rademacher embedding achieves a comparably close performance.
\vspace{-.2cm}   
\section{Numerical Experiments and Concluding 
Discussion}\label{sec:experiments}\vspace{-.1cm}  
Among various applications of the proposed framework, a key contribution lies in enhancing the accuracy of predictive HDC models. Practical implementation details of G-Nets beyond standard fully connected layers---such as convolutional, pooling, classification, etc---are provided in Section \ref{sec:GNetImp} of the Appendix. Also, a detailed and extended version of the numerical experiments presented in this section, along with additional accuracy and robustness analysis across other datasets, is included in Section \ref{sec:NumericalExp}. Here, we selectively present classification results on MNIST, CIFAR-10, and human activity recognition (HAR-WSS \cite{uciHAR2012}).

\begin{figure}[!htbp]
\begin{center}
\begin{overpic}[trim={0.17cm -.25cm  0.25cm 0},clip,height=1.725in]{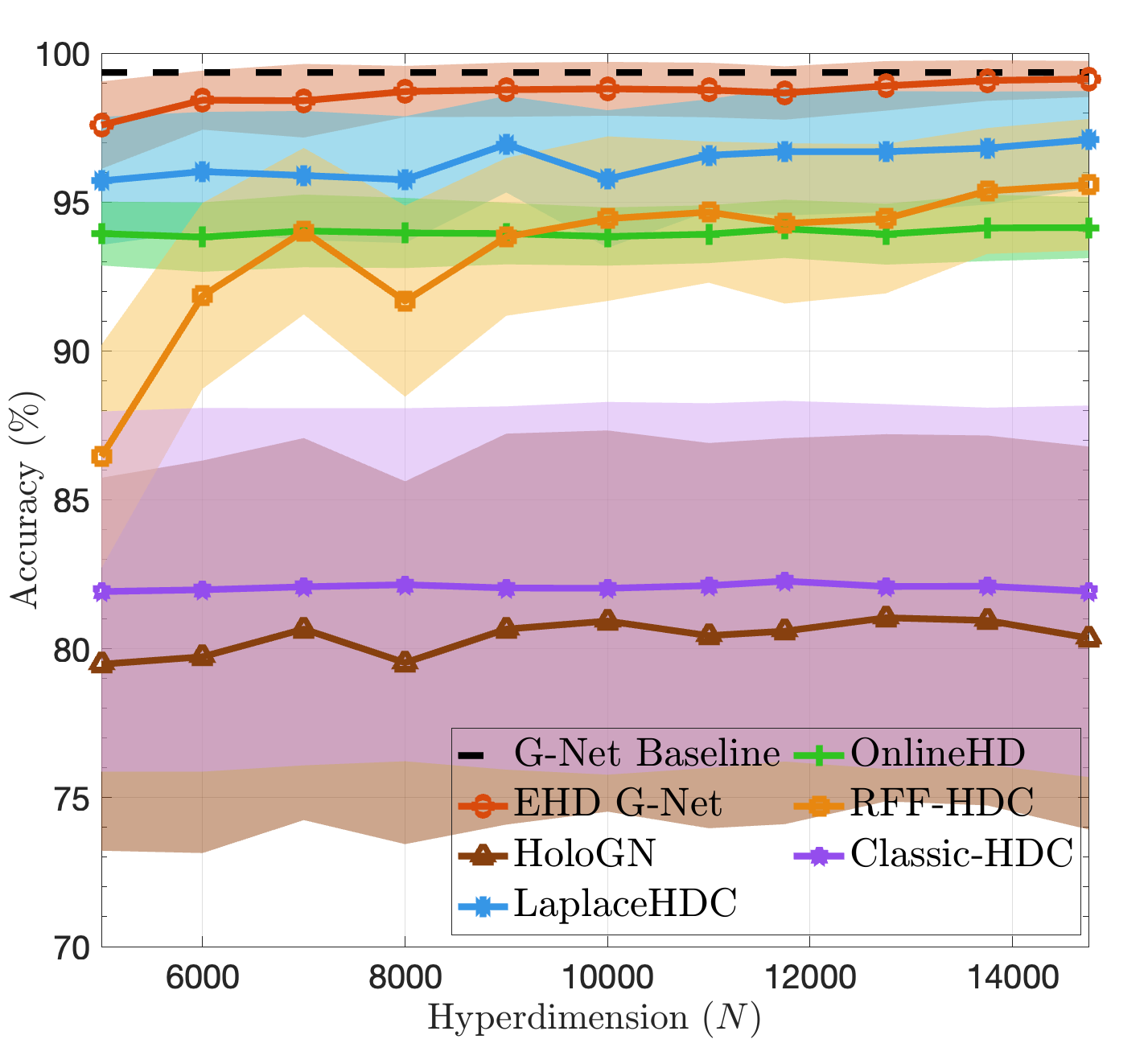}
\put (50,-3) {\scalebox{.7}{\rotatebox{0}{(a)}}}
\end{overpic}\hspace{.0cm}
\begin{overpic}[trim={0.17cm -.25cm  0.25cm 0},clip,height=1.725in]{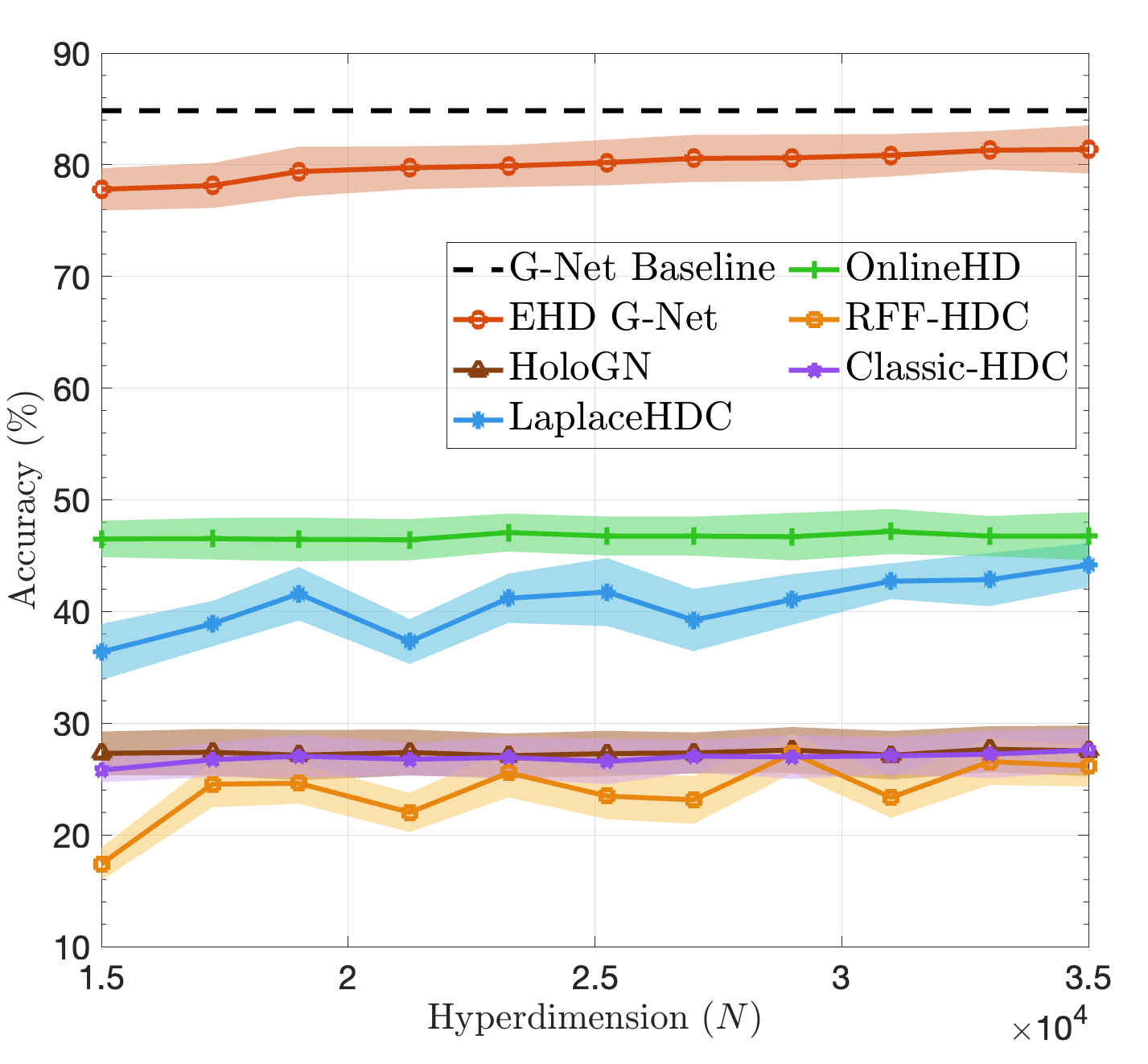}
\put (50,-3) {\scalebox{.7}{\rotatebox{0}{(b)}}}
\end{overpic}
\begin{overpic}[trim={0.17cm -.25cm  0.25cm 0},clip,height=1.725in]{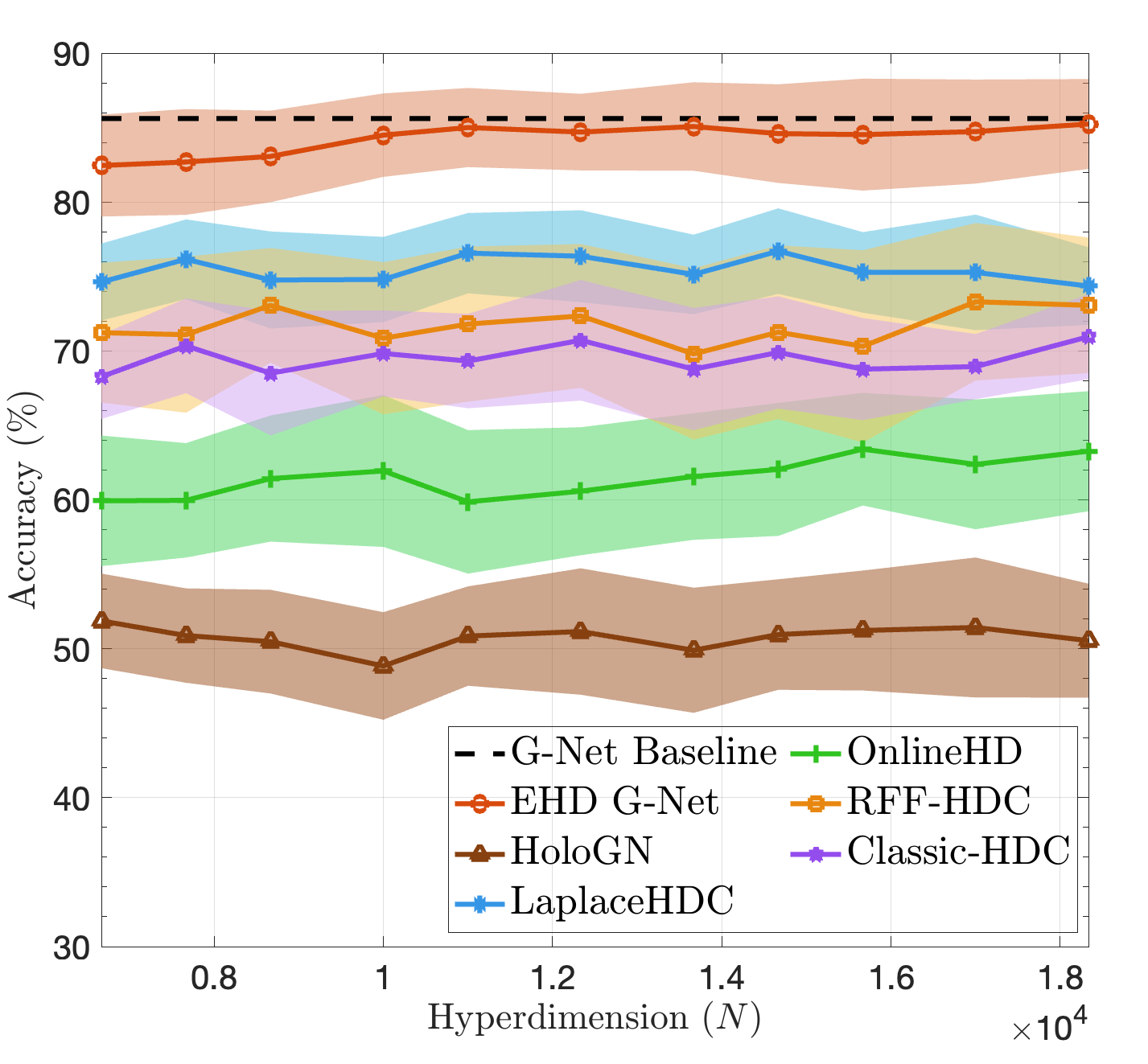}
\put (50,-3) {\scalebox{.7}{\rotatebox{0}{(c)}}}
\end{overpic}\\[.2cm]
\begin{overpic}[trim={0.17cm -.25cm  0.25cm 0},clip,height=1.725in]{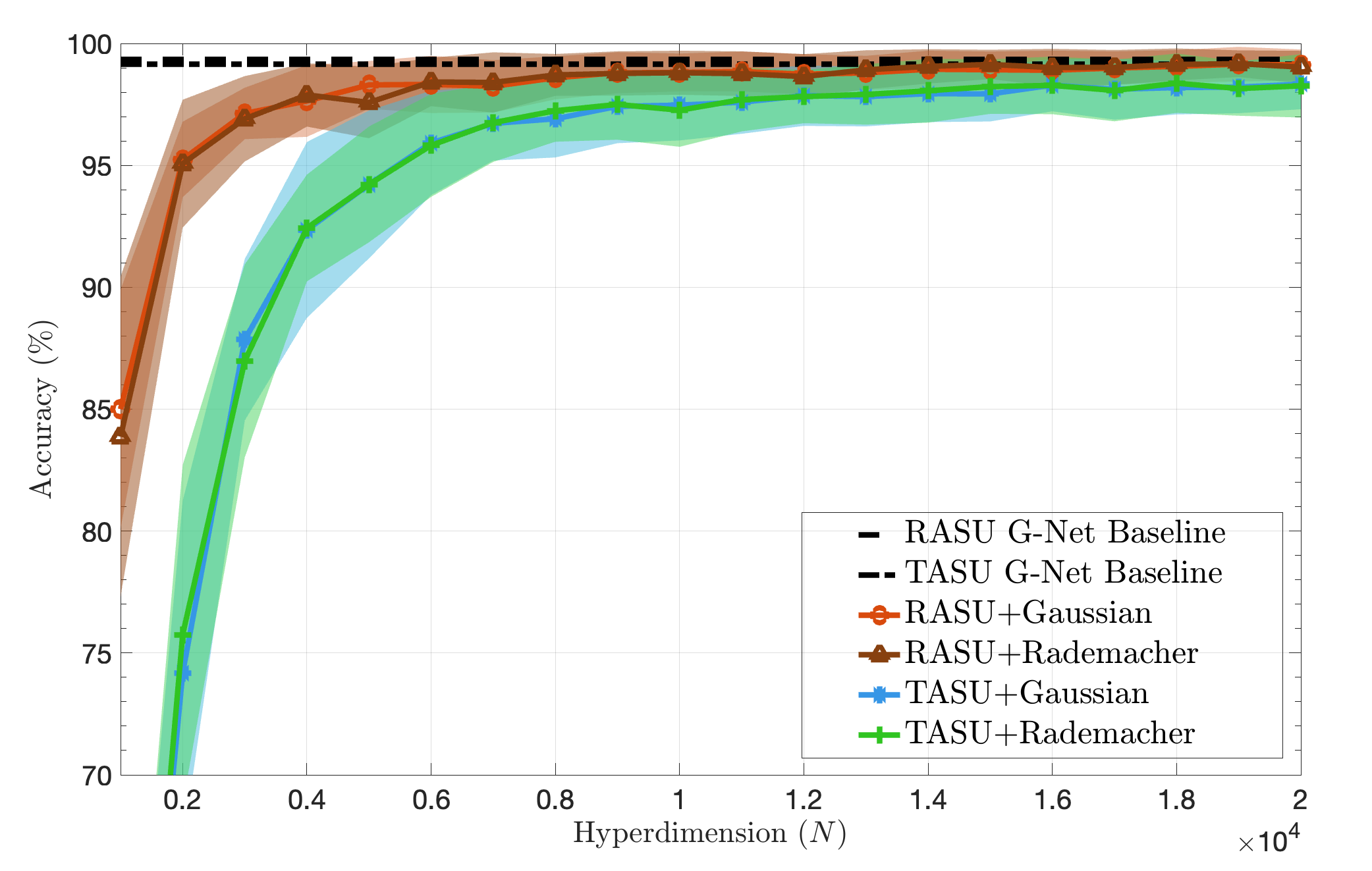}
\put (50,-3) {\scalebox{.7}{\rotatebox{0}{(d)}}}
\end{overpic}
\begin{overpic}[trim={0.17cm -.25cm  0.25cm 0},clip,height=1.725in]{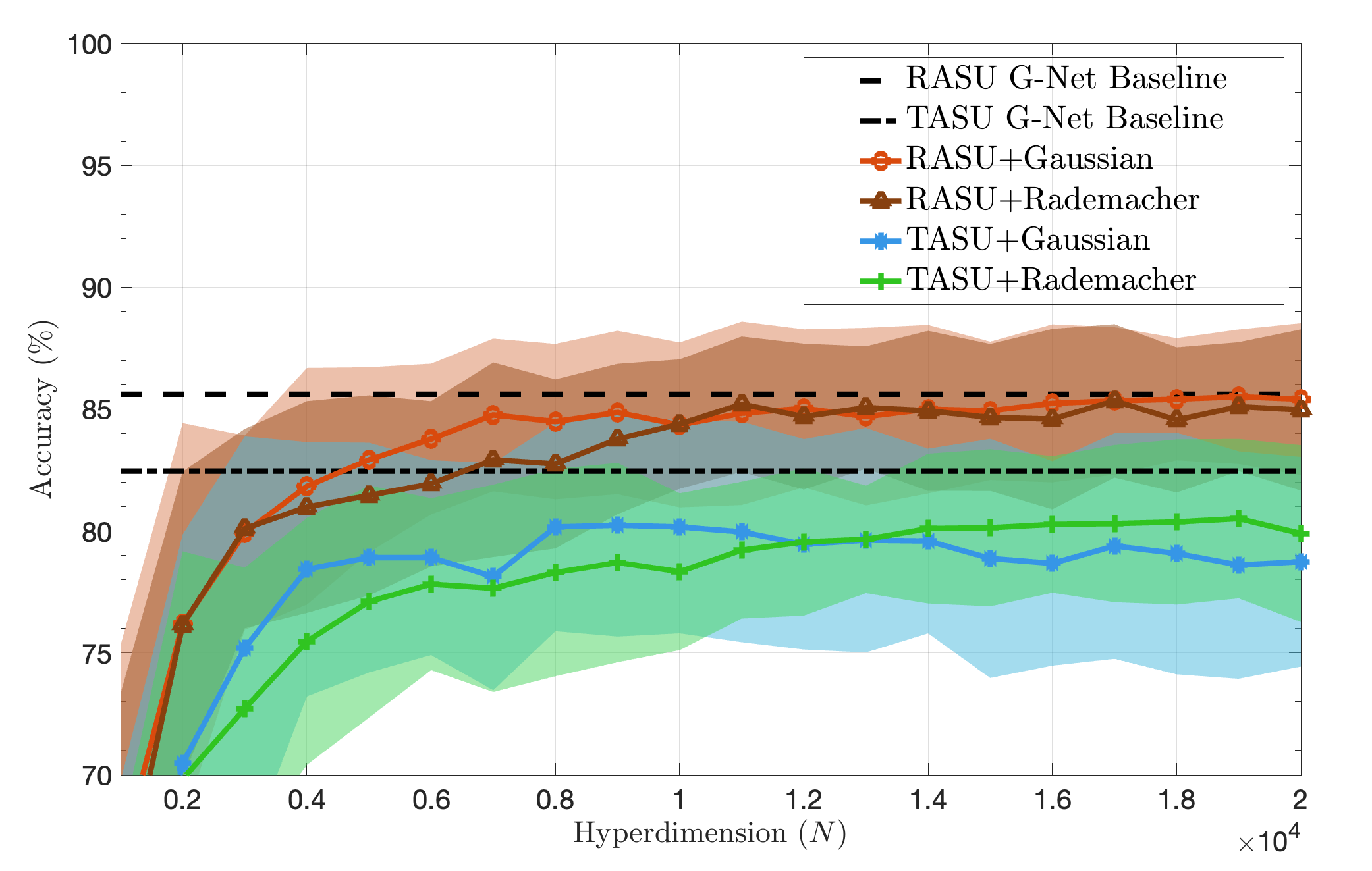}
\put (50,-3) {\scalebox{.7}{\rotatebox{0}{(e)}}}
\end{overpic}
\end{center}\vspace{-.2cm}
\caption{(a–c) Comparison of Rademacher RASU G-Net with other HDC methods (classic HDC \cite{Ge2020Classification}, HoloGN \cite{ExHoloGN}, Laplace HDC \cite{LaplaceHDC}, OnlineHD \cite{onlineHD2021} and RFF-HDC\cite{yu2022understanding}) on MNIST, CIFAR-10, and WSS (left to right); (d,e) performance of different G-Nets on MNIST (left) and WSS (right)}\label{figMCW} \vspace{-.6cm}
\end{figure}

The experiments involve fitting a G-Net to the original training data, followed by evaluation in the binary hyperspace by applying the corresponding EHD G-Net to the random sign embedding of test data. Owing to the G-Net's normalization property, all experiments required at most three convolutional and two fully connected layers, enabling fast and efficient training in the primal space, while still achieving strong baseline accuracies for the G-Net. Panels (a-c) of Figure \ref{figMCW} report the average test accuracies of EHD G-Net and other HDC methods. The reported hyperdimension $N$ represents the average $N_\ell$ across G-Net layers and the dimension used in other methods. For each $N$, the conversion of the reference G-Net to an EHD G-Net was repeated multiple times with different random matrices; the same number of repetitions was applied to other HDC techniques. The plots show the resulting mean accuracies along with $\pm 1$ standard deviation.

The G-Net architecture used in panels (a–c) is a RASU network. Although Gaussian embedding yields slightly better results, we report EHD G-Net accuracies with Rademacher embedding due to its hardware efficiency. One observes that without training a large binary model---and by training only a compact network in the primal space followed by inexpensive binary encoding---we achieve classifiers that outperform state-of-the-art HDC models by a significant margin. For example, HDC accuracies exceed 99\% on MNIST and 81\% on CIFAR-10, rivaling real-valued convolutional neural networks. Accuracy can be even further improved by employing larger G-Nets and higher hyperdimensions, as EHD G-Net performance asymptotically approaches that of the original G-Net with increasing $N$, as evidenced in the plots. Panels (d,e) present G-Net and EHD G-Net accuracies for both RASU and TASU networks, as well as for Gaussian and Rademacher embeddings. While RASU networks tend to yield higher accuracies, TASU remains competitive---for instance, achieving 98.4\% on MNIST. As theory suggests, TASU-based EHD G-Nets converge more slowly to their G-Net baselines. These plots also reinforce the earlier statement regarding the close performance of Gaussian and Rademacher embeddings.
Additional experiments and discussion are provided in Section \ref{sec:NumericalExp} of the Appendix.

\paragraph{Relevance to Other Binary Neural Network Designs} The proposed framework is rooted in hyperdimensional computing: it constructs binary, high-dimensional models without heavy training pipelines. Leveraging random matrix theory, we derive sample-complexity and concentration guarantees for the resulting embeddings. This contrasts with conventional BNN training, which poses a challenging discrete optimization problem and is therefore commonly addressed with heuristics that offer limited theoretical guarantees. G-Net sidesteps this difficulty by learning a continuous, low-dimensional generator whose sign embedding yields the binary model. Furthermore, thanks to the randomized nature of the process, a single trained G-Net can instantiate combinatorially many EHD G-Nets, providing architectural diversity that can be exploited for privacy and security on edge devices. In the extended experiments (Appendix, Section~\ref{sec:NumericalExp}), we also observe that BNNs initialized via an EHD G-Net attain peak accuracy within a few epochs, whereas BNNs trained from scratch typically require substantially many more epochs. 

\paragraph{Computational Cost of EHD G-Net versus a Floating Point G-Net} 
At the hardware level, any floating-point network can be viewed as a ``binary'' system by interpreting each weight as a bit sequence manipulated by floating-point operators. Unlike the EHD G-Net, however, these bit sequences contain significance structure, and the induced operations are substantially more complex. In Appendix Section~\ref{sec:compcost}, we provide a mathematical comparison of computational cost, both time and memory, between EHD G-Net and its floating-point G-Net counterpart. We show that the binary encoding not only reduces computational complexity via a simpler representation, but also confers markedly higher robustness. More specifically, EHD G-Nets are significantly more resilient to bit flips than the corresponding floating-point models where weights are viewed as bit sequences.

In conclusion, this paper introduces a new class of randomized binary neural networks formed by inexpensive sampling from a meta-distribution with controllable mean and variance. As a primary application, the proposed framework demonstrates that hyperdimensional computing can serve as a practical and competitive computational paradigm, achieving performance comparable to conventional deep learning methods, if multi-layer encoding or inference pipelines are considered. Additionally, the paper advances the theoretical foundations of HDC by establishing connections between hyperdimension size and model performance. Moreover, this paper opens several promising directions for future research. One avenue involves fine-tuning and compressing the theoretically grounded models introduced here, including the possibility of directly training EHD G-Nets through easy optimization techniques. Another direction is expanding the theoretical understanding of the Rademacher-based embeddings of the hypersphere to the hypercube we introduced. Additionally, exploring the robustness of the proposed randomized binary neural networks---especially under adversarial conditions---remains a compelling avenue.

In terms of limitations, the proposed framework is primarily motivated by HDC applications where high-dimensional models are typical. Accordingly, EHD G-Nets---derived without training in the binary space---are also high-dimensional, which could be seen as a limitation in certain use cases. Moreover, the network analysis in Section \ref{sec:NetConsistency} focuses on the base ASU activations to streamline the theoretical development. While the cascade analysis of TASU and RASU layers involves many similar steps, a comprehensive treatment requires additional analysis, which is deferred to a more extended presentation.

\section*{Appendix}
\appendix

\startcontents
\printcontents{}{1}{\section*{Contents}}{}
\clearpage 

\section{An Overview of Hyperdimensional Computing}\label{sec:HDCTax}
Hyperdimensional computing (HDC) is a machine learning paradigm inspired by the structure and operation of the brain. In HDC systems, data are represented by high-dimensional vectors, called hypervectors, typically on the order of hundreds or thousands of dimensions. Vector entries may be taken from any number of sets; various models have used real-valued entries \cite{Plate1995}, complex entries \cite{Plate2003}, and binary entries \cite{LaplaceHDC,Gayler1998}. In the latter of these cases, hyperdimensional computing schemes benefit from greatly simplified arithmetic, in contrast to traditional neural networks with float-valued arithmetic. HDC computations are also highly parallelizable and tolerant to stochastic computation environments \cite{yu2022understanding,Verges2023,Menon2021}.

An HDC scheme is principally comprised of two stages: embedding and inference. Consequently, the differences in HDC schemes primarily stem from the differences in their respective embedding and inference methods. In the embedding stage, input data $x$ in some input space $\mathcal X$ is embedded into a high-dimensional vector space (hyperspace) $\mathcal{Y}$ as a hypervector through an embedding map $\phi\colon \mathcal X \to \mathcal Y$. Regardless of the embedding method, the mapping to $\mathcal{Y}$ should ensure that similar input data are represented by similar hypervectors.

For classification tasks with $k$ output labels, the inference stage involves constructing a set of class representatives $\psi_c \in \mathcal{Y}$ for each class $c \in \{1, \ldots, k\}$. Predicting the class of an input $x$ then reduces to comparing its embedding $y$ with all class representatives and selecting the label corresponding to the representation most similar to $y$.
It is worth noting that our proposed framework can be interpreted within the HDC framework, where in the EHD G-Net architecture of Figure \ref{FigGNet:LayNet}(b), the first $L-1$ layers function as a multi-stage embedding pipeline, and the final classification layer operates as the inference module.

\subsection{Hypervector Arithmetic}
In classic HDC, with information embedded as hypervectors, one may use operations on these hypervectors to form composite hypervectors, and measure similarity of hypervectors with a similarity metric. The vector arithmetic operations often used are superposition, binding, and permutation as detailed below.

\textbf{Superposition.} The superposition operation $+\colon \mathcal Y \times \mathcal Y \to \mathcal Y$ combines hypervectors in a way that the similarity of the original operands is preserved \cite{reviewpaper}. In most cases, the superposition operation is taken as typical element-wise vector addition, motivating the choice of notation. 

\textbf{Binding.} The binding operation $\circ\colon \mathcal Y \times \mathcal Y \to \mathcal Y$ associates two hypervectors into a new hypervector, which is dissimilar from both its operands. This operation is often used to combine component hypervectors into composite hypervectors, retaining some information of each its factors. In the special case of binary-valued hypervectors when binding is taken to be elementwise multiplication, the resulting vector arithmetic is particularly amenable to hardware-level implementation \cite{LaplaceHDC}.

\textbf{Permutation.} The permutation operation, as the name suggests, permutes the entries of a hypervector. Denoting by $\Pi$ a permutation matrix, the product $\Pi y$ permutes the entries of $y$ according to $\Pi$.

\textbf{Similarity.}
A similarity measure is a function $\delta\colon \mathcal Y\times \mathcal Y \to \mathbb R$.
The measure is typically used to determine to which class a hypervector belongs, e.g. by selecting 
\[\mathrm{class}(y) = {\arg\max}_{c\in\{1,\ldots,k\}} ~\delta(y, \psi_c).\]
A linear classifier is obtained by selecting $\delta(y, y') = y^\top y'$. Naturally, the choice of embedding method and similarity measure should be amenable in the sense that, $\delta(\phi(x),\phi(x'))$ correlates with the similarity of $x$ and $x'$ in $\mathcal X$.

\subsection{Encoding Methods}
We now briefly review several representative encoding techniques, nearly all of which incorporate some form of randomness.

\textbf{Record-Based Encoding.}
This encoding framework applies when the feature values are discrete or can be closely approximated by discrete levels. This encoding method
employs two types of hypervectors, representing the
feature positions and their level values, respectively.  The  encoding involves the binding
and bundling of random hypervectors associated with the feature positions and level values \cite{imani2019quanthd}. In this framework, 
the position hypervectors are independent and uncorrelated, while the level hypervectors are correlated for neighboring levels. This way correlated vectors in $\mathcal X$ remain correlated in $\mathcal Y$.

\textbf{N-Gram-Based Encoding.} This method also applies to feature vectors with discrete levels. First level hypervectors are randomly generated. Then the encoding happens by binding level hypervectors that are rotationally permuted according to their feature position \cite{Ge2020Classification}.

\textbf{Fractional Power Encoding.}
This method enables the representation of continuous numerical values using complex-valued hypervectors. It begins by sampling a random basis hypervector with complex values (each component corresponds to a random phasor). A value $x$ is encoded by applying fractional self-binding of the basis hypervector $x$ number of times. The similarity kernel of the resulting encoding depends on the distribution generating the basis hypervector \cite{komer2020,Frady2022}. 

\textbf{Random Projection Encoding.} This is also a general encoding scheme which incorporates random projections as the core operation \cite{rachkovskij2015formation}.
For hyperspace $\mathcal Y$ of dimension $N$ and input space $\mathcal X\subseteq\R^p$ this encoding is performed by 
$\phi(x) = \sigma(\mathrnd{G}x)$, where $\mathrnd{G} \in \R^{N \times p}$ has i.i.d. standard normal entries, and $\sigma$ is a quantization operation such as a sign function.  A bias $b\in \mathcal Y$ may optionally be added to the argument of $\sigma$, as in the method described in \cite{Rahimi2007}.

\textbf{Random Fourier Feature Encoding} This method proposed in \cite{yu2022understanding} begins by prescribing a desired symmetric similarity matrix $M\in R^{l\times l}$ among $l$ level hypervectors. Then, binary hypervectors are generated by computing the  singular value decomposition (SVD) of $\sin(\tfrac{\pi}{2}M)$ to obtain left singular values $U$ and diagonal matrix $S_+$, which is identical to the typical $S$ with its negative entries set to 0, and $X\in \mathbb{R}^{n\times D}$ has i.i.d. standard normal entries. The Laplace-HDC model by Pourmand \textit{et al.} \cite{LaplaceHDC} extends this work and shows that choosing $M$ to be a Laplacian kernel is particularly amenable to this framework.

\subsection{Inference Methods}
In this section, we will briefly describe several HDC schemes present in recent literature. We will discuss both broad classes of architectures, as well as some selected realizations of these architectures. We emphasize the following techniques are not mutually exclusive within any particular HDC model. As the array of methods present in HDC models is quite large, we will discuss a relatively small number of selected methods. For a more complete survey of HDC methods, see \cite{reviewpaper}. 

\textbf{Centroid Classifiers.}
Centroid classifiers are the simplest type of HDC classifiers \cite{Kleyko2015}. In these methods, a hypervector is generated for each class by adding the hypervectors of the training data that belong to that class. 

\textbf{Adaptive Training.}
Adaptive learning techniques are able to adapt to changing data, and are suitable for real-time predictions. Essentially, they operate by updating class representatives only when the classifier would incorrectly predict the class of a training datum. The update to the class representative may be done directly proportional to the misclassified hypervector, as in the AdaptHD model \cite{Imani2019a}, or by scaling the misclassified vector by some expression relating to similarity of the misclassified vector with its predicted class, as in OnlineHD \cite{onlineHD2021}.

\textbf{Regenerative Training.}
Regenerative training techniques attempt to improve the selection of class representatives by passing through the data several times. In each pass, a prescribed number of entries of the class representatives carrying the least information (determined by entries with the smallest variance, tracked over the updates) are dropped and re-selected. Notable examples of HDC schemes utilizing this technique are NeuralHD \cite{Zou2021} and DistHD \cite{Wang2023}.

\textbf{Dimensionality Reduction.} A number of HDC models reduce the dimensionality of the hyperspace, thereby increasing the time and memory efficiency of computation. With fewer dimensions in which to embed data, HDC models utilizing dimensionality reduction techniques often incorporate adaptive and regenerative techniques. The SparseHD \cite{Imani2019b} first trains a standard dense centroid classifier, and compares its performance against a sparsified copy of the classifier, with retraining in the case the sparsification reduces the model accuracy by too large a margin. CompHD \cite{Morris2019} partitions hypervectors into $s$ equally-sized components, then binding each component with another hypervector based on its position, and superimposing each component to yield a compressed model. 

\textbf{Optimization-Based Methods.} Many HDC models involve tunable parameters, and a broad class of these methods optimize performance by jointly tuning them. In this sense, a subset of HDC approaches can be regarded as optimization-based. As an example, LeHDC \cite{Duan2022} learns a classifier as a neural network, and binarizes its classification matrix to produce a small binary classifier. 

We emphasize that these models all share the common two-step classification pipeline of embedding and inference. While classification is a prototypical task for HDC, many HDC models perform other machine learning tasks, such as data regression \cite{Hernandez2021b}, reinforcement learning \cite{Chen2022,Ni2022} and multi-task learning \cite{Chang2020,Chang2021}.

\section{An Overview of Binary Neural Networks}\label{sec:BNNTax}
Binary neural networks (BNNs) are a class of neural networks which use binary-valued weights during operation. There is a rich existing literature on BNN architecture, for which we provide here a non-exhaustive review. Extant methods for developing binary neural networks can be broadly divided into several types. In no particular order, first are those that convert an existing pre-trained neural net operating with weights of a non-binary datatype, often float into an equivalent one operating with binary-valued weights. Second are those which train a binary-weighted network from scratch without the need for float-valued weights during any step of training or deployment. Additionally, there are several schemes which hybridize these types, using some combination of float- and binary-valued weights during training. For a more extensive review of BNN methods and architectures, see \cite{YuanChunyu2023Acro, SayedRatshih2023ASLR, ZhaoWenyu2021ARoR}.

A notable subclass of BNN methods is the related XNOR/XOR-Nets, a class of methods which approximate the convolution operations in a network with binary-valued weights, in the former by using a combination of XNOR and bitcount, and in the latter with XOR in place of XNOR \cite{RastegariMohammad2016XICU}, \cite{ ZhuShien2020XAEC}. In either variety, these networks train a float-valued network and update an approximate binary-valued network which is updated in parallel in every training pass \cite{RastegariMohammad2016XICU,LiZewen2022ASoC, ZhuShien2020XAEC}, in contrast with our method of binarizing a G-Net into an EHD G-Net, which occurs as distinct steps in our pipeline. The Bi-Real Net framework proposed by Liu et al. \cite{LiuZechun2018BNEt}, connects the float-valued activations to those of consecutive blocks to increase the representational capacity of the network \cite{LeDuyH.2020IBNw, LiuZechun2020BNBD}. In this sense, layers in a Bi-Real net have a degree of cross-communication which is not present in our model. Another approach at network binarization is the Accurate-Binary-Convolution Network (ABC-Net) which converts a pre-trained CNN into a BNN by determining a basis of binary filters which, together with the appropriate weights, yield approximations of the original layer \cite{lin2017towards}. In contrast, our EHD G-Nets need only a single binary weight tensor in each layer.

\section{Proof of Proposition \ref{G:Isometry}}
\begin{tcolorbox}[colback=teal!20]
    \renewcommand\theproposition{\ref{G:Isometry}}
\begin{proposition}
    For a standard Gaussian matrix $\mathrnd{G}\in\R^{N\times n}$, consider the mapping  $\mathcal{A}(x) = \texttt{sign}(\mathrnd{G}x)$ from $\mathcal{X} = \eS^{n-1}$ (equipped with a geodesic distance $\Dg$) to $\mathcal{Y}=\{-1,1\}^N$ (equipped with a normalized Hamming distance $\Dh$). Fix $\varepsilon\in(0,1)$, then there exist constants $c_1$ and $c_2$ such that if $N\geq c_1n/\varepsilon^2$, with probability exceeding $1-\exp(-c_2 n)$, $\mathcal{A}$ is an $\varepsilon$-near-isometry. That is
    \begin{equation*}
    \forall x,y\in \eS^{n-1}: \qquad \left | \mathcal{D}_{\mathcal{H}}\left( \texttt{sign}\left( \mathrnd{G}x\right), \texttt{sign}\left( \mathrnd{G}y\right)\right) -    \Dg(x,y)  \right| \leq \varepsilon. 
    \end{equation*}
\end{proposition}
\end{tcolorbox}

\begin{proof}
    The proof is a direct application of the following result from \cite{oymak2015near} which provides an optimal dependency on the dimensions and $\varepsilon$:
    \begin{tcolorbox}
\begin{lemma}[Theorem 2.2 in \cite{oymak2015near}] \label{oymak:lem}Suppose $\mathrnd{G}\in\R^{N\times n}$ has independent $\mathcal{N}(0,1)$ entries. For a given a set $K\subseteq\eS^{n-1}$, define the Gaussian width as
\[
\omega(K) = \EE_{g\sim \mathcal{N}(0,I)} \left[ \sup_{x\in K} g^\top x\right],
\]
and denote $\text{cone}(K) = \left\{\alpha v: \alpha\geq 0, v\in K \right\}$. If $\text{cone}(K)$ is a subspace, then for a fixed $\varepsilon\in(0,1)$, there exist constants $c_1$ and $c_2$ such that when $N\geq c_1\frac{\omega^2(K)}{\varepsilon^2}$, with probability exceeding $1-\exp\left( -c_2 \delta^2 N\right)$:
\begin{equation}\label{eq:Oymak}
\forall x,y\in \eS^{n-1}: \qquad \left | \mathcal{D}_{\mathcal{H}}\left( \texttt{sign}\left( \mathrnd{G}x\right), \texttt{sign}\left( \mathrnd{G}y\right)\right) -    \Dg(x,y)  \right| \leq \varepsilon. 
\end{equation}
Here $\text{angle}(x,y)$ denotes the smaller angle between $x$ and $y$.     
\end{lemma} 
\end{tcolorbox}
Based on Lemma \ref{oymak:lem}, to use the result we only need to compute the Gaussian width of the unit sphere:
\begin{align*}
    \omega\left(\eS^{n-1} \right) = \EE_{\mathrnd{g}\sim \mathcal{N}(0,I)} \left[ \sup_{x\in \eS^{n-1}} \mathrnd{g}^\top x\right] &= \EE_{\mathrnd{g}\sim \mathcal{N}(0,I)} \|\mathrnd{g}\|_2 = \frac{\sqrt{2}\Gamma(n/2+1/2)}{\Gamma(n/2)}\leq \sqrt{n}, 
\end{align*}
where we used the fact that $\|g\|_2$ has a chi distribution with $n$ degrees of freedom, and applied Gautschi's inequality. 
\end{proof}
As an alternative presentation of Lemma \ref{oymak:lem}, one may be interested in a version of \eqref{eq:Oymak}, which uses the same metric on both the source and encoded spaces. To that end, notice that 
\begin{align*}
    \mathcal{D}_{\mathcal{H}}\left( \texttt{sign}\left( \mathrnd{G}x\right), \texttt{sign}\left( \mathrnd{G}y\right)\right) &= \frac{1}{4N}\left\| \texttt{sign}\left( \mathrnd{G}x\right)- \texttt{sign}\left( \mathrnd{G}y\right) \right\|^2 \\& = \frac{1}{2} - \frac{1}{2}\left\langle \frac{1}{\sqrt{N}} \texttt{sign}\left( \mathrnd{G}x\right) ,  \frac{1}{\sqrt{N}} \texttt{sign}\left( \mathrnd{G}y\right)\right\rangle \\& = \frac{1}{2} - \frac{1}{2}\cos\left( \text{angle}\left( \frac{1}{\sqrt{N}} \texttt{sign}\left( \mathrnd{G}x\right) ,  \frac{1}{\sqrt{N}} \texttt{sign}\left( \mathrnd{G}y\right)\right) \right) \\&= \sin^2\left( \frac{1}{2}\text{angle}\left( \frac{1}{\sqrt{N}} \texttt{sign}\left( \mathrnd{G}x\right) ,  \frac{1}{\sqrt{N}} \texttt{sign}\left( \mathrnd{G}y\right)\right) \right)\\ & = \sin^2\left( \frac{\pi}{2}\Dg\left( \frac{1}{\sqrt{N}} \texttt{sign}\left( \mathrnd{G}x\right) ,  \frac{1}{\sqrt{N}} \texttt{sign}\left( \mathrnd{G}y\right)\right) \right),
\end{align*}
where $\text{angle}(u,v)$ represents the smaller angle between two vectors $u$ and $v$. Therefore, an alternative presentation of the $\varepsilon$-near isometry in \eqref{eq:Oymak} is
\begin{equation*}
\forall x,y\in \eS^{n-1}: \qquad \left | \sin^2\left( \frac{\pi}{2}\Dg\left( \frac{1}{\sqrt{N}} \texttt{sign}\left( \mathrnd{G}x\right) ,  \frac{1}{\sqrt{N}} \texttt{sign}\left( \mathrnd{G}y\right)\right) \right)  -    \Dg(x,y)  \right| \leq \varepsilon, 
\end{equation*}
where a geodesic distance is used for the source vectors $x,y$ and their encodings $\texttt{sign}\left( \mathrnd{G}x\right)$ and $\texttt{sign}\left( \mathrnd{G}y\right)$. 

\section{Proof of Lemma \ref{grothendieck} (Grothendieck Identity)}
\begin{tcolorbox}[colback=teal!20]
    \renewcommand\thelemma{\ref{grothendieck}}
\begin{lemma}
Let $\mathrnd{g}$ be a standard Gaussian vector in $\mathbb{R}^p$.  Then, for  fixed vectors $u,v \in \mathbb{R}^{p}$:
$$
\mathbb{E}\left[ \sign(\mathrnd{g}^\top {u}) \sign(\mathrnd{g}^\top {v}) \right] = \frac{2}{\pi} \arcsin\left( \frac{{u}^\top {v}}{\|u\|_2\|v\|_2} \right).
$$
\end{lemma}
\end{tcolorbox}
\begin{proof}
We show that for  fixed vectors $u,v \in \eS^{p-1}$:
$$
\mathbb{E}\left[ \sign(\mathrnd{g}^\top {u}) \sign(\mathrnd{g}^\top {v}) \right] = \frac{2}{\pi} \arcsin\left( {u}^\top {v}\right).
$$
Among different ways to show this identity, the geometric argument is the most straightforward. By the rotational invariance of the Gaussian distribution, we can restrict our attention to the two-dimensional subspace spanned by $u$ and $v$. Consider the set $S = \left\{ z : u^\top z \le 0 \text{ and } v^\top z \ge 0 \right.\}$. If $\alpha$ is the angle between $u$ and $v$, then the angle between the two lines that define the boundary of $S$, which are perpendicular to $u$ and $v$, respectively, is also $\alpha$, see Figure \ref{figgeo}.
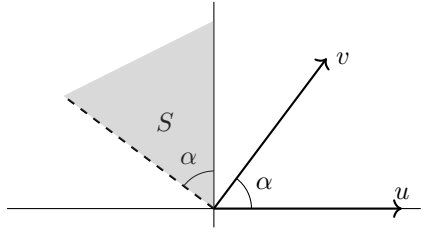
\begin{figure}[h!]
\centering
\begin{tikzpicture}[scale =.5]
  \draw[-] (-5.5,0)--(5.5,0) node[right]{};
  \draw[-] (0,-0.5)--(0,5.5) node[above]{};
  
  \coordinate (Origin) at (0,0);
  \coordinate (U) at (5,0);
  \coordinate (V) at (3,4);
  \coordinate (P) at (-4,3);
  \coordinate (Q) at (4,-3);
  \coordinate (Y) at (0,5);
  \coordinate (mY) at (0,-5);
  \coordinate (A) at (-1.3,2.3);

  \draw[->, black, thick] (Origin)--(U) node[midway, below]{};
  \draw[->, black, thick] (Origin)--(V) node[midway, above]{};
  \draw[-,dashed, black, thick] (Origin)--(P) node[midway, above]{};

  \node at (U) [above] {$u$};
  \node at (V) [right] {$v$};
  \node at (A) {$S$};
    
  \pic [draw, -, "$\alpha$", angle eccentricity=1.5] {angle = U--Origin--V};  
  \pic [draw, -, "$\alpha$", angle eccentricity=1.5] {angle = Y--Origin--P};  

  \fill[gray, opacity=0.3] (0,0) -- (-4,3) -- (0,5) -- cycle;
  
\end{tikzpicture}
\caption{The set $S = \left\{ z : u^\top z \le 0 \text{ and } v^\top z \ge 0 \right\}$}

 \label{figgeo}
\end{figure}

\noindent By  similar reasoning, the set $\left\{ z : u^\top z \ge 0 \text{ and } v^\top z \le 0 \right\}$ is a radial set of angle $\alpha$. Thus,  $\sign(u^\top z)=\sign(v^\top z)$
outside of radial sets of combined angle $2\alpha$. Thanks to the rotation invariance of the Gaussian vectors:
$$
\mathbb{P} \left\{\sign(u^\top \mathrnd{g}) = \sign(v^\top\mathrnd{g})\right\} = \frac{2\pi  - 2\alpha}{2\pi} 
=  \frac{1}{2} + \frac{\arcsin( u^\top v )}{\pi} ,
$$
where we used the fact that
$\alpha = \arccos( u^\top v ) = \frac{\pi}{2} - \arcsin( u^\top v )$. Finally, 
\begin{align*}
    \mathbb{E}\left[ \sign(\mathrnd{g}^\top {u}) \sign(\mathrnd{g}^\top {v}) \right] &= \mathbb{P} \left\{\sign(u^\top \mathrnd{g}) = \sign(v^\top\mathrnd{g})\right\} - \mathbb{P} \left\{\sign(u^\top \mathrnd{g}) \neq \sign(v^\top\mathrnd{g})\right\} \\& = 2\mathbb{P} \left\{\sign(u^\top \mathrnd{g}) = \sign(v^\top\mathrnd{g})\right\} - 1\\ &= \frac{2}{\pi}\arcsin\left( u^\top v\right).
\end{align*}
\end{proof}

\section{Proof of Theorem \ref{RASU:layer}}
\begin{tcolorbox}[colback=teal!20]
    \renewcommand\thetheorem{\ref{RASU:layer}}
\begin{theorem}
Consider a RASU (or ASU) layer with input $x \in \mathbb{S}^{p-1}$ and a weight matrix $W \in \mathbb{R}^{n \times p}$ whose rows are $\ell_2$-normalized. Let the layer output be given by $y = \texttt{RASU}(Wx)$ (or $y = \texttt{ASU}(Wx)$). Define the embedded output as $\tilde{\mathrnd{y}} = \texttt{ReLU}\left(\texttt{sign}(W \mathrnd{G}^\top)\, \texttt{sign}(\mathrnd{G} x)\right)$ (or $\tilde{\mathrnd{y}} =\texttt{Id}\left( \texttt{sign}(W \mathrnd{G}^\top)\, \texttt{sign}(\mathrnd{G} x)\right)$), where $\mathrnd{G} \in \mathbb{R}^{N \times p}$ is a standard Gaussian matrix. Then, for any $c > 0$, with probability at least $1 - \exp(-c)$,
\[
\left\| \frac{1}{N} \tilde{\mathrnd{y}} - y \right\|_2 \leq \sqrt{\frac{2(c + \log 2n )n}{N}}.
\]
\end{theorem}
\end{tcolorbox}

\begin{proof}We present the following general result, applicable to both RASU and TASU layers, with its proof provided at the end of this section. 
\begin{tcolorbox}
\begin{theorem}\label{th:yVect}
   Consider a random matrix $\mathrnd{X}\in{\R^{n\times N}}$ with entries $\mathrnd{x}_{ij}$, such that $a\leq \mathrnd{x}_{i,j}\leq b$ almost surely. Define a vector $\mathrnd{e}\in \R^n$ with entries $\mathrnd{e}_i = f(N^{-1}\sum_{j=1}^N\mathrnd{x}_{ij}) - f(\EE[N^{-1}\sum_{j=1}^N\mathrnd{x}_{ij}])$, where $f$ is a $\rho$-Lipschitz function in $[a,b]$. Then
  \begin{itemize}
      \item[(a)] When $\mathrnd{X}$ maintains independence both along the rows and columns (i.e., independent entries), then for all $c>0$, with probability exceeding $1-\exp(-c)$:
        \[
        \|\mathrnd{e}\|_2\leq \frac{\rho(b-a)\sqrt{n}}{2\sqrt{N}} + \left({\frac{c}{c'}}\right)^{1/4}\frac{\rho(b-a)n^{1/4}}{\sqrt{N}} = \mathcal{O}\left( \rho(b-a)\sqrt{\frac{n}{N}}\right),
        \]
     where $c'$ is a positive constant. 
      \item[(b)] When $\mathrnd{X}$ maintains independence only along the columns, then for all $c>0$, with probability exceeding $1-\exp(-c)$:
\[
\|\mathrnd{e}\|_2\leq \sqrt{\frac{(c+\log 2n)n\rho^2(b-a)^2}{2N}} = \mathcal{O}\left(\rho(b-a)\sqrt{\frac{n\log n}{N}}\right).
\]
  \end{itemize}
\end{theorem}
\end{tcolorbox}
We now proceed by applying this result to a RASU/ASU layer. Recall that 
\[
y = \texttt{RASU}(Wx), \qquad\mbox{and}\qquad  \tilde{\mathrnd{y}} = \texttt{ReLU}\left(\texttt{sign}(W\mathrnd{G}^\top)\texttt{sign}(\mathrnd{G}x)\right).
\]
The discrepancy between the two layers can be written as $\left\| \frac{1}{N}\tilde{\mathrnd{y}} - y \right\|^2 = \sum_{i=1}^n\mathrnd{e_i}^2$, where
\begin{align*}
    \mathrnd{e_i} &= \frac{1}{N}\text{ReLU}\left(\sum_{j=1}^N\sign\left(\mathrnd{g}_j^\top w_i\right)\sign\left(\mathrnd{g_j}^\top x\right)\right) - \text{ReLU}\left( \frac{2}{\pi}\arcsin\left(w_i^\top x\right) 
     \right) \\&= \text{ReLU}\left(\frac{1}{N}\sum_{j=1}^N\sign\left(\mathrnd{g}_j^\top w_i\right)\sign\left(\mathrnd{g_j}^\top x\right)\right) - \text{ReLU}\left(\EE\left[\frac{1}{N}\sum_{j=1}^N\sign\left(\mathrnd{g}_j^\top w_i\right)\sign\left(\mathrnd{g_j}^\top x\right)\right]\right).
\end{align*}
This is clearly an instance of Theorem \ref{th:yVect} part (b), with $\texttt{ReLU}$
as the Lipschitz function and the quantities $ \sign\left(\mathrnd{g}_j^\top w_i\right)\sign\left(\mathrnd{g_j}^\top x\right)$ as $\mathrnd{x}_{ij}$. Since the Lipschitz constant for $\texttt{ReLU}$ and the identity function corresponding to an $\texttt{ASU}$ layer are both 1, and $ -1\leq \sign\left(\mathrnd{g}_j^\top w_i\right)\sign\left(\mathrnd{g_j}^\top x\right)\leq 1$, this immediately implies that with probability exceeding $1-\exp(-c)$:
\[
\left\|\frac{1}{N}\tilde{\mathrnd{y}} - y \right\|_2\leq \sqrt{\frac{2(c+\log 2n)n}{N}} = \mathcal{O}\left(\sqrt{\frac{n\log n}{N}}\right).
\]
\end{proof}
If the construction were modified such that the elements of $\mathrnd{e}_i$ became independent, Theorem~\ref{th:yVect}(a) implies that the discrepancy could be reduced to $\mathcal{O}(\sqrt{n/N})$. However, achieving such independence would require employing a separate embedding matrix for each row of $W$, which increases the memory and computational overhead. The following example suggests that the maximum possible reduction achievable through promoting independence is limited to a factor of $\log n$.

\subsection{Example}
Suppose that $f(x) = \rho x$, and let $\mathrnd{x}_{1j}$ be i.i.d. random variables for $j = 1,\ldots,N$ such that
$
\mathbb{P}[\mathrnd{x}_{1j} = +a] = \mathbb{P}[\mathrnd{x}_{1j} = -a] = 1/2.
$
Assume that $\mathrnd{x}_{ij} = \mathrnd{x}_{1j}$ for $i = 2,\ldots,n$ and $j = 1,\ldots,N$. 
Applying the upper bound on $\|\mathrnd{e}\|_2$ from Theorem \ref{th:yVect} part (b) to this example gives 
$$
\|\mathrnd{e}\|_2 = \mathcal{O}\left( \rho a \sqrt{\frac{n \log n}{N}} \right),
$$
with high probability. 
For comparison, observe that
$$
\mathbb{E} \| \mathrnd{e}\|_2 = \frac{\rho \sqrt{n}}{N} \left|\sum_{j=1}^N x_{1j} \right| =  \Theta\left( \rho a \sqrt{\frac{n}{N}} \right),
$$
as $N \rightarrow \infty$.
Hence, the upper bound is only a factor of $\mathcal{O}(\sqrt{\log n})$ greater than the expected value, at least in some cases.

\subsection{Proof of Theorem \ref{th:yVect}}
\begin{proof}[Proof of part (a)]
We start the proof by using Hoeffding's inequality stated below:
\begin{tcolorbox}
\begin{lemma}[Hoeffding's Inequality]\label{Hoef:Lem}
    Let $\chi_1,\ldots,\chi_N$ be independent random variables such that $a_j\leq \chi_j\leq b_j$ almost surely. Then for all $t\geq 0$:
    \begin{equation}\notag 
        \PP\left\{\left|\sum_{j=1}^N\chi_j - \EE\left[\sum_{j=1}^N\chi_j\right]\right|\geq t\right\} \leq 2\exp\left( -\frac{2t^2}{\sum_{j=1}^N(b_j-a_j)^2}\right).
    \end{equation}
\end{lemma}
\end{tcolorbox}
\noindent Applying this inequality to $\mathrnd{y}_i = N^{-1}\sum_{j=1}^N\mathrnd{x}_{ij}$ reveals that for all $t>0$: 
\begin{equation}\notag 
     \PP\left\{\left(\mathrnd{y}_i - \EE[\mathrnd{y}_i]\right)^2\geq t\right\} \leq 2\exp\left( -\frac{2Nt}{(b-a)^2}\right).
\end{equation}
A direct consequence of the Lipschitz continuity of $f$ is that if $(f(z_1) - f(z_2))^2\geq t$, then $(z_1 - z_2)^2\geq t/\rho^2$, which implies
\begin{align}\notag 
    \PP\left\{\mathrnd{e}_i^2\geq t\right\} =
        \PP\left\{\left(f(\mathrnd{y}_i) - f\left(\EE[\mathrnd{y}_i]\right)\right)^2\geq t\right\} &\leq \PP\left\{\left(\mathrnd{y}_i - \EE[\mathrnd{y}_i]\right)^2\geq \frac{t}{\rho^2}\right\} \\ &\leq 2\exp\left( -\frac{2Nt}{\rho^2(b-a)^2}\right).\notag 
\end{align}
This implies that $\mathrnd{e}_i^2$ is a sub-exponential random variable with sub-exponential norm 
\[
\|\mathrnd{e}_i^2\|_{\psi_1} = \mathcal{O}\left(\rho^2(b-a)^2N^{-1}\right).\]
This also indicates that $\|\mathrnd{e}_i^2-\EE[\mathrnd{e}_i^2]\|_{\psi_1} \leq  c'\rho^2(b-a)^2N^{-1}$ (see \S 2.7 of \cite{vershynin2018high}). Next, we will use Bernstein's inequality for the sum of sub-exponential random variables.
\begin{tcolorbox}
\begin{lemma}[Bernstein's Inequality]\label{Bern:Lem}
    Let $\chi_1,\ldots,\chi_n$ be independent zero-mean sub-exponential random variables. Then for all $t\geq 0$:
    \begin{equation}\notag 
        \PP\left\{\sum_{i=1}^n\chi_i\geq t\right\} \leq \exp\left( -\tilde c\min\left(\frac{t^2}{\sum_{i=1}^n\|\chi_i\|^2_{\psi_1}},\frac{t}{\max_{1\leq i\leq N}\|\chi_i\|_{\psi_1}}\right) \right),
    \end{equation}
    where $\tilde c$ is a positive constant. 
\end{lemma}
\end{tcolorbox}
\noindent Applying Lemma \ref{Bern:Lem} to the sub-exponential random variables $\mathrnd{e}_i^2-\EE[\mathrnd{e}_i^2]$ gives
\begin{align}
\PP\left\{ \frac{1}{n}\sum_{i=1}^n \mathrnd{e}_i^2 -\frac{1}{n}\sum_{i=1}^n\EE\left[\mathrnd{e}_i^2\right]\geq t\right\} &\leq \exp\left( -\tilde c n\min\left(\frac{N^2t^2}{c'^2\rho^4(b-a)^4},\frac{Nt}{c'\rho^2(b-a)^2}\right)\right).\label{berns:ei}
\end{align}
Setting $c_0 = \tilde c/c'^2$, when $n$ is sufficiently large, picking 
\[
t = \sqrt{\frac{c}{c_0 }}\frac{\rho^2(b-a)^2}{\sqrt{n}N}
\]
in \eqref{berns:ei} guarantees that for all $c>0$, with probability exceeding $1-\exp(-c)$:
\[
\|\mathrnd{e}\|_2^2 \leq {\sum_{i=1}^n\EE\left[\mathrnd{e}_i^2\right]} + \sqrt{\frac{c}{c_0}}\frac{\rho^2(b-a)^2\sqrt{n}}{N},
\]
or equivalently
\begin{equation}\label{ei:indep}
\|\mathrnd{e}\|_2 \leq  \sqrt{\sum_{i=1}^n\EE\left[\mathrnd{e}_i^2\right]}+ \left({\frac{c}{c_0}}\right)^{1/4}\frac{\rho(b-a)n^{1/4}}{\sqrt{N}}.
\end{equation}
We now proceed by bounding $\EE[\mathrnd{e}_i^2]$ as follows:
\begin{align}\notag 
    \sum_{i=1}^n\EE[\mathrnd{e}_i^2] &= \sum_{i=1}^n\EE\left[\left(f\left(\frac{1}{N}\sum_{j=1}^N\mathrnd{x}_{ij}\right) - f\left(  
 \frac{1}{N}\sum_{j=1}^N\EE[\mathrnd{x}_{ij}]\right)\right)^2 \right]\\\notag &\leq \rho^2\sum_{i=1}^n\EE\left[\left(\frac{1}{N}\sum_{j=1}^N\mathrnd{x}_{ij} -  
 \frac{1}{N}\sum_{j=1}^N\EE[\mathrnd{x}_{ij}]\right)^2 \right]
 \\ \notag &=\frac{\rho^2}{N^2}\sum_{i=1}^n\sum_{j=1}^N\var [\mathrnd{x}_{ij}]\\&\leq \frac{\rho^2(b-a)^2n}{4N},\label{popov:ineq}
\end{align}
where the first inequality is thanks to the Lipschitz continuity of $f$, and the second inequality is a direct result of the Popoviciu's inequality which states that for a random variable $\mathrnd{x}$, bounded  between $a$ and $b$, $\var[x]\leq (b-a)^2/4$. Combining \eqref{popov:ineq} and \eqref{ei:indep} proves that with probability exceeding $1-\exp(-c)$:
\[
\|\mathrnd{e}\|_2\leq \frac{\rho(b-a)\sqrt{n}}{2\sqrt{N}} + \left({\frac{c}{c_0}}\right)^{1/4}\frac{\rho(b-a)n^{1/4}}{\sqrt{N}} = \mathcal{O}\left( \rho(b-a)\sqrt{\frac{n}{N}}\right).
\]
\end{proof}

\begin{proof}[Proof of part (b)]
Since $\mathrnd{X}$ offers independence along the columns, setting $\mathrnd{y}_i = N^{-1}\sum_{j=1}^N\mathrnd{x}_{ij}$, and appealing to the Hoeffding's inequality in Lemma \ref{Hoef:Lem} yields 
\begin{equation}\notag 
      \forall t>0:\qquad   \PP\left\{\left|\mathrnd{y}_i - \EE[\mathrnd{y}_i]\right|\geq t\right\} \leq 2\exp\left( -\frac{2Nt^2}{(b-a)^2}\right).
\end{equation}
A direct implication of the Lipschitz continuity of $f$ is that if $|f(z_1) - f(z_2)|\geq t$, then $|z_1 - z_2|\geq t/\rho$, which yields 
\begin{align}\notag 
    \PP\left\{\mathrnd{e}_i^2\geq \frac{t^2}{n}\right\} =
        \PP\left\{\left|f(\mathrnd{y}_i) - f\left(\EE[\mathrnd{y}_i]\right)\right|\geq \frac{t}{\sqrt{n}}\right\} &\leq \PP\left\{\left|\mathrnd{y}_i - \EE[\mathrnd{y}_i]\right|\geq \frac{t}{\rho\sqrt{n}}\right\} \\ &\leq 2\exp\left( -\frac{2Nt^2}{n\rho^2(b-a)^2}\right).\label{yi:conc}
\end{align}
For arbitrary real-valued random variables $\chi_1,\ldots\chi_n$ and scalars $t_1,\ldots,t_n$, one has
\begin{align}\notag 
    \PP\left\{ \sum_{i=1}^n \chi_i \geq \sum_{i=1}^n t_i\right\}&\leq \PP\left\{ \bigcup_{i=1}^n\left\{\chi_i\geq t_i\right\}\right\}  \\ &\leq \sum_{i=1}^N \PP\left\{ \chi_i\geq t_i\right\},\label{genIneq}
\end{align}
where the first inequality is valid since 
\[
 \left\{ (\chi_1,\ldots,\chi_n):\sum_{i=1}^n \chi_i \geq \sum_{i=1}^n t_i\right\}\subseteq \bigcup_{i=1}^n\left\{ \chi_i:\chi_i\geq t_i\right\}, 
\]
and the second inequality is thanks to the union bound. An application of \eqref{genIneq} to \eqref{yi:conc} yields 
\begin{align*}
 \PP\left\{\|\mathrnd{e}\|_2\geq t\right\} =  \PP\left\{\|\mathrnd{e}\|_2^2\geq t^2\right\} &\leq 2n\exp\left( -\frac{2Nt^2}{n\rho^2(b-a)^2}\right)\\& = \exp\left( \log (2n) -\frac{2Nt^2}{n\rho^2(b-a)^2}\right),
\end{align*}
which means for all $c>0$, with probability exceeding $1-\exp(-c)$:
\[
\|\mathrnd{e}\|_2\leq \sqrt{\frac{(c+\log(2n))n\rho^2(b-a)^2}{2N}} = \mathcal{O}\left(\rho(b-a)\sqrt{\frac{n\log n}{N}}\right).
\]
\end{proof}

\section{Proof of Theorem \ref{TASU:layer}}
\begin{tcolorbox}[colback=teal!20]
    \renewcommand\thetheorem{\ref{TASU:layer}}
\begin{theorem}
 Consider a TASU layer with output $y = \texttt{TASU}_\kappa( Wx)$, where $x \in \mathbb{S}^{p-1}$ and $W \in \mathbb{R}^{n \times p}$ has normalized rows $w_i^\top$ ($\|w_i\|_2=1$ for $i=1,\ldots,n$). Define the embedded layer output as $\tilde{\mathrnd{y}} = \texttt{sign}\left(\texttt{sign}(W \mathrnd{G}^\top)\, \texttt{sign}(\mathrnd{G} x)\right)$, where $\mathrnd{G} \in \mathbb{R}^{N \times p}$ is a standard Gaussian matrix. Assume 
    \begin{equation*}
        \left|w_i^\top x\right|\geq \ell_{\min}>0, \qquad i=1,\ldots,n.
    \end{equation*}
    Pick a target discrepancy $\varepsilon\leq \sqrt{n}$ and set
$\kappa \geq \frac{\pi}{2\ell_{\min}}\log \frac{4\sqrt{n}}{\varepsilon}.$
Then, for any scalar $c>0$, selecting $N \geq 8(c+\log 2n)n\kappa^2/\varepsilon^2$ guarantees that with probability exceeding $1-3\exp(-c)$: $\|y-\tilde{\mathrnd{y}}\|_2\leq \varepsilon$.
\end{theorem}
\end{tcolorbox}
\begin{proof}Recall that
\[
y = \texttt{TASU}_\kappa( Wx), \qquad \mbox{and}\qquad  \tilde{\mathrnd{y}} = \texttt{sign}\left(\texttt{sign}(W\mathrnd{G}^\top)\texttt{sign}(\mathrnd{G}x)\right). 
\]
Denoting the discrepancy between the two layers as $\mathrnd{e} = \tilde{\mathrnd{y}}- y$, by triangle inequality we have
\[
\|\mathrnd{e}\| = \|\mathrnd{e}'\|_2 + \|\mathrnd{e}''\|_2, 
\]
where 
\begin{align*}
\mathrnd{e}'_i &=\tanh\left( \frac{\kappa }{N}\sum_{j=1}^N\sign\left(\mathrnd{g}_j^\top w_i\right)\sign\left(\mathrnd{g_j}^\top x\right)
     \right) - \tanh\left( \frac{2\kappa}{\pi}\arcsin\left(w_i^\top x\right)\right)\\ &= \tanh\left( \frac{\kappa }{N}\sum_{j=1}^N\sign\left(\mathrnd{g}_j^\top w_i\right)\sign\left(\mathrnd{g_j}^\top x\right)
     \right) - \tanh\left(\frac{\kappa}{N} \EE\left[\sum_{j=1}^N\sign\left(\mathrnd{g}_j^\top w_i\right)\sign\left(\mathrnd{g_j}^\top x\right)\right]\right),
\end{align*}
and 
\[
\mathrnd{e}''_i = \sign\left(\frac{1}{N}\sum_{j=1}^N\sign\left(\mathrnd{g}_j^\top w_i\right)\sign\left(\mathrnd{g_j}^\top x\right)\right) - \tanh\left( \frac{\kappa }{N}\sum_{j=1}^N\sign\left(\mathrnd{g}_j^\top w_i\right)\sign\left(\mathrnd{g_j}^\top x\right)
     \right).
\]
The first term $\|\mathrnd{e}'\|_2$ can be related to Theorem \ref{th:yVect}(b) with $f(x)=\tanh(\kappa x)$ as the Lipschitz function, for which it is easy to verify that $\rho=\kappa$. Therefore, with probability exceeding $1-\exp(-c)$:
\begin{equation}\label{ep:bound}
\|\mathrnd{e}'\|_2\leq \sqrt{\frac{2(c+\log(2n))n\kappa^2}{N}}. 
\end{equation}
For the second term, using the fact 
\[
\left| \sign(z) - \tanh(\kappa z)\right|\leq 2\exp(-2\kappa |z|), 
\]
yields 
\begin{align*}
     |\mathrnd{e}_i''|\leq 2\exp\left( -2\kappa\left|\frac{1 }{N}\sum_{j=1}^N\sign\left(\mathrnd{g}_j^\top w_i\right)\sign\left(\mathrnd{g_j}^\top x\right)\right|
     \right),
\end{align*}
bounding which requires finding a lower bound for $|\frac{1 }{N}\sum_{j=1}^N\sign\left(\mathrnd{g}_j^\top w_i\right)\sign\left(\mathrnd{g_j}^\top x\right)|$. Notice that for general $u$ and $v$, the inequality $|v|-|u|> t$ implies $|u-v|> t$. Therefore, using Hoeffding's inequality 
\begin{align*}
    \PP&\left\{ \frac{2}{\pi}\left| \arcsin(w_i^\top x)\right| - \left|\frac{1 }{N}\sum_{j=1}^N\sign\left(\mathrnd{g}_j^\top w_i\right)\sign\left(\mathrnd{g_j}^\top x\right)\right| \geq t\right\}\\&\leq  \PP\left\{ \left|\frac{1 }{N}\sum_{j=1}^N\sign\left(\mathrnd{g}_j^\top w_i\right)\sign\left(\mathrnd{g_j}^\top x\right) - \frac{2}{\pi} \arcsin(w_i^\top x)\right|\geq t\right\}\leq 2\exp\left( -\frac{Nt^2}{2}\right),
\end{align*}
which implies that with probability exceeding $1-2\exp(-c)$:
\begin{align*}
    \left|\frac{1 }{N}\sum_{j=1}^N\sign\left(\mathrnd{g}_j^\top w_i\right)\sign\left(\mathrnd{g_j}^\top x\right)\right| &> \frac{2}{\pi}\left| \arcsin(w_i^\top x)\right| - \frac{\sqrt{2c}}{\sqrt{N}}\\ &\geq  \frac{2}{\pi}\left| w_i^\top x\right| - \frac{\sqrt{2c}}{\sqrt{N}}\\&\geq \frac{2}{\pi} \ell_{\min}  - \frac{\sqrt{2c}}{\sqrt{N}}.
\end{align*}
This indicates that with probability exceeding $1-2\exp(-c)$:
\begin{align*}
     |\mathrnd{e}_i''|^2\leq 4\exp\left( -\frac{8\kappa}{\pi}\ell_{\min} + \kappa\sqrt{\frac{32c}{N}}
     \right),
\end{align*}
which, after applying a union bound analogous to \eqref{genIneq}, implies that with probability at least $1-2n\exp(-c)$:
\[
\|\mathrnd{e}''\|_2^2\leq 4n\exp\left( -\frac{8\kappa}{\pi}\ell_{\min} + \kappa\sqrt{\frac{32c}{N}}
     \right),
\]
or equivalently, with probability exceeding $1-2\exp(-c)$:
\[
\|\mathrnd{e}''\|_2\leq 2\exp\left(\log \sqrt{n} -\frac{4\kappa}{\pi}\ell_{\min} + \kappa\sqrt{\frac{8(c+\log n)}{N}}
     \right).
\]
After combining this result with \eqref{ep:bound} we can claim that with probability exceeding $1-3\exp(-c)$:
\[
\|\mathrnd{e}\|_2\leq \sqrt{\frac{2(c+\log(2n))n\kappa^2}{N}} + 2\exp\left(\log \sqrt{n} -\frac{4\kappa}{\pi}\ell_{\min} + \kappa\sqrt{\frac{8(c+\log n)}{N}}
     \right).
\]
Setting $N = 8(c+\log 2n)n\kappa^2/\varepsilon^2$ gives
\begin{align}\label{e:2term}
    \|\mathrnd{e}\|_2\leq \frac{\varepsilon}{2} + 2\exp\left( \log \sqrt{n} -\frac{4\kappa}{\pi}\ell_{\min} + \frac{\varepsilon}{\sqrt{n}} \right).
\end{align}
Notice that 
\begin{align*}
   \frac{4\kappa}{\pi}\ell_{\min} \geq 2\log \frac{4\sqrt{n}}{\varepsilon}\geq \log \sqrt{n}  + \frac{\varepsilon}{\sqrt{n}} - \log \frac{\varepsilon}{4},
\end{align*}
where the first inequality is thanks to the assumption on $\kappa$ and the second inequality is valid as long as $n\geq \varepsilon^2$. Together with \eqref{ep:bound} and \eqref{e:2term}, this implies that with probability exceeding $1-3\exp(-c)$:
\begin{align*}
    \|\mathrnd{e}\|_2\leq \frac{\varepsilon}{2} + 2\exp\left( \log \frac{\varepsilon}{4}\right) = \varepsilon. 
\end{align*}
\end{proof}

\section{Proof of Theorem \ref{th:netConsistency}}
\begin{tcolorbox}[colback=teal!20]
    \renewcommand\thetheorem{\ref{th:netConsistency}}
\begin{theorem} Consider the cascade of $L$ G-Net layers and the corresponding hyperdimensional embedding as \eqref{eq:seqNet}. Assume that each layer $\ell$ of the network is consistent with near isometry of $\texttt{ASU}$ in $\eS^{n_{\ell-1}-1}$ with parameters $\beta_\ell$ and $\varepsilon_\ell$, as stated in \eqref{near:isomx-y}. Then with probability exceeding $1-L\exp(-c)$:
\begin{equation*}
  \|\delta_L\|_2 \leq   c' \sum_{\ell=1}^L \left(\sqrt{\frac{(c+\log n_\ell)n_\ell}{N_\ell\beta_\ell\left(1-\varepsilon_\ell\right)}} + \sqrt{\frac{\varepsilon_\ell}{1+\varepsilon_\ell}}\right),
\end{equation*}
where $c'$ is an absolute constant. \vspace{-.2cm}
\end{theorem}
\end{tcolorbox}
\subsection{Auxiliary Lemmas} To prove Theorem \ref{th:netConsistency}, we first need to state some auxiliary lemmas, which follow.  
\begin{tcolorbox}
\begin{lemma} \label{lem:isomxy}   If \eqref{near:isomx-y} holds then then for all $x,y\in\mathcal{D}$:
\begin{equation}\label{near:isomxy}
     x^\top y - \frac{3}{2}\varepsilon\leq \beta^{-1} \left\langle \texttt{ASU}\left(Wx\right),\texttt{ASU}\left(Wy\right) \right\rangle \leq  x^\top y + \frac{3}{2}\varepsilon,
\end{equation}
\end{lemma}
\end{tcolorbox}
\begin{proof}
By the left-hand side of \eqref{near:isomx-y} we have
\begin{align*}
    \|x\|^2 + \|y\|^2 - 2x^\top y - \varepsilon &\leq \beta^{-1}\left\| \texttt{ASU}\left(Wx\right)\right\|^2 + \beta^{-1}\left\| \texttt{ASU}\left(Wy\right)\right\|^2 - 2\beta^{-1}\left\langle \texttt{ASU}\left(Wx\right),\texttt{ASU}\left(Wy\right) \right\rangle\\&\leq \|x\|^2 + \varepsilon + \|y\|^2 + \varepsilon  - 2\beta^{-1}\left\langle \texttt{ASU}\left(Wx\right),\texttt{ASU}\left(Wy\right) \right\rangle,
\end{align*}
where in the second inequality we used $\beta^{-1}\|\texttt{ASU}(Wu)\|_2^2\leq \|u\|_2^2+\varepsilon$ twice, once by setting $u$ to $x$ and once to $y$. This implies that 
\[
\beta^{-1} \left\langle \texttt{ASU}\left(Wx\right),\texttt{ASU}\left(Wy\right) \right\rangle \leq  x^\top y + \frac{3}{2}\varepsilon.
\]
Also by the second side of the inequality we have
\begin{align*}
    \|x\|^2 + \|y\|^2 - 2x^\top y + \varepsilon &\geq \beta^{-1}\left\| \texttt{ASU}\left(Wx\right)\right\|^2 + \beta^{-1}\left\| \texttt{ASU}\left(Wy\right)\right\|^2 - 2\beta^{-1}\left\langle \texttt{ASU}\left(Wx\right),\texttt{ASU}\left(Wy\right) \right\rangle\\&\geq \|x\|^2 - \varepsilon + \|y\|^2 - \varepsilon  - 2\beta^{-1}\left\langle \texttt{ASU}\left(Wx\right),\texttt{ASU}\left(Wy\right) \right\rangle,
\end{align*}
which implies that 
\[
\beta^{-1} \left\langle \texttt{ASU}\left(Wx\right),\texttt{ASU}\left(Wy\right) \right\rangle \geq  x^\top y - \frac{3}{2}\varepsilon.
\]

\end{proof}
\begin{tcolorbox}
    \begin{lemma}\label{lema:normdif}
    Consider two arbitrary non-zero vectors $y$ and $\tilde y\in\R^n$. If $\|\tilde y-y\|_2\leq \epsilon$ and $\|y\|_2\geq \kappa>0$, then
    \[
    \left\| \frac{\tilde y}{\|\tilde y\|_2} - \frac{y}{\|y\|_2}\right\|_2\leq \frac{2\epsilon}{\kappa}.
    \]
\end{lemma}
\end{tcolorbox}

\begin{proof}
A direct implication of the assumptions is
\begin{equation}\label{eqlem1}
    \left\| \frac{\tilde y}{\|y\|_2}- \frac{y}{\|y\|_2}\right\|_2\leq \frac{\epsilon}{\kappa}.
\end{equation}
    On the other hand
\begin{equation}\label{eqlem2}
    \left\| \frac{\tilde y}{\|\tilde y\|_2}- \frac{\tilde y}{\|y\|_2}\right\|_2= \left| \frac{1}{\|\tilde y\|_2}- \frac{1}{\|y\|_2} \right| \|\tilde y\|_2 = \frac{\left|\|y\|_2 - \|\tilde y\|_2 \right|}{\|y\|_2}\leq \frac{\left\|y - \tilde y\right \|_2 }{\|y\|_2}\leq \frac{\epsilon}{\kappa}, 
\end{equation}    
where we used the reverse triangle inequality. Now using the triangle inequality along with \eqref{eqlem1} and \eqref{eqlem2} we get
\begin{align*}
    \left\| \frac{\tilde y}{\|\tilde y\|_2} - \frac{y}{\|y\|_2}\right\|_2&\leq  \left\| \frac{\tilde y}{\|\tilde y\|_2}- \frac{\tilde y}{\|y\|_2}\right\|_2 +  \left\| \frac{\tilde y}{\|y\|_2}- \frac{y}{\|y\|_2}\right\|_2\leq \frac{2\epsilon}{\kappa}, 
\end{align*}
which completes the proof. 
\end{proof}

\begin{tcolorbox}
\begin{lemma} \label{lemma:normalized:isometry}Consider $W\in\R^{n\times p}$ which is consistent with near isometry of $\texttt{ASU}$ in $\eS^{p-1}$, as stated in \eqref{near:isomx-y}. Then for all $x,y\in\eS^{p-1}$: 
    \begin{equation}\label{near:isom:normalized}
    \left\| \frac{\texttt{ASU}\left(Wx\right)}{\left\|\texttt{ASU}\left(Wx\right)\right\|}-\frac{\texttt{ASU}\left(Wy\right)}{\left\|\texttt{ASU}\left(Wy\right)\right\|} \right\|_2^2 \leq  \left\|x-y\right\|_2^2 + \frac{5\varepsilon}{1+\varepsilon}.
\end{equation}
\end{lemma}

\end{tcolorbox}
\begin{proof}
    Expanding the left-hand side of \eqref{near:isom:normalized} gives
    \begin{align*}
         \left\| \frac{\texttt{ASU}\left(Wx\right)}{\left\|\texttt{ASU}\left(Wx\right)\right\|}-\frac{\texttt{ASU}\left(Wy\right)}{\left\|\texttt{ASU}\left(Wy\right)\right\|} \right\|_2^2 &= \left\| \frac{\texttt{ASU}\left(Wx\right)}{\left\|\texttt{ASU}\left(Wx\right)\right\|}\right\|_2^2 + \left\| \frac{\texttt{ASU}\left(Wy\right)}{\left\|\texttt{ASU}\left(Wy\right)\right\|}\right\|_2^2-\frac{2\left\langle \texttt{ASU}\left(Wx\right),\texttt{ASU}\left(Wy\right) \right\rangle}{\left\|\texttt{ASU}\left(Wx\right)\right\|_2\left\|\texttt{ASU}\left(Wy\right)\right\|_2}\\
         &=2  -\frac{2\beta^{-1}\left\langle \texttt{ASU}\left(Wx\right),\texttt{ASU}\left(Wy\right) \right\rangle}{\beta^{-1/2}\left\|\texttt{ASU}\left(Wx\right)\right\|_2\beta^{-1/2}\left\|\texttt{ASU}\left(Wy\right)\right\|_2}
         \\
         &\leq 2 - 2\frac{x^\top y-\frac{3}{2}\varepsilon}{1+\varepsilon}\\
         & = \|x-y\|_2^2 +  \frac{\varepsilon}{1+\varepsilon}\left( 3+ 2x^\top y\right)\\
         &\leq  \|x-y\|_2^2 +\frac{5\varepsilon}{1+\varepsilon},
    \end{align*}
where the first inequality is thanks to Lemma \ref{lem:isomxy}. This completes the proof.     
\end{proof}

\subsection{Proof of the Main Theorem}
\begin{proof}
We are now ready to prove Theorem \ref{th:netConsistency}. By the triangle inequality we have
\begin{align}\label{deltal:bound}
    \left\|\delta_\ell\right\|_2\leq \left\| \frac{\tilde{\mathrnd{y}}_\ell}{\|\tilde{\mathrnd{y}}_\ell\|_2} - \frac{\texttt{RASU}\left(W_\ell\frac{\tilde{\mathrnd{y}}_{\ell-1}}{\|\tilde{\mathrnd{y}}_{\ell-1}\|}\right)}{\left\| \texttt{RASU}\left(W_\ell\frac{\tilde{\mathrnd{y}}_{\ell-1}}{\|\tilde{\mathrnd{y}}_{\ell-1}\|}\right) \right\|_2} \right\|_2 + \left\| \frac{\texttt{RASU}\left(W_\ell\frac{\tilde{\mathrnd{y}}_{\ell-1}}{\|\tilde{\mathrnd{y}}_{\ell-1}\|}\right)}{\left\| \texttt{RASU}\left(W_\ell\frac{\tilde{\mathrnd{y}}_{\ell-1}}{\|\tilde{\mathrnd{y}}_{\ell-1}\|}\right) \right\|_2} - \frac{y_\ell}{\|y_\ell\|_2} \right\|_2.
\end{align}
We start bounding $\|\delta_\ell\|_2$ by first bounding the first term on the right-hand side of \eqref{deltal:bound}. By the layer consistency we know that given $\tilde{\mathrnd{y}}_{\ell-1}$, with probability exceeding $1-p_\ell$ (where as shown in Theorem \ref{RASU:layer}, $p_\ell=\exp(-c)$ for some desired $c>0$):
\begin{align}\label{eq:disc:layerl}
    \left\| \frac{1}{N_\ell}\tilde{\mathrnd{y}}_\ell - \texttt{RASU}\left(W_\ell\frac{\tilde{\mathrnd{y}}_{\ell-1}}{\|\tilde{\mathrnd{y}}_{\ell-1}\|}\right) \right\|_2 = \mathcal{O}\left(\sqrt{\frac{(c+\log n_\ell)n_\ell}{N_\ell}}\right).
\end{align}
By the near isometry consistency of the $\ell$-th layer we have
\begin{equation}\label{eq:norm:layerl}
\left\|\texttt{RASU}\left(W_\ell\frac{\tilde{\mathrnd{y}}_{\ell-1}}{\|\tilde{\mathrnd{y}}_{\ell-1}\|}\right)\right\|_2 \geq \sqrt{\beta_\ell\left(1-\varepsilon_\ell\right)}.
\end{equation}
Using Lemma \ref{lema:normdif}, the bounds \eqref{eq:disc:layerl} and \eqref{eq:norm:layerl} imply that given $\tilde{\mathrnd{y}}_{\ell-1}$, with  probability exceeding $1-p_\ell$:
\begin{equation}\label{term1:bound}
    \left\| \frac{\tilde{\mathrnd{y}}_\ell}{\|\tilde{\mathrnd{y}}_\ell\|_2} - \frac{\texttt{RASU}\left(W_\ell\frac{\tilde{\mathrnd{y}}_{\ell-1}}{\|\tilde{\mathrnd{y}}_{\ell-1}\|}\right)}{\left\| \texttt{RASU}\left(W_\ell\frac{\tilde{\mathrnd{y}}_{\ell-1}}{\|\tilde{\mathrnd{y}}_{\ell-1}\|}\right) \right\|_2} \right\|_2= \mathcal{O}\left(\sqrt{\frac{(c+\log n_\ell)n_\ell}{N_\ell\beta_\ell\left(1-\varepsilon_\ell\right)}}\right). 
\end{equation}
The second term on the right-hand side of \eqref{deltal:bound} can be bounded using Lemma \ref{lemma:normalized:isometry} as 
\begin{align}\notag 
    \left\| \frac{\texttt{RASU}\left(W_\ell\frac{\tilde{\mathrnd{y}}_{\ell-1}}{\|\tilde{\mathrnd{y}}_{\ell-1}\|}\right)}{\left\| \texttt{RASU}\left(W_\ell\frac{\tilde{\mathrnd{y}}_{\ell-1}}{\|\tilde{\mathrnd{y}}_{\ell-1}\|}\right) \right\|_2} - \frac{y_\ell}{\|y_\ell\|_2} \right\|_2^2 &=  \left\| \frac{\texttt{RASU}\left(W_\ell\frac{\tilde{\mathrnd{y}}_{\ell-1}}{\|\tilde{\mathrnd{y}}_{\ell-1}\|}\right)}{\left\| \texttt{RASU}\left(W_\ell\frac{\tilde{\mathrnd{y}}_{\ell-1}}{\|\tilde{\mathrnd{y}}_{\ell-1}\|}\right) \right\|_2} - \frac{\texttt{RASU}\left(W_\ell\frac{ y_{\ell-1}}{\| y_{\ell-1}\|}\right)}{\left\| \texttt{RASU}\left(W_\ell\frac{ y_{\ell-1}}{\| y_{\ell-1}\|}\right) \right\|_2}\right\|_2^2\\&\leq \left\| \frac{\tilde{\mathrnd{y}}_{\ell-1}}{\|\tilde{\mathrnd{y}}_{\ell-1}\|_2} - \frac{y_{\ell-1}}{\|y_{\ell-1}\|_2} \right\|_2^2 + \frac{5\varepsilon_\ell}{1+\varepsilon_\ell}\notag \\&=\|\delta_{\ell-1}\|_2^2 + \frac{5\varepsilon_\ell}{1+\varepsilon_\ell}\label{term2:bound}.
\end{align}
Using \eqref{term2:bound} and \eqref{term1:bound} in \eqref{deltal:bound} we conclude that given $\tilde{\mathrnd{y}}_{\ell-1}$, with probability exceeding $1- p_\ell$, 
\begin{equation}\label{recursive:deltabound}
  \|\delta_\ell\|_2 \leq \|\delta_{\ell-1}\|_2 +  c'\left(  \sqrt{\frac{(c+\log n_\ell)n_\ell}{N_\ell\beta_\ell\left(1-\varepsilon_\ell\right)}} + \sqrt{\frac{\varepsilon_\ell}{1+\varepsilon_\ell}}\right),\qquad \ell =1,\ldots,L,
\end{equation}
where $c'$ is an absolute  constant and $\delta_0=0$. 
Now consider $A_\ell$ to be the event that \eqref{eq:disc:layerl} holds. Then we have
\begin{align*}
    \PP\left\{A_L\right\}&\geq \PP\left\{A_{L-1}\right\} \PP\left\{A_L|A_{L-1}\right\}\\&\geq \PP\left\{A_{L-2}\right\} \PP\left\{A_L|A_{L-1}\right\}\PP\left\{A_{L-1}|A_{L-2}\right\}\\&\vdots\\&\geq \PP\left\{A_{1}\right\} \prod_{\ell=2}^L\PP\left\{A_L|A_{L-1}\right\}\\&=\prod_{\ell=1}^L(1-p_\ell).
\end{align*}
By the Weierstrass inequality we know if $p_\ell\in[0,1]$ for $\ell=1,\ldots,L$, then
\[
\prod_{\ell=1}^L(1-p_\ell)\geq 1- \sum_{\ell=1}^L p_\ell.
\]
Together with \eqref{recursive:deltabound}, this implies that with probability exceeding $1-L\exp(-c)$:
\begin{equation}\label{recursive:deltaTot}
  \|\delta_L\|_2 \leq   c' \sum_{\ell=1}^L \left(\sqrt{\frac{(c+\log n_\ell)n_\ell}{N_\ell\beta_\ell\left(1-\varepsilon_\ell\right)}} + \sqrt{\frac{\varepsilon_\ell}{1+\varepsilon_\ell}}\right),
\end{equation}
where $c'$ is an absolute constant. 
\end{proof}

\section{Proof of Theorem \ref{near:ISO:Th}}
\begin{tcolorbox}[colback=teal!20]
    \renewcommand\thetheorem{\ref{near:ISO:Th}}
\begin{theorem}
        Consider $\mathrnd{W}\in\R^{n\times p}$, where $n\gtrsim p\geq 27$, following the construction format in \eqref{eq:W}. Let $\mathrnd{g}_i\sim\mathcal{N}(0,I_p)$ be independent standard normal vectors. Then, for all $x, y\in \mathbb{B}_2^p$, with probability exceeding $1-c\exp(-c'p)$:
\begin{equation*}
    \left\|x-y \right\|_2^2 - \varepsilon_{n,p}^{l}\leq \beta_{n,p}^{-1}\left\|\texttt{ASU}(\mathrnd{W} x) - \texttt{ASU}(\mathrnd{W} y)\right\|^2  \leq \left\|x-y \right\|_2^2 + \varepsilon_{n,p}^{u},
\end{equation*}
where 
\begin{equation*}
\beta_{n,p}^{-1} = \frac{\pi^2 p}{4\left(\sqrt{n}+\sqrt{p}\right)^2}, ~~~ \varepsilon_{n,p}^l = c_l\frac{\sqrt{p}}{ \sqrt{n}+\sqrt{p}}, ~~~  \varepsilon_{n,p}^u = c_u\left( \frac{\sqrt{p}}{ \sqrt{n}+\sqrt{p}} + \frac{n+p^2}{p\left( \sqrt{n}+\sqrt{p}\right)^2}\right),
\end{equation*}
and $c, c', c_l$ and $c_u$ are absolute numerical constants. 
\end{theorem}
\end{tcolorbox}
\begin{proof}
    We present the proof as four main steps and for each step provide the related lemmas which are all proved at the end of this section.\\ 

\paragraph{\textbf{-- Step 1. Bounding the $\texttt{ASU}$ Variations:}}
    Using basic calculus, one could establish the following inequality:
\begin{tcolorbox}
\begin{lemma}\label{rasu:bound}Consider $x,y\in[-1,1]$. Then
\begin{align*}%
     \frac{4}{\pi^2}(x-y)^2 \leq \left(\texttt{ASU}(x) - \texttt{ASU}(y)\right)^2 &\leq \frac{4}{\pi^2}(x-y)^2 + 2\left(1-\frac{4}{\pi^2}\right)\left(x^4+y^4\right)\\ &\leq \frac{4}{\pi^2}(x-y)^2 + \frac{6}{5}\left(x^4+y^4\right).
\end{align*}
    \end{lemma}
\end{tcolorbox} 
\begin{proof}
    See Section \ref{flfu:sec}. 
\end{proof}
By Lemma \ref{rasu:bound}, for each row of $\mathrnd{W}$ we have
    \begin{align*}%
     \frac{4}{\pi^2}\left(\mathrnd{w}_i^\top x-\mathrnd{w}_i^\top y\right)^2 \leq \left(\texttt{ASU}(\mathrnd{w}_i^\top x) - \texttt{ASU}(\mathrnd{w}_i^\top y)\right)^2 &\leq \frac{4}{\pi^2}(\mathrnd{w}_i^\top x-\mathrnd{w}_i^\top y)^2 + \frac{6}{5}\left(\left(\mathrnd{w}_i^\top x\right)^4+\left(\mathrnd{w}_i^\top y\right)^4\right).
\end{align*}
Summing along the rows and scaling both sides gives
\begin{equation}\label{ASU:doubleBound}
    0\leq \frac{\pi^2}{4}\left\|\texttt{ASU}(\mathrnd{W} x) - \texttt{ASU}(\mathrnd{W} y)\right\|^2 - \left\|\mathrnd{W}(x-y) \right\|^2\leq \frac{3\pi^2}{10}\left( \left\| \mathrnd{W}x\right\|_4^4 + \left\| \mathrnd{W}y\right\|_4^4\right).
\end{equation}
Consider $s_{\min}(\mathrnd{W})$ and $s_{\max}(\mathrnd{W})$ to denote the smallest and largest singular values of $\mathrnd{W}$. Then for any $x$ and $y$:
\begin{equation}\label{Singular:bounds}
  s_{\min}^2(\mathrnd{W})\|x-y\|^2_2\leq  \left\|\mathrnd{W}(x-y) \right\|^2 \leq s_{\max}^2(\mathrnd{W})\|x-y\|^2_2.
\end{equation}
On the other hand, using the notion of induced norm, for any $x\in \mathbb{B}_2^p$ one has
\begin{equation}\label{ind:norm}
\|\mathrnd{W}x\|_4^4 \leq \|\mathrnd{W}\|_{2\to 4}^4,
\end{equation}
where 
\[
\|\mathrnd{W}\|_{2\to 4} = \sup_{x:\|x\|_2\leq 1}{\|\mathrnd{W}x\|_4}.
\]
Using \eqref{ind:norm} and \eqref{Singular:bounds} in \eqref{ASU:doubleBound} we get the following general bound for all $x, y\in \mathbb{B}_2^p$:
\begin{equation}\label{ASU:doubleBound2}
    s_{\min}^2(\mathrnd{W})\|x-y\|^2_2\leq \frac{\pi^2}{4}\left\|\texttt{ASU}(\mathrnd{W} x) - \texttt{ASU}(\mathrnd{W} y)\right\|^2 \leq  s^2_{\max}(\mathrnd{W})\|x-y\|^2_2 + \frac{6\pi^2}{10} \|\mathrnd{W}\|_{2\to 4}^4.
\end{equation}
The next step is using high probability bounds for  $s_{\min}(\mathrnd{W}), s_{\max}(\mathrnd{W})$ and $\|\mathrnd{W}\|_{2\to 4}$ to express them in terms of the problem dimensions $n$ and $p$. \\

\paragraph{\textbf{-- Step 2. High Probability Bounds for  $s_{\min}(\mathrnd{W})$ and $s_{\max}(\mathrnd{W})$}:} In this step we will focus on bounding the singular values of $\mathrnd{W}$ in \eqref{eq:W}, where $\mathrnd{g}_i\sim\mathcal{N}(0,I_p)$ are independent standard normal vectors. A central component of the proofs of this step and the next step is the following lemma, which characterizes the moment generating functions of rows of $\mathrnd{W}$. Since parts (a) and (b) are closely related, they are consolidated as one result, while only part (a) is used in this step. 
\begin{tcolorbox}
    \begin{lemma}\label{tale:ineq}
        Consider $x\in\eS^{p-1}$ where $p\geq 27$, and $\mathrnd{g} = [\mathrnd{g}_1,\ldots,\mathrnd{g}_p]^\top\sim\mathcal{N}(0,I)$. Define the random variable 
\[
\mathrnd{u} = \frac{\left| x^\top \mathrnd{g}\right|}{\|\mathrnd{g}\|}.
\]
Then
\begin{itemize}
    \item[(a)] For all $\tau\in\left [0, 2/5\right ]$:
\[
 \EE \exp\left(\tau^2 p \mathrnd{u}^2\right) \leq \exp\left( \frac{25}{4}\tau^2  \right).
\]
    \item[(b)] For all $\tau\in\left [0, \log(2)/6\right ]$:
\begin{align}\notag 
     \EE \exp\left(\tau p \mathrnd{u}^4\right)\leq \exp\left(\frac{13\tau}{p} \right).
\end{align}    
\end{itemize}

    \end{lemma}
\end{tcolorbox}
\begin{proof}
    See Section \ref{MGF:proof}. 
\end{proof}
As a well-known result, a random variable $\mathrnd{s}$ is sub-Gaussian if there exists a constant $k>0$ such that for all $\tau$ satisfying $|\tau|\leq k^{-1}$, one has $\EE \exp\left(\tau^2 \mathrnd{s}\right) \leq \exp\left( k^2\tau^2 \right)$ (e.g., see Proposition 2.5.2 of \cite{vershynin2018high}).
Defining $\mathrnd{u} = \left| x^\top \mathrnd{g}\right|/\|\mathrnd{g}\|$, it is immediate from part (a) of Lemma \ref{tale:ineq}  that $\sqrt{p}\mathrnd{u}$ is a sub-Gaussian random variable. Furthermore, since Lemma \ref{tale:ineq} considers $x\in\eS^{p-1}$ as any arbitrary unit vector on the sphere, this guarantees that the normalized vector $\mathrnd{g}/\|\mathrnd{g}\|$ is generally a sub-Gaussian random vector. Notably, $\mathrnd{g}/\|\mathrnd{g}\|$ follows a uniform distribution on the unit sphere, which is guaranteed to follow a sub-Gaussian distribution (see Theorem 3.4.6 of \cite{vershynin2018high}). However, in our analysis, we provide a more detailed characterization of this property with explicit constants, which are later needed in part (b) of  Lemma \ref{tale:ineq} and the analysis in step 3.

The sub-Gaussian property of $\sqrt{p}\mathrnd{u}$ allows using the following uniform result:
\begin{tcolorbox}
    \begin{lemma}[Theorem 4.6.1 of \cite{vershynin2018high}]\label{versh:th}
        Let $\mathrnd{S}$ be an $n\times p$ matrix whose rows $\mathrnd{s}_i^\top$ are i.i.d, zero-mean, sub-gaussian isotropic random
vectors in $\R^p$. Let $\kappa = \|\mathrnd{s}_i\|_{\psi_2}$. Then for any $t\geq 0$ and all $x\in \eS^{p-1}$:
\[
\sqrt{n} - c\kappa^2\left( \sqrt{p}+t\right)\leq \|\mathrnd{S}x\|_2 \leq \sqrt{n} + c\kappa^2\left( \sqrt{p}+t\right),
\]
with probability at least $1 - 2 \exp \left(-t^2\right)$. 
\end{lemma}
\end{tcolorbox}
Lemma \ref{versh:th} provides an immediate way of uniformly bounding the singular values of $\sqrt{p}\mathrnd{W}$. To that end, notice that for $\mathrnd{g}\sim \mathcal{N}(0,I_p)$:
\[
\EE\left[\frac{\mathrnd{g}\mathrnd{g}^\top}{\|\mathrnd{g}\|^2}\right] = \frac{1}{p}I,
\]
which can be easily verified by symmetry for diagonal elements and negative symmetry for off-diagonal elements. This observation makes the rows of $\sqrt{p}\mathrnd{W}$ isotropic. Moreover, part (a) of Lemma \ref{tale:ineq} guarantees that the sub-Gaussian norm of $\sqrt{p}\mathrnd{u}$ is a constant, i.e., $\|\sqrt{p} \mathrnd{u}\|_{\psi_2}=\mathcal{O}(1)$. Now appealing to Lemma \ref{versh:th} and setting $t = c\sqrt{p}$ guarantees that with probability exceeding $1 - 2 \exp \left(-cp\right)$, for all $x\in \eS^{p-1}$:
\[
\sqrt{n} - c'\sqrt{p}\leq \|\sqrt{p}\mathrnd{W} x\|_2 \leq \sqrt{n} + c'\sqrt{p},
\]
or equivalently, with probability exceeding $1 - 2 \exp \left(-cp\right)$: 
\begin{equation}\label{eq:singVals}
    \frac{\sqrt{n} - c'\sqrt{p}}{\sqrt{p}}\leq s_{\min}(\mathrnd{W})\leq s_{\max}(\mathrnd{W}) \leq \frac{\sqrt{n} + c'\sqrt{p}}{\sqrt{p}}.
\end{equation}

\paragraph{\textbf{-- Step 3. High Probability Bounds for  $\|\mathrnd{W}\|_{2\to 4}$}:} In this section we focus on bounding $\|\mathrnd{W}\|_{2\to 4}$ using a covering argument. While the induced norm of random matrices with bounded and independent values has been studied in the literature  \cite{bennett1975norms, guedon2017expectation, lata2025operatorellptoellqnormsgaussian}, to the best of our knowledge, no such result is available for random matrices like \eqref{eq:W} with dependence across the columns. Here using part (b) of Lemma \ref{tale:ineq}, we will provide concentration bounds on $\|\mathrnd{W}\|_{2\to 4}$. 

Consider a random matrix $\mathrnd{W}\in\R^{n\times p}$ following the construction in \eqref{eq:W}. For a given $x\in\eS^{p-1}$ we have
\[
\|\mathrnd{W}x\|_4^4 = \sum_{i=1}^n \frac{|\mathrnd{g}_i^\top x|^4}{\|\mathrnd{g}_i\|^4}.
\]
For a fixed $\tau$ within the designated region of Lemma \ref{tale:ineq} we have
\begin{align}\notag 
     \EE \exp\left(\tau p \|\mathrnd{W}x\|_4^4 \right)\leq \exp\left(\frac{13\tau n}{p} \right).
\end{align} 
By the Markov's inequality for a non-negative random variable $\mathrnd{r}$ and all $t>0$: $\PP\{\mathrnd{r}\geq t\}\leq \EE \mathrnd{r}/t $. Fixing $\tau = \log(2)/6$ and applying the Markov's inequality to $\mathrnd{r} = \exp\left(\tau p \|\mathrnd{W}x\|_4^4 \right)$ with $t = \exp(13\tau n/p + \gamma p)$ guarantees that for any $\gamma>0$:
\begin{equation}\label{Markov}
      \PP\left\{ \tau p \|\mathrnd{W}x\|_4^4 \geq \frac{13\tau n}{p} + \gamma p\right\} = \PP\left\{ p^2 \|\mathrnd{W}x\|_4^4 \geq 13 n + \frac{6}{\log 2}\gamma p^2\right\}\leq  \exp(-\gamma p). 
\end{equation}
We now proceed by applying a union bound to the tale bound above using a covering argument. The following standard result (e.g., see Corollary 4.2.13 of \cite{vershynin2018high}) bounds the size of an $\epsilon$-net on $\eS^{p-1}$:
\begin{tcolorbox}
\begin{lemma}\label{lema:covering}
    The covering number of the unit sphere $\eS^{p-1} = \{x\in\R^p: \|x\|_2 =  1\}$ satisfies the following for any $\varepsilon\in(0,1]$:
    \[
    N(\eS^{p-1},\varepsilon)\leq \left( \frac{2}{\varepsilon}+1\right)^p,
    \]
    where $N(\eS^{p-1},\varepsilon)$ represents an $\varepsilon$-net of $\eS^{p-1}$, i.e., 
    \[
    \forall~\!x\in \eS^{p-1}, ~\exists~\!x'\in N(\eS^{p-1},\varepsilon): ~~ \|x-x'\|\leq \varepsilon. 
    \]
\end{lemma}
\end{tcolorbox}
We next relate $\|\mathrnd{W}\|_{2\to 4}$ to a version of it computed on $N(\eS^{p-1},\varepsilon)$ by fixing $\varepsilon\in(0,1]$ and using a standard covering argument for the norms.
Let $x\in\eS^{p-1}$ be a vector such that $\|\mathrnd{W}x\|_4 = \|\mathrnd{W}\|_{2\to 4}$. By the triangle inequality for any $x'\in N(\eS^{p-1},\varepsilon)$ we have:
\begin{align*}
    \|\mathrnd{W}x'\|_4\geq \|\mathrnd{W}x\|_4 - \|\mathrnd{W}(x-x')\|_4\geq \|\mathrnd{W}\|_{2\to 4} - \|\mathrnd{W}\|_{2\to 4}\|x-x'\|_2\geq (1-\varepsilon)\|\mathrnd{W}\|_{2\to 4}. 
\end{align*}
After fixing $\varepsilon = 1/2$, dividing both sides of this inequality by $1-\varepsilon$ and taking a maximum with respect to $x'$ we get
\begin{align}
    \|\mathrnd{W}\|_{2\to 4} \leq \frac{1}{1-\varepsilon}\sup_{x\in N(\eS^{p-1},\varepsilon)} \|\mathrnd{W}x\|_4=  2\sup_{x\in N(\eS^{p-1},1/2)} \|\mathrnd{W}x\|_4.
\end{align}
Note that by Lemma \ref{lema:covering} the cardinality of the net can be bounded by $|N(\eS^{p-1},1/2)|\leq 5^p$, which admits applying a union bound as follows:
\begin{align}\notag 
    \PP\left\{ \sqrt{p} \|\mathrnd{W}\|_{2\to 4} \geq 2\left(13 n + \frac{6}{\log 2}\gamma p^2\right)^{1/4}\!\right\}\! &\leq \!\PP\left\{ \sqrt{p} \!\!\!\sup_{x\in N(\eS^{p-1},1/2)}\!\!\! \|\mathrnd{W}x\|_4 \geq \left(13 n + \frac{6}{\log 2}\gamma p^2\right)^{1/4}\right\}\\
    \notag 
    &\leq \sum_{x\in N(\eS^{p-1},1/2)}\!\!\!\!\!\! \PP\left\{ \sqrt{p} \|\mathrnd{W}x\|_4 \geq \left(13 n + \frac{6}{\log 2}\gamma p^2\right)^{1/4}\right\}\\
    \notag &\leq 5^p \exp(-\gamma p)\\\notag 
    & = \exp\left((\log 5-\gamma)p \right),
\end{align}
which implies that for positive constants $\tilde c$ and $\tilde c'$, with probability exceeding $1-\exp(-\tilde c'p)$ we have $p^2\|\mathrnd{W}\|_{2\to 4}^4 \leq \tilde c(n+p^2)$, or equivalently 
\begin{equation}\label{indNorm:UB}
\|\mathrnd{W}\|_{2\to 4}^4 \leq \tilde c\frac{(n+p^2)}{p^2}. 
\end{equation}
\paragraph{\textbf{-- Step 4. Combining the Results of Previous Steps}:} Let's define
\[
\mu_{n,p} = \frac{p}{\left(\sqrt{n}+\sqrt{p}\right)^2},
\]
and multiply both sides of \eqref{ASU:doubleBound2} by $\mu_{n,p}$ to get
\begin{align}\label{ASU:doubleBound2-p}
    \mu_{n,p}s_{\min}^2(\mathrnd{W})\|x-y\|^2_2\leq \beta_{n,p}^{-1}&\left\|\texttt{ASU}(\mathrnd{W} x)\! -\! \texttt{ASU}(\mathrnd{W} y)\right\|^2 \leq  \mu_{n,p}\!\left(\!s^2_{\max}(\mathrnd{W})\|x-y\|^2_2 + \frac{6\pi^2} {10} \|\mathrnd{W}\|_{2\to 4}^4\!\!\right)\!,
\end{align}
where
\[
\beta_{n,p}^{-1} = \frac{\pi^2 p}{4\left(\sqrt{n}+\sqrt{p}\right)^2}.
\]
By \eqref{eq:singVals} for all $x,y\in \mathbb{B}_2^p$ and with probability exceeding $1 - 2 \exp \left(-cp\right)$ the left side of \eqref{ASU:doubleBound2-p} can be lower bounded as:
\begin{align}\notag 
    \mu_{n,p}s^2_{\min}(\mathrnd{W})\|x-y\|_2^2 \geq \left(\frac{\sqrt{n} - c'\sqrt{p}}{\sqrt{n} +\sqrt{p}}\right)^2\|x-y\|_2^2   &= \left(1- \frac{  (c'+1)\sqrt{p}}{\sqrt{n} +\sqrt{p}}\right)^2\|x-y\|_2^2 \\ \notag &\geq \|x-y\|^2 - \frac{  2(c'+1)\sqrt{p}}{\sqrt{n} +\sqrt{p}}\|x-y\|_2^2 \\ & \geq \|x-y\|_2^2 -  \frac{  8(c'+1)\sqrt{p}}{\sqrt{n} +\sqrt{p}}\label{smin:low}.
\end{align}
On the other hand, a combination of \eqref{indNorm:UB} and \eqref{eq:singVals} guarantees that for all $x,y\in \mathbb{B}_2^p$, with probability exceeding $1 - \exp(-\tilde c'p) -2\exp(-cp)$ the right side of \eqref{ASU:doubleBound2-p} can be upper-bounded as 
\begin{align}\notag 
  \mu_{n,p}s^2_{\max}(\mathrnd{W})\|x-y\|^2_2 + \frac{6\pi^2\mu_{n,p}}{10} \|\mathrnd{W}\|_{2\to 4}^4 &\leq \left(1 + \frac{ (c'-1)\sqrt{p}}{\sqrt{n} +\sqrt{p}}\right)^2 \|x-y\|_2^2 + \frac{6\tilde c\pi^2}{10}\frac{(n+p^2)}{p\left(\sqrt{n}+\sqrt{p}\right)^2}\\ &\leq \|x-y\|_2^2 + c_u\left( \frac{\sqrt{p}}{ \sqrt{n}+\sqrt{p}} + \frac{n+p^2}{p\left( \sqrt{n}+\sqrt{p}\right)^2}\right),\label{smax:high}
\end{align}
for some positive constant $c_u$. In the second inequality above we used the fact that $\|x-y\|_2^2\leq 4$ and $\frac{ (c'-1)\sqrt{p}}{\sqrt{n} +\sqrt{p}}<1$ for sufficiently large $n$. The outcomes of \eqref{smin:low} and \eqref{smax:high} combined with \eqref{ASU:doubleBound2-p} validate the claims made in Theorem \ref{near:ISO:Th}, and the proof is complete. 

\end{proof}
\subsection{Proof of Lemma \ref{rasu:bound}}\label{flfu:sec}
\begin{proof}
    Consider the following functions:
    \[
     f_\ell(x,y) =  2\left(1-\frac{4}{\pi^2}\right)\left(x^4+y^4\right) - \frac{4}{\pi^2}(x-y)^2 +  \frac{4}{\pi^2}\left(\arcsin(x) - \arcsin(y)\right)^2, 
    \]
    and 
    \[
    f_u(x,y) = 2\left(1-\frac{4}{\pi^2}\right)\left(x^4+y^4\right) + \frac{4}{\pi^2}(x-y)^2 -  \frac{4}{\pi^2}\left(\arcsin(x) - \arcsin(y)\right)^2. 
    \]
    To prove the lemma, it suffices to show that for $(x,y)\in[-1,1]\times [-1, 1]$, $f_\ell(x,y)\geq 0$ and $f_u(x,y)\geq 0$. To establish the inequality for $f_\ell(x,y)$, notice that by the Lipschitz continuity of the $\sin$ function:
    \begin{equation*}
        (\sin(\alpha) - \sin(\beta))^2\leq (\alpha-\beta)^2, 
    \end{equation*}
    which by setting $\alpha = \arcsin(x)$ and $\beta = \arcsin(y)$ and multiplying both sides by $4/\pi^2$ yields
    \begin{equation}\label{lips:Asin}
    \frac{4}{\pi^2}\left(\arcsin(x) - \arcsin(y)\right)^2 - \frac{4}{\pi^2}(x-y)^2 \geq 0,  
    \end{equation}
    from which it immediately follows that $f_\ell(x,y)\geq 0$.

    To show the inequality $f_u(x,y)\geq 0$, we begin by rearranging the terms of $f_u$ and bounding them as
    \begin{align} 
    \nonumber f_u(x,y)=&~2\left(1-\frac{4}{\pi^2}\right)\left(x^4+y^4\right)+\frac{4}{\pi^2}\left(x^2+y^2\right)-\frac{4}{\pi^2}\left(\arcsin^2(x)+\arcsin^2(y)\right)\\ \notag 
    & + 2 \left(\frac{4}{\pi^2}\right)(\arcsin(x)\arcsin(y)-xy) \\ \label{fu:ineq} \geq &~2\left(1-\frac{4}{\pi^2}\right)(x^4+y^4)+\frac{8}{\pi^2}(x^2+y^2)-\frac{8}{\pi^2}(\arcsin^2(x)+\arcsin^2(y))\\=& ~\frac{8}{\pi^2}\left( h(x) + h(y)\right),\notag
    \end{align}
    where
    \[
    h(x) = \left(\frac{\pi^2}{4}-1 \right)x^4 + x^2 - \arcsin^2(x).
    \]
    In the derivations above, inequality \eqref{fu:ineq} uses the fact
    \begin{equation*}
   \left(\arcsin(x) + \arcsin(y)\right)^2  \geq (x+y)^2,  
    \end{equation*}
    which is a direct implication of \eqref{lips:Asin}, when one uses $x$ and $-y$ as the arguments. Thanks to symmetry in $x$ and $y$, to show $f_u(x,y)\geq 0$ it suffices to show that $h(x)\geq 0$, and since $h$ is an even function, we can restrict the domain to $x\in[0,1]$. Notice that since $x^2\geq \frac{4}{\pi^2}\arcsin^2(x)$: 
    \begin{align*}
        h(x) &= \left(\frac{\pi^2}{4}-1 \right)x^4 + x^2 - \arcsin^2(x)\\ &\geq \left(1-\frac{4}{\pi^2}\right)x^2\arcsin^2(x) + x^2 - \arcsin^2(x) \\&\geq0,
    \end{align*}
where the second inequality is thanks to the elementary inequality 
\[
\arcsin(x)\leq \frac{x}{\sqrt{1-\left(1-\frac{4}{\pi^2}\right)x^2}},
\]
which is valid for $x\in[0,1]$ (e.g., see Theorem 1 of \cite{bagul2021simple}). This completes the proof. 
\end{proof}
\subsection{Proof of Lemma \ref{tale:ineq}}\label{MGF:proof}
\subsubsection{Auxiliary Lemmas Needed to Prove Lemma \ref{tale:ineq}}To prove the lemma, we first need to state two integral bounds (Lemma \ref{q2:intBound} and Lemma \ref{q4:intBound}) which will be later used in the main proof. Also to bound the moment generating functions, we need a tail bound which is stated in Lemma \ref{lem:CDF} below. We first present and prove these lemmas, and then combine them to prove Lemma \ref{tale:ineq}. 

\begin{tcolorbox}
    \begin{lemma}\label{q2:intBound}
    For all integers $p\geq 27$: 
    \[
    \int_0^1 \frac{dx}{(1+x^2)^{\frac{p}{6}-1}}\leq \sqrt{\frac{6}{p-6}}.
    \]
    \end{lemma}
\end{tcolorbox}
\begin{proof}
Consider $k\in\left\{\frac{n}{6}:n\in\mathbb{N}\right\}$. Defining $I_k$ as below and doing an integration by parts we get
\begin{align}\label{eq:Ik}
    I_{k} &:= \int_0^1 \frac{dx}{(1+x^2)^{k}} \\ \notag &= \left[ \frac{x}{(1+x^2)^k}\right]_0^1 + 2k\int_{0}^1\frac{x^2dx}{(1+x^2)^{k+1}}\\ \notag & = \frac{1}{2^k} + 2k\left( \int_{0}^1\frac{dx}{(1+x^2)^{k}} - \int_{0}^1\frac{dx}{(1+x^2)^{k+1}} \right),
\end{align}
which implies 
\begin{equation}\label{int:rec}
    I_{k+1} = \frac{1}{k 2^{k+1}} + \frac{2k-1}{2k}I_k.
\end{equation}
We now use an inductive argument and show if $I_k\leq k^{-1/2}$, then $I_{k+1}\leq {(k+1)}^{-1/2}$. To this end, we state the following lemma which is proved at the end of this section:
\begin{tcolorbox}
\begin{lemma}\label{lem:ineq}
    For any  $k\in\left\{\frac{n}{6}:n\geq 21, ~n\in\mathbb{N} \right\}$:
    \begin{equation}\label{ineq:lbub}
        \frac{1}{k2^{k+1}}< \frac{3}{10k^2\sqrt{k}}< \frac{1}{\sqrt{k+1}} - \frac{2k-1}{2k\sqrt{k}}.
    \end{equation}
\end{lemma}
\end{tcolorbox}
\noindent Using Lemma \ref{lem:ineq}, assuming $I_k\leq k^{-1/2}$, by the recursive equation \eqref{int:rec} we get
\begin{align}\notag 
    I_{k+1} &= \frac{1}{k2^{k+1}} + \frac{2k-1}{2k}I_k\\\notag &\leq \frac{3}{10k^2\sqrt{k}} + \frac{2k-1}{2k\sqrt{k}}\\&\leq \frac{1}{\sqrt{k+1}},
\end{align}
proving the inductive step. Notice that we have unit inductive step increments in \eqref{int:rec}, while the increments in $k$ are integer multiples of $1/6$. Hence, to establish the induction base case, it suffices to show that for some $n_0$, the integrals $I_{\frac{n_0}{6}}, I_{\frac{n_0+1}{6}}, \ldots I_{\frac{n_0+5}{6}}$ are dominated by $\sqrt{6/n_0},\sqrt{{6}/{(n_0+1)}},\ldots \sqrt{{6}/{(n_0+5)}},$ respectively. This can be verified for $n_0=6$. More specifically,  $I_1 = 0.7854<1$, $I_{7/6} = 0.7575<\sqrt{6/7}$, \ldots $I_{11/6} = 0.6628<\sqrt{6/11}$. Together with the common inductive step, this verifies that 
\begin{equation}\label{induction:result}
    \forall k\in\left\{\frac{n}{6}:n\geq 21, ~n\in\mathbb{N} \right\}: ~~ I_k\leq k^{-1/2}.
\end{equation} 
For a given integer $p\geq 27$, setting $k=p/6-1$ and appealing to \eqref{induction:result} gives
    \[
    \int_0^1 \frac{dx}{(1+x^2)^{\frac{p}{6}-1}}\leq \sqrt{\frac{6}{p-6}},
    \]
which completes the proof. We are only left with providing the proof of Lemma \ref{lem:ineq}, which is presented in the sequel. 

The first inequality in \eqref{ineq:lbub} is straightforward as the exponential expression $k2^{k+1}$ dominates the monomial $\frac{10}{3}k^2\sqrt{k}$ for sufficiently large $k$. It can be easily verified that the domination happens after $k\geq 21/6$. To prove the upper-bound, it suffices to show that
\[
\frac{3}{10k^2} + \frac{2k-1}{2k}<\sqrt{\frac{k}{k+1}},
\]
which after squaring both sides and rearranging the terms is equivalent to showing that
\[
(k+1)(3+10k^2-5k)^2 < k\left(10k^2\right)^2.
\]
Taking both expressions to one side and expanding the terms reveals that
\begin{align}
    k\left(10k^2\right)^2 - (k+1)(3+10k^2-5k)^2 = 15k^3 - 55k^2+21k -9,
\end{align}
which is a strictly positive quantity, simply because $15k^3+21k$ consistently dominates $55k^2+9$ for every integer $k\geq 3$. This verifies the upper-bound in \eqref{ineq:lbub}, and the proof of Lemma \ref{lem:ineq} is complete. 
\end{proof}
\begin{tcolorbox}
    \begin{lemma}\label{q4:intBound}
    For all integers $p\geq 27$: 
    \[
    \int_0^1 \frac{x^2 dx}{(1+x^2)^{\frac{p}{3}-1}}\leq \frac{3^{3/2}}{2(p-6)^{3/2}},
    \]
    \end{lemma}
\end{tcolorbox}
\begin{proof}
    Let $k\in\left\{\frac{n}{6}:n\in\mathbb{N}\right\}$, then
\begin{align}\notag 
    \int_0^1 \frac{x^2 dx}{(1+x^2)^{k} }&= \int_0^1 \frac{(1+x^2) dx}{(1+x^2)^{k}} - \int_0^1 \frac{ dx}{(1+x^2)^{k}}= I_{k-1} - I_k,
\end{align}
where $I_k$ follows the formulation in \eqref{eq:Ik}. By the recursive equation in \eqref{int:rec} which is valid for $k\in\left\{\frac{n}{6}:n\geq 21, ~n\in\mathbb{N} \right\}$ we have
\begin{align*}
    I_{k-1} - I_{k} &= -\frac{1}{(k-1) 2^{k}} + \frac{1}{2k-2}I_{k-1}\\&= -\frac{1}{(k-1) 2^{k}} + \frac{1}{2(k-1)\sqrt{k-1}}\\&\leq \frac{1}{2(k-1)^{3/2}},
\end{align*}
where the second inequality uses the fact $I_k\leq k^{-1/2}$ which was proved after equation \eqref{int:rec}. Setting $k = p/3-1$ (which still keeps $k$ an integer multiple of $1/6$) gives
 \[
    \int_0^1 \frac{x^2 dx}{(1+x^2)^{\frac{p}{2}-1}}\leq  \frac{3^{3/2}}{2(p-6)^{3/2}}.
    \]
This completes the proof. 
\end{proof}
The next lemma provides a tail bound for the random vectors of interest in Lemma \ref{tale:ineq}. 
\begin{tcolorbox}
    \begin{lemma}\label{lem:CDF}
        Consider $x\in\eS^{p-1}$ where $p\geq 3$, and $\mathrnd{g} = [\mathrnd{g}_1,\ldots,\mathrnd{g}_p]^\top\sim\mathcal{N}(0,I)$. For all $\lambda\in(0,1]$:
        \begin{equation*}
            \PP\left\{ \frac{\left| x^\top \mathrnd{g}\right|}{\|\mathrnd{g}\|}\geq \lambda \right\} \leq \sqrt{\frac{2}{\pi (p-2)}}\frac{(1+\lambda^2)^{1-\frac{p}{2}}}{\lambda}.
        \end{equation*}
    \end{lemma}
\end{tcolorbox}
\begin{proof}
Since $x\in\eS^{p-1}$ and the standard normal distribution is rotation invariant, we can set $x=e_1$ (the first canonical basis) which implies
 \begin{align}\notag 
            q_\lambda &= \PP\left\{ \frac{\left| x^\top \mathrnd{g}\right|}{\|\mathrnd{g}\|}\geq \lambda \right\} \\\notag & = \PP\left\{ \frac{|\mathrnd{g}_1|}{\|\mathrnd{g}\|}\geq \lambda \right \}\\ \notag & = \PP\left\{ |\mathrnd{g}_1|\geq \lambda\sqrt{\mathrnd{g}_1^2+\mathrnd{g}_2^2\ldots + \mathrnd{g}_p^2}\right\} \\& \notag \leq \PP\left\{ |\mathrnd{g}_1|\geq \lambda\sqrt{\mathrnd{g}_2^2+\mathrnd{g}_3^2\ldots + \mathrnd{g}_p^2}\right\}
            \\&  = \PP\left\{ \frac{\left(\sum_{i=2}^p \mathrnd{g}_i^2\right)/(p-1)}{\mathrnd{g}_1^2}\leq \frac{1}{\lambda^2(p-1)}\right\},\label{fcdf}
\end{align}
where the inequality is thanks to the fact that 
\[\left\{ z\in\R^p: |z_1|\geq \lambda\sqrt{z_1^2+z_2^2\ldots + z_p^2}\right\}\subseteq \left\{ z\in\R^p: |z_1|\geq \lambda\sqrt{z_2^2+z_3^2\ldots + z_p^2}\right\}.
\]
Notice that the random variable $\frac{\left(\sum_{i=2}^p \mathrnd{g}_i^2\right)/(p-1)}{\mathrnd{g}_1^2}$ is the ratio of two independent chi-squared random variables each normalized by their degrees of freedom,  hence follows an $F$-distribution with $p-1$ and $1$ degrees of freedom, denoted as $\mathrnd{F}_{p-1,1}$. Generally, the cumulative distribution function (CDF) of $\mathrnd{F}_{d_1,d_2}$ at a given point $z$ is evaluated as
\begin{align*}
     \PP\left\{ \mathrnd{F}_{d_1,d_2}\leq z\right\} = \frac{B\left(\frac{d_1z}{d_1z+d_2};\frac{d_1}{2},\frac{d_2}{2}\right)}{B\left(1;\frac{d_1}{2},\frac{d_2}{2}\right)},
\end{align*}
where $B$ represents an incomplete beta function:
\[
B(z;a,b) = \int_{0}^z t^{a-1}(1-t)^{b-1}dt.
\]
Notice that the probability in \eqref{fcdf} is basically the CDF of $\mathrnd{F}_{p-1,1}$ evaluated at $\left(\lambda^2(p-1)\right)^{-1}$, therefore
\begin{align*}
    q_\lambda &\leq \PP\left\{ \mathrnd{F}_{p-1,1}\leq \frac{1}{\lambda^2(p-1)}\right\} = \frac{B\left(\frac{1}{1+\lambda^2};\frac{p-1}{2},\frac{1}{2}\right)}{B\left(1;\frac{p-1}{2},\frac{1}{2}\right)} = \frac{B\left(\frac{1}{1+\lambda^2};\frac{p-1}{2},\frac{1}{2}\right)\Gamma\left(\frac{p}{2}\right)}{\Gamma(\frac{1}{2})\Gamma\left(\frac{p-1}{2}\right)},
\end{align*}
where in the last equation we used the well-known relationship between the beta function and the gamma function:
\[
B(1;a,b) = \frac{\Gamma(a)\Gamma(b)}{\Gamma(a+b)}. 
\]
Since $\Gamma(1/2)=\sqrt{\pi}$, and for $p\geq 1$, by the Gautschi's inequality:
\[
\frac{\Gamma\left( \frac{p-1}{2}\right)}{\Gamma\left(\frac{p}{2}\right)}\geq \sqrt{\frac{2}{p}},
\]
one can bound $q_\lambda$ as
\begin{align*}
    q_\lambda &\leq \sqrt\frac{p}{2\pi} B\left(\frac{1}{1+\lambda^2};\frac{p-1}{2},\frac{1}{2}\right)\\& = \sqrt\frac{p}{2\pi} \int_0^{\frac{1}{1+\lambda^2}} \frac{t^{\frac{p-3}{2}}dt}{\sqrt{1-t}}\\&\leq \sqrt\frac{p}{2\pi} \sqrt{\frac{1+\lambda^2}{\lambda^2}}\int_0^{\frac{1}{1+\lambda^2}}t^{\frac{p-3}{2}}dt\\ & = \sqrt\frac{2p}{\pi(p-1)^2}  \frac{\left(1+\lambda^2 \right)^\frac{2-p}{2}}{\lambda}
    \\ & \leq  \sqrt\frac{2}{\pi(p-2)}  \frac{\left(1+\lambda^2 \right)^\frac{2-p}{2}}{\lambda},
\end{align*}
where in the second inequality we used $\sqrt{1-t}\geq \sqrt{\frac{\lambda^2}{1+\lambda^2}}$ when $0\leq t\leq \frac{1}{1+\lambda^2}$. This completes the proof. 
\end{proof}
\subsubsection{Main Proof of Lemma \ref{tale:ineq}}
We are now ready to use the results of the auxiliary lemmas stated above to prove Lemma \ref{tale:ineq}. 

Consider $x\in\eS^{p-1}$ where $p\geq 3$, and $\mathrnd{g} = [\mathrnd{g}_1,\ldots,\mathrnd{g}_p]^\top\sim\mathcal{N}(0,I)$. We define the random variable 
\[
\mathrnd{u} = \frac{\left| x^\top \mathrnd{g}\right|}{\|\mathrnd{g}\|}.
\]
Notice that by construction $0\leq \mathrnd{u}\leq 1$, which combined with the result of Lemma \ref{lem:CDF} gives
\begin{equation}
    \PP\left\{ \mathrnd{u}> \lambda \right\}: \left\{\begin{array}{lc}=0 & \lambda\geq 1\\ \leq \sqrt{\frac{2}{\pi (p-2)}}\frac{(1+\lambda^2)^{1-\frac{p}{2}}}{\lambda} & 0<\lambda <1\\ =1 & \lambda\leq 0 \end{array} \right..
\end{equation}
As a result, for all $t\geq 0$ and integers $q\geq 2$:
\begin{align}\notag 
    \EE \exp\left(t\mathrnd{u}^q\right) = \int_0^1 \exp\left(t\lambda^q\right)d\left(\PP\left\{ \mathrnd{u}\leq  \lambda \right\} \right) &= - \int_0^1 \exp\left(t\lambda^q\right)d\left(\PP\left\{ \mathrnd{u}>  \lambda \right\} \right)\\ \notag & = 1 + \int_0^1 tq\lambda^{q-1}\exp\left(t\lambda^q\right) \PP\{\mathrnd{u}>\lambda\}d\lambda \\& \leq 1+ tq\sqrt{\frac{2}{\pi (p-2)}}\int_0^1 \lambda^{q-2}\exp\left(t\lambda^q\right) (1+\lambda^2)^{1-\frac{p}{2}}d\lambda \label{expq:bound}
\end{align}
where the third equality is thanks to integration by parts and the fact that $\PP\{\mathrnd{u}>1\} = 1-\PP\{\mathrnd{u}>0\}=0$. 
\paragraph{\textbf{Part (a) Proof}}
Consider $q = 2$, then by \eqref{expq:bound} we get
\begin{align}\notag 
 \EE \exp\left(t\mathrnd{u}^2\right) &\leq  1+2t\sqrt{\frac{2}{\pi (p-2)}}\int_0^1 \exp\left(t\lambda^2\right) (1+\lambda^2)^{1-\frac{p}{2}}d\lambda  \\ \notag & \leq 1+2t\sqrt{\frac{2}{\pi (p-2)}}\int_0^1 \left( 1+\lambda^2\right)^{\frac{t}{\log 2}} (1+\lambda^2)^{1-\frac{p}{2}}d\lambda
 \\&= 1+2t\sqrt{\frac{2}{\pi (p-2)}}\int_0^1 (1+\lambda^2)^{1 + \frac{t}{\log 2} -\frac{p}{2}}d\lambda,\label{MGF:upper}
\end{align}
where in the second inequality we used the fact that for all $z\in[0,1]$: $\exp(z)\leq (1+z)^{\frac{1}{\log 2}}$ which implies that 
\begin{equation*}
    \forall z\in[0,1], ~ t\geq 0: ~~ \exp(tz)\leq (1+z)^{\frac{t}{\log 2}}.
\end{equation*}

For all $\tau\in\left [0, \sqrt{\log(2)}/\sqrt{3}\right ]$,
setting $t = \tau^2 p$ in \eqref{MGF:upper} gives 
\begin{align}\notag 
     \EE \exp\left(\tau^2 p \mathrnd{u}^2\right)&\leq 1 + 2\tau^2 p \sqrt{\frac{2}{\pi (p-2)}}\int_0^1 (1+\lambda^2)^{1 + \frac{\tau^2 p}{\log 2} -\frac{p}{2}}d\lambda\\ &\notag  \leq 1 + 2\tau^2 p \sqrt{\frac{2}{\pi (p-2)}}\int_0^1 (1+\lambda^2)^{1 -\frac{p}{6}}d\lambda.
\end{align}
Now using Lemma \ref{q2:intBound}, we have  for all $\tau\in\left [0, \sqrt{\log(2)}/\sqrt{3}\right ]$ and $p\geq 27$:
\begin{align*}
     \EE \exp\left(\tau^2 p \mathrnd{u}^2\right)&  \leq 1 + 2\tau^2 p \sqrt{\frac{12}{\pi (p-2)(p-6)}} \\\notag &\leq 1 + \frac{25}{4}\tau^2 
     \\&\leq \exp\left( \frac{25}{4}\tau^2  \right).
\end{align*}
Since $[0,2/5]\subset \left [0, \sqrt{\log(2)}/\sqrt{3}\right ]$ the advertised claim is valid. 

\paragraph{\textbf{Part (b) Proof}} Notice that
by \eqref{expq:bound} we have
\begin{align*}
 \EE \exp\left(t\mathrnd{u}^4\right) \leq  1+4t\sqrt{\frac{2}{\pi (p-2)}}\int_0^1 \lambda^2 (1+\lambda^2)^{1 + \frac{t}{\log 2} -\frac{p}{2}}d\lambda.
\end{align*}
For all $\tau\in\left [0, \log(2)/6\right ]$,
setting $t = \tau p$ gives 
\begin{align}\notag 
     \EE \exp\left(\tau p \mathrnd{u}^4\right)&\leq 1 + 4\tau p \sqrt{\frac{2}{\pi (p-2)}}\int_0^1 \lambda^2 (1+\lambda^2)^{1 + \frac{\tau p}{\log 2} -\frac{p}{2}}d\lambda\\ &\notag  \leq 1 + 4\tau p \sqrt{\frac{2}{\pi (p-2)}}\int_0^1 \lambda^2 (1+\lambda^2)^{1 -\frac{p}{3}}d\lambda.
\end{align}
Now using Lemma \ref{q4:intBound} we get
\begin{align}\notag 
     \EE \exp\left(\tau p \mathrnd{u}^4\right)&\leq 1 + 2\tau p \sqrt{\frac{54}{\pi (p-2)(p-6)^3}}\leq 1 + \frac{13\tau}{p}\leq \exp\left(\frac{13\tau}{p} \right).
\end{align}

\section{Proof of Theorem \ref{thm1} }
\begin{tcolorbox}[colback=teal!20]
    \renewcommand\thetheorem{\ref{thm1}}
\begin{theorem}
Suppose that $u,v \in \eS^{p-1}$ are unit vectors. Let $\mathrnd{r}$ be a Rademacher random vector in $\mathbb{R}^p$. Then,
$$
\left| \mathbb{E} \left[ \sign\left( \mathrnd{r}^\top u \right) \sign \left( \mathrnd{r}^\top v \right) \right]
- \frac{2}{\pi} \arcsin(u^\top v) \right| \le c g\left(u,\frac{v - \langle u,v\rangle u}{\|v - \langle u,v\rangle u\|}\right),
$$
where $c = 264$ is an absolute constant, and for two unit vectors $w$ and $w'$ in $\eS^{p-1}$:
\begin{equation}\label{g:equation}
g(w,w') = \sum_{i=1}^p (w_i^2+{w_i'}^2)^{3/2}.
\end{equation}
\end{theorem}
\end{tcolorbox}
\begin{proof}
A main component of the proof is the following result due to Rai\^c  \cite{Rai2019}. 
\begin{tcolorbox}
\begin{theorem}[Multivariate Berry-Esseen \cite{Rai2019}]
\label{berryesseen}
Let $\mathrnd{x}_1,\ldots,\mathrnd{x}_p$ be independent random vectors in $\mathbb{R}^d$. Assume that $\mathbb{E} \mathrnd{x}_i = 0$ for all $i$ and that $\sum_{i=1}^p \Cov(\mathrnd{x}_i) = I_d$. Let $\mathrnd{y}= \sum_{i=1}^p \mathrnd{x}_i$. Then, for all convex sets $A \subseteq \mathbb{R}^d$, we have
$$
\left|\mathbb{P}\{\mathrnd{y} \in A\} - \mathbb{P}\{\mathrnd{z} \in A\}\right| \le (42 d^{1/4} + 16) \sum_{i=1}^p \mathbb{E} \|\mathrnd{x}_i\|_2^3,
$$
where $\mathrnd{z}$ is a random vector with a standard normal distribution $\mathcal{N}(0,I_d)$.
\end{theorem}
\end{tcolorbox}
Consider $\mathcal{W}$ be the subspace formed by $u$ and $v$ and columns of $W \in \R^{p\times 2}$ forming an orthonormal basis for $\mathcal{W}$. Consider $W = \begin{bmatrix} w,w'\end{bmatrix}$. 
Projecting any vector $x$ onto $\mathcal{W}$ is done through $
\mathcal{P}_{\mathcal{W}}(x) = W^\top x$. 
Since projection does not expand the angles we should have
\begin{align}\label{projExpand}
    \sign\left(\mathrnd{r}^\top u\right) &= \sign\left(\mathrnd{r}^\top W W^\top \mathrnd{u}\right) = \sign\left(\mathrnd{r}^\top W W^\top u/\left\| W^\top u\right\|\right ),
\end{align}
and a similar relation holds for $v$. Define
\[
\tilde u := W^\top u/\left\| W^\top u\right\|, \qquad \tilde v := W^\top v/\left\| W^\top v\right\|.
\]
The random vector $W^\top \mathrnd{r}$ can be viewed as the sum $W^\top \mathrnd{r} = \sum_{i=1}^p \mathrnd{x}_i$ for 
\[
\mathrnd{x}_i = \mathrnd{r}_i\begin{pmatrix}
    w_i\\w'_i
\end{pmatrix}, ~ i=1,\ldots,p, 
\]
where the $i$ subscript denotes the $i$-th element of each vector. Note that by construction $\mathrnd{x}_i$ are independent, $\EE\mathrnd{x}_i=0$ and
\begin{align*}
    \sum_{i=1}^p \Cov(\mathrnd{x}_i) = \EE\left[ W^\top \mathrnd{r}\mathrnd{r}^\top W \right]  =W^\top \EE\left[\mathrnd{r}\mathrnd{r}^\top\right]W = W^\top I_n W = I_2, 
\end{align*}
which indicates that $\mathrnd{x}_i$ meet the conditions stated in Theorem \ref{berryesseen}. 
We now have
\begin{align*}
    \mathbb{E} \left[ \sign\left( \mathrnd{r}^\top u \right)\sign\left( \mathrnd{r}^\top v \right)\right]  &= \mathbb{E} \left[ \sign\left( \mathrnd{r}^\top W \tilde u \right)\sign\left( \mathrnd{r}^\top W \tilde v \right)\right] \\ & = 2\PP \left\{ \mathrnd{r}^\top W \tilde u> 0, \mathrnd{r}^\top W \tilde v> 0 \right\} - 2\PP \left\{ \mathrnd{r}^\top W \tilde u> 0, \mathrnd{r}^\top W \tilde v<0 \right\}, 
\end{align*}
where the first equality is thanks to \eqref{projExpand} and the second equality is a simple expansion of the expectation, using the fact that $\sign(0)=0$ and due to symmetry of $\mathrnd{r}$:
\begin{align*}
\PP \left\{ \mathrnd{r}^\top W \tilde u>0, \mathrnd{r}^\top W \tilde v>0 \right\} &= \PP \left\{ \mathrnd{r}^\top W \tilde u<0, \mathrnd{r}^\top W \tilde v<0 \right\}, 
\\ \PP \left\{ \mathrnd{r}^\top W \tilde u>0, \mathrnd{r}^\top W \tilde v<0 \right\} &= \PP \left\{ \mathrnd{r}^\top W \tilde u<0, \mathrnd{r}^\top W \tilde v>0 \right\}.
\end{align*}
For $\mathrnd{z}\sim\mathcal{N}(0,I_2)$, we have
\begin{align*}
\PP \left\{ \mathrnd{z}^\top \tilde u>0, ~\pm \mathrnd{z}^\top \tilde v>0 \right\} =  \frac{1}{4} \pm \frac{\arcsin(\tilde u^\top \tilde v)}{2\pi} = \frac{1}{4} \pm \frac{\arcsin(u^\top  v)}{2\pi},
\end{align*}
where the second equality holds since the angle between $\tilde u$ and $\tilde v$ is the same as that of $u$ and $v$. Applying the triangle inequality and twice appealing to Theorem \ref{berryesseen} we get
\begin{align}\notag 
     \bigg|\mathbb{E} \left[ \sign\left( \mathrnd{r}^\top u \right) \sign \left( \mathrnd{r}^\top v \right) \right]
- & \frac{2}{\pi} \arcsin(u^\top v)\bigg|\\ \notag \leq ~&2\left|\PP \left\{ \mathrnd{r}^\top W \tilde u>0, \mathrnd{r}^\top W \tilde v>0 \right\} -  \PP \left\{ \mathrnd{z}^\top \tilde u>0,  \mathrnd{z}^\top \tilde v>0 \right\} \right|\\ \notag &+ 2\left|\PP \left\{ \mathrnd{r}^\top W \tilde u>0, \mathrnd{r}^\top W \tilde v<0 \right\} -  \PP \left\{ \mathrnd{z}^\top \tilde u>0,  \mathrnd{z}^\top \tilde v<0 \right\} \right|\\ \notag \leq ~&  4\left(42 (2)^{1/4}+16\right)\sum_{i=1}^p \mathbb{E} \|\mathrnd{x}_i\|_2^3
\\ \notag \leq ~&  264\sum_{i=1}^p \mathbb{E} \|\mathrnd{x}_i\|_2^3\\ =& ~264\sum_{i=1}^p \left(w_i^2 + {w'_i}^2\right)^{3/2}.\label{discBound}
\end{align}
Since this inequality holds for all selections of $w$ and $w'$, we can specifically pick the ones offered by the Gram–Schmidt procedure as 
    \begin{equation}\label{GS}
        w = u, \qquad w' = \frac{v - \langle u,v\rangle u}{\|v - \langle u,v\rangle u\|}, 
    \end{equation}
which implies that
\begin{equation}\label{main:ineq}
\left| \mathbb{E} \left[ \sign\left( \mathrnd{r}^\top u \right) \sign \left( \mathrnd{r}^\top v \right) \right]
- \frac{2}{\pi} \arcsin(u^\top v) \right| \le C g\left(u,\frac{v - \langle u,v\rangle u}{\|v - \langle u,v\rangle u\|}\right).
\end{equation}
In the sequel we show that the Gram-Schmidt selection is already tight and the right-hand side expression cannot be made any tighter. For this purpose, assume that we pick a different $W$ matrix with columns $\tilde w$ and $\tilde w'$. Since $\tilde w$ and $\tilde w'$ must reside within the same plane as $w$ and $w'$ in \eqref{GS}, and still meet the orthonormality conditions, we must have
\begin{align*}
    \tilde w &= \sin(\alpha) w + \cos(\alpha) w'\\
    \tilde w' &= -\cos(\alpha) w + \sin(\alpha) w'
\end{align*}
for some $\alpha\in [0,2\pi)$. Now, notice that
\begin{align*}
    g(\tilde w, \tilde w') &= \sum_{i=1}^p\left( \tilde w_i^2 + {\tilde w'}_i {}^2\right)^{3/2}\\ & = \sum_{i=1}^p\left( \left(\sin(\alpha) w_i + \cos(\alpha) w'_i\right)^2 + \left(-\cos(\alpha) w_i + \sin(\alpha) w_i'\right)^2\right)^{3/2}\\ & = \sum_{i=1}^p\left(  w_i^2 +  {w'_i}^2\right)^{3/2}\\& = g(w,w'),
\end{align*}
which indicates that the right-hand side of the inequality \eqref{main:ineq} is oblivious to the selection of $W$, and all such selections present an identical right-hand side. 
\end{proof}

\section{Proof of Corollary \ref{corr:rad}}
\begin{tcolorbox}[colback=teal!20]
    \renewcommand\thecorollary{\ref{corr:rad}}
\begin{corollary}
Consider unit vectors $u, v\in\eS^{p-1}$ such that $\|u\|_\infty=\mathcal{O}(p^{-1/2})$,  $\|v\|_\infty=\mathcal{O}(p^{-1/2})$, and there exists some constant $c>0$ such that $|\langle u,v\rangle|\leq 1-c$. Then
    \[
 \left|\mathbb{E} \left[ \sign\left( \mathrnd{r}^\top u \right) \sign \left( \mathrnd{r}^\top v \right) \right]
- \frac{2}{\pi} \arcsin(u^\top v)\right|= \mathcal{O}\left(p^{-1/2}\right).
\]
\end{corollary}
\end{tcolorbox}

\begin{proof}
    We only need to show that $g(w,w') = \mathcal{O}(p^{-1/2})$ for $w$ and $w'$ picked as \eqref{GS} and $g$ defined as \eqref{g:equation}. Clearly, by construction $\|w\|_\infty=\mathcal{O}(p^{-1/2})$. On the other hand
    \begin{align}\notag 
        \|v - \langle u,v\rangle u\|_\infty &\leq \|v\|_\infty + |\langle u,v\rangle|\|u\|_\infty \\ \notag &\leq  \|v\|_\infty + \|u\|_2\|v\|_2\|u\|_\infty
        \\ \notag &=  \|v\|_\infty + \|u\|_\infty\\ &=\mathcal{O} (p^{-1/2}).\label{ineq1}
    \end{align}
    Moreover,
    \begin{align}\notag 
        \|v - \langle u,v\rangle u\|_2 &\geq \|v\|_2 - |\langle u,v\rangle|\|u\|_2 \\ \notag &=  1  - |\langle u,v\rangle|\\&\geq c.\label{ineq2}
    \end{align}
Combining \eqref{ineq1} and \eqref{ineq2} implies $\|w'\|_\infty = \mathcal{O}(p^{-1/2})$, and therefore 
\begin{align*}
    g(w,w') &= \sum_{i=1}^p (w_i^2 + {w'_i}^2)^{3/2} \\&\leq \max_{i=1,\ldots,p}\left\{ \left\|\begin{pmatrix} w_i \\ w'_i\end{pmatrix} \right\| \right\}\sum_{i=1}^p (w_i^2 + {w'_i}^2)\\ & = 2\max_{i=1,\ldots,p}\left\{ \left\|\begin{pmatrix} w_i \\ w'_i\end{pmatrix} \right\|\right\} \\& = \mathcal{O}(p^{-1/2}). 
\end{align*}
\end{proof}

\section{~~Proof of Proposition \ref{RASU:layerRad}}
\begin{tcolorbox}[colback=teal!20]
    \renewcommand\theproposition{\ref{RASU:layerRad}}
\begin{proposition}
Consider a similar RASU (or ASU) layer as Theorem \ref{RASU:layer}, where  for fixed constants $C,\delta > 0$, $\|x\|_\infty \le C p^{-1/2}$, $\|w_i\|_\infty \leq  C p^{-1/2}$, and $|w_i^\top x| \le 1 -\delta$ for all $i \in \{1,\ldots,n\}$. Define the embedded output as $\tilde{\mathrnd{y}} = \texttt{ReLU}\left(\texttt{sign}(W \mathrnd{R}^\top)\, \texttt{sign}(\mathrnd{R} x)\right)$ (or $\tilde{\mathrnd{y}} =\texttt{Id}\left( \texttt{sign}(W \mathrnd{R}^\top)\, \texttt{sign}(\mathrnd{R} x)\right)$), where $\mathrnd{R} \in \mathbb{R}^{N \times p}$ is a Rademacher matrix. Then, with high probability:
$$
\left\| \frac{1}{N} \tilde{\mathrnd{y}} - y \right\|_2 \leq \mathcal{O}\left(\sqrt{\frac{n \log n }{N}} + \sqrt{\frac{n}{p}}\right).
$$
\end{proposition}
\end{tcolorbox}
\begin{proof}
The main difference between Theorem \ref{RASU:layer} 
and Proposition \ref{RASU:layerRad}  is that the Gaussian matrix $\mathrnd{G} \in \mathbb{R}^{N \times p}$ is replaced by a Radamacher matrix $\mathrnd{R} \in \mathbb{R}^{N \times p}$.
Recall that $\mathrnd{g}_j^\top$ denotes the $j$-th row of $\mathrnd{G}$ and note that $\mathrnd{g}_j$ enters the proofs of Theorem \ref{RASU:layer} through the random variables 
\begin{equation} \label{Z}
\mathrnd{z}_{ij} = \sign\left(\mathrnd{g}_j^\top w_i\right)\sign\left(\mathrnd{g_j}^\top x\right).
\end{equation}
These proofs rely on three properties of these random variables:
\begin{enumerate}[label=(P\arabic{*}), ref=(P\arabic{*})]
\item $-1 \le \mathrnd{z}_{ij} \le 1$ for all $(i,j) \in \{1,\ldots,n\} \times \{1,\ldots,p\}$, \label{Zprop1}
\item For fixed $i \in \{1,\ldots,n\}$, the random variables $\mathrnd{z}_{i1},\ldots,\mathrnd{z}_{ip}$ are i.i.d., \label{Zprop2}
\item The expected value $\mathbb{E} \mathrnd{z}_{ij} = \frac{2}{\pi} \arcsin( w_i^\top x)$. \label{Zprop3}
\end{enumerate}
When we replace the rows $\mathrnd{g}_j^\top$ of $\mathrnd{G}$ with the rows $\mathrnd{r}_j^\top$ of $\mathrnd{R}$, and define
\begin{equation} \label{approxZ}
\tilde{\mathrnd{z}}_{ij} = \sign\left(\mathrnd{r}_j^\top w_i\right)\sign\left(\mathrnd{r_j}^\top x\right),
\end{equation}
properties \ref{Zprop1} and \ref{Zprop2} continue to hold for $\tilde{\mathrnd{z}}_{ij}$, and by Corollary \ref{corr:rad}, property \ref{Zprop3} approximately holds
\begin{equation} \label{ZapproxE}
\mathbb{E}(\tilde{\mathrnd{z}}_{ij}) = \frac{2}{\pi} \arcsin(w_i^\top x) + \mathcal{O}(p^{-1/2}).
\end{equation}
If we define $y'_i = \text{ReLU} (\mathbb{E}(\tilde{\mathrnd{z}}_{i1}))$ (or $y_i' = \mathbb{E}(\tilde{\mathrnd{z}}_{i1})$ for the \texttt{ASU} case) for $i \in \{1,\ldots,n\}$, then the proof of Theorem \ref{RASU:layer} implies that
for any $c > 0$, 
$$
\left\| \frac{1}{N} \tilde{\mathrnd{y}} - y' \right\|_2 \leq \sqrt{\frac{2(c + \log 2n )n}{N}},
$$
with probability at least $1 - \exp(-c)$. Using \eqref{ZapproxE} and the triangle inequality we get
\begin{equation} \label{Zfinalest}
\left\| \frac{1}{N} \tilde{\mathrnd{y}} - y \right\|_2 \leq \mathcal{O}\left(\sqrt{\frac{n \log n }{N}} + \sqrt{\frac{n}{p}}\right)
\end{equation}
with high probability, which completes the proof.
\end{proof}

\section{~~Proof of Proposition \ref{TASU:layerRad}}
\begin{tcolorbox}[colback=teal!20]
\renewcommand\theproposition{\ref{TASU:layerRad}}
\begin{proposition}
  Consider a similar TASU layer as Theorem \ref{TASU:layer},  where additionally for fixed constants $C,\delta > 0$, $\|x\|_\infty \le C p^{-1/2})$, $\|w_i\|_\infty \leq  C p^{-1/2}$, and $0<\ell_{\min}\leq |w_i^\top x| \le 1 -\delta$ for all $i \in \{1,\ldots,n\}$.  Define the embedded layer output as $\tilde{\mathrnd{y}}= \texttt{sign}\left(\texttt{sign}(W \mathrnd{R}^\top)\, \texttt{sign}(\mathrnd{R} x)\right)$, where $\mathrnd{R} \in \mathbb{R}^{N \times p}$ is a Rademacher matrix. Assume $C p^{-1/2} \le \ell_\text{min}/\pi$. Fix $\varepsilon\leq \sqrt{n}$ and set
$\kappa \geq \frac{\pi}{\ell_{\min}}\log \frac{4\sqrt{n}}{\varepsilon}.$
Then, picking $N = \mathcal{O}(n\kappa^2 \log n/\varepsilon^2)$ guarantees that $\|y-\tilde{\mathrnd{y}}\|_2 = \mathcal{O}(\varepsilon + \kappa \sqrt{n/p})$ with high probability.
\end{proposition}
\end{tcolorbox}
\begin{proof}
The proof is similar to that of Theorem \ref{TASU:layer} and we outline the main steps. The discrepancy between the two layers can be bounded as $\|\mathrnd{e}\|_2 = \|\tilde{\mathrnd{y}}- y\|_2 \leq \|\mathrnd{e}'\|_2 + \|\mathrnd{e}''\|_2$, 
where following \eqref{approxZ} as the notation:
\begin{align*}
\mathrnd{e}'_i &=\tanh\left( \frac{\kappa }{N}\sum_{j=1}^N \tilde{\mathrnd{z}}_{ij}
     \right) - \tanh\left( \frac{2\kappa}{\pi}\arcsin\left(w_i^\top x\right)\right),
\end{align*}
and 
\[
\mathrnd{e}''_i = \sign\left(\frac{1}{N}\sum_{j=1}^N \tilde{\mathrnd{z}}_{ij}\right) - \tanh\left( \frac{\kappa }{N}\sum_{j=1}^N \tilde{\mathrnd{z}}_{ij}
     \right).
\]
The term $\mathrnd{e}'$ in the proof of Theorem \ref{TASU:layer} can be handled in a similar way as in the proof of Theorem \ref{RASU:layerRad}, to indicate that with probability exceeding $1-\exp(-c)$:
$$ 
 \|\mathrnd{e}'\|_2 \leq \sqrt{\frac{2(c+\log(2n))n\kappa^2}{N}} +  C\sqrt{ \frac{\kappa^2 n}{p}},
$$
where the additional term $\sqrt{\kappa^2 n/p}$ comes from the Lipschitz continuity of $\tanh(\kappa\cdot)$ and a triangle inequality similar to \eqref{ZapproxE} and \eqref{Zfinalest}. 

For the second term we similarly have
\begin{align*}
     |\mathrnd{e}_i''|\leq 2\exp\left( -2\kappa\left|\frac{1 }{N}\sum_{j=1}^N \tilde{\mathrnd{z}}_{ij}\right|
     \right).
\end{align*}
Applying Hoeffding's inequality and the triangle inequality guarantees that with probability exceeding $1-2\exp(-c)$:
\begin{align*}
    \left|\frac{1 }{N}\sum_{j=1}^N\tilde{\mathrnd{z}}_{ij}\right| > \frac{2}{\pi} \ell_{\min}  - \frac{\sqrt{2c}}{\sqrt{N}} - \frac{C}{\sqrt{p}}.
\end{align*}
This indicates that with probability exceeding $1-2\exp(-c)$:
\[
\|\mathrnd{e}''\|_2\leq 2\exp\left(\log \sqrt{n} -\frac{4\kappa}{\pi}\ell_{\min} + \kappa\sqrt{\frac{8(c+\log n)}{N}}
    + 2\frac{C\kappa }{\sqrt{p}} \right).
\]
Setting $N = 8(c+\log 2 n)n\kappa^2/\varepsilon^2$, guarantees with probability exceeding $1-3\exp(-c)$:
\begin{align*}
    \|\mathrnd{e}\|_2\leq \frac{\varepsilon}{2} + C\sqrt{ \frac{\kappa^2 n}{p}} + 2\exp\left( \log \sqrt{n} -\frac{4\kappa}{\pi}\ell_{\min} + \frac{\varepsilon}{\sqrt{n}} + 2\frac{C\kappa }{\sqrt{p}} \right).
\end{align*}
If we pick $\kappa$ in a way that
\begin{equation}\label{cond:kappa}
\frac{4\kappa}{\pi}\ell_{\min}\geq 2\log \frac{4\sqrt{n}}{\varepsilon} + 2\frac{C\kappa }{\sqrt{p}},
\end{equation}
then by a similar argument as before we will have
\begin{align*}
   \frac{4\kappa}{\pi}\ell_{\min} \geq 2\log \frac{4\sqrt{n}}{\varepsilon} + 2\frac{C\kappa }{\sqrt{p}} \geq  \log \sqrt{n}  + \frac{\varepsilon}{\sqrt{n}} + 2\frac{C\kappa }{\sqrt{p}} - \log \frac{\varepsilon}{4},
\end{align*}
and subsequently
\begin{align*}
    \|\mathrnd{e}\|_2\leq \varepsilon +  C\sqrt{ \frac{\kappa^2 n}{p}}. 
\end{align*}
It only remains to see that when $C p^{-1/2} \le \ell_\text{min}/\pi$ and
$\kappa \geq \frac{\pi}{\ell_{\min}}\log \frac{4\sqrt{n}}{\varepsilon},$ condition \eqref{cond:kappa} is met. 
\end{proof}

\section{Minor Distribution Shift of Wide Neural Networks}\label{sec:MinorSh}
\begin{figure}[h!]
\centering
\begin{overpic}[trim={0.4cm .4cm  0.4cm 0.4cm},clip,height=3.77in]{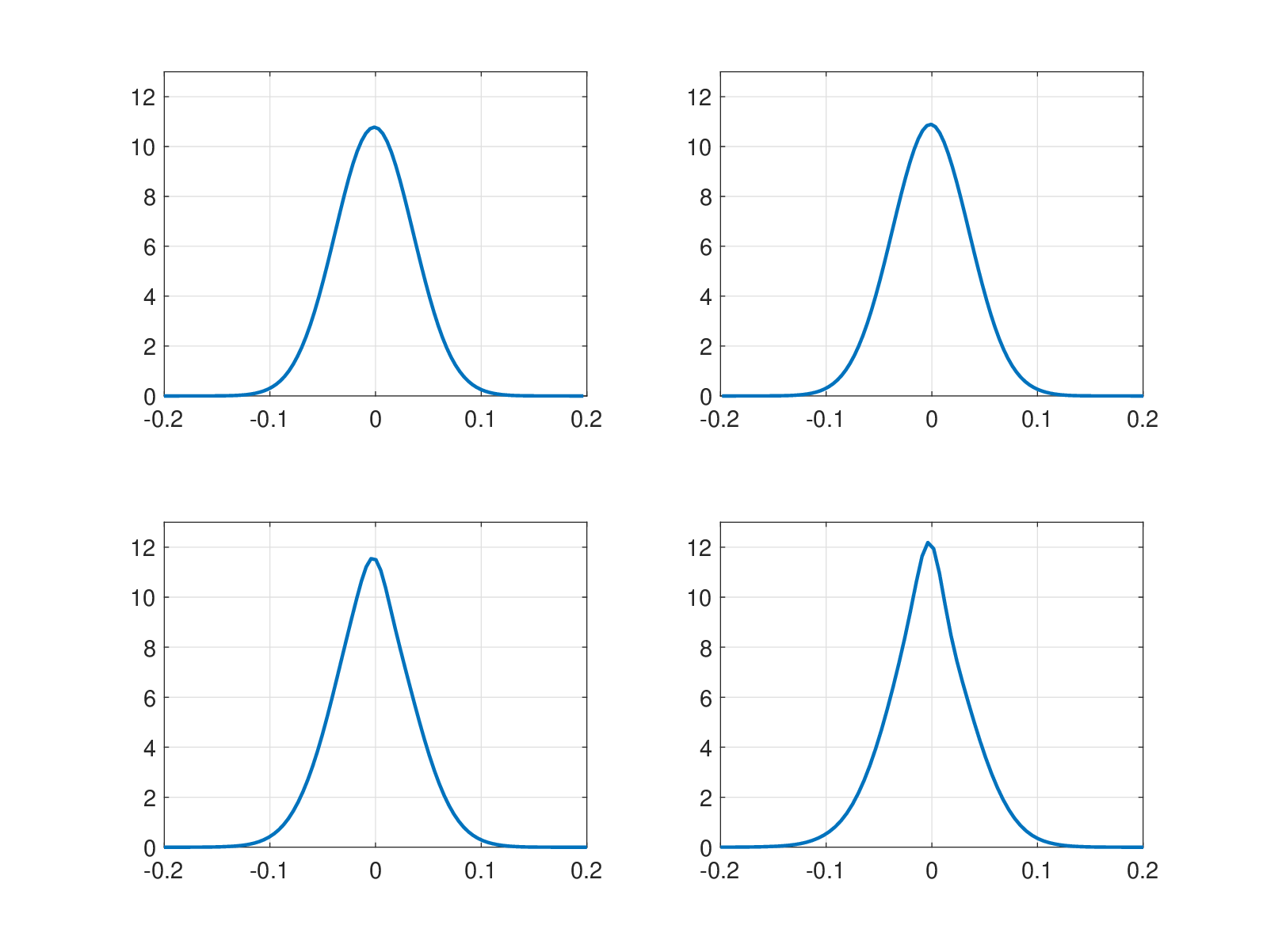}
\put (25,38) {\scalebox{.7}{\rotatebox{0}{Initialization}}}
\put (71,38) {\scalebox{.7}{\rotatebox{0}{1 epoch}}}
\put (25,0) {\scalebox{.7}{\rotatebox{0}{10 epochs}}}
\put (71,0) {\scalebox{.7}{\rotatebox{0}{25 epochs}}}
\end{overpic}
\caption{Minor weight distribution shift for wide layers: the distribution of the weights of a fully connected G-Net Layer at epochs $0$, $1$, $10$, and $25$ }
\label{fig:minorshift}
\end{figure}
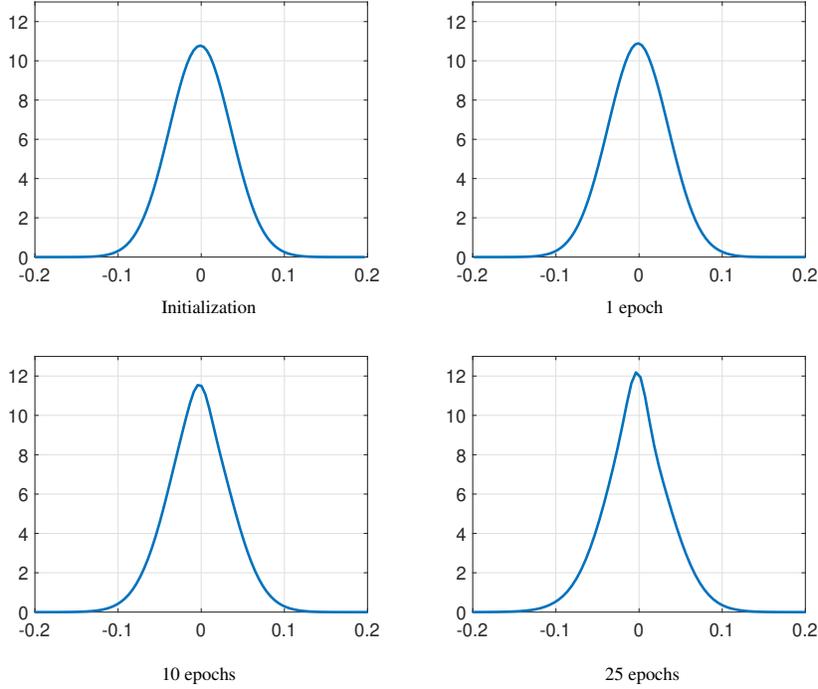

It is well-known that in wide neural networks, the weight distribution shifts only slightly over the course of training \cite{jacot2018neural}. In this section, we present a numerical experiment to demonstrate this property. For this purpose, we train a fully connected G-Net on MNIST data with a wide first layer (width 
1024) initialized using row-normalized Gaussian weights. During training, we monitor the evolution of the layer’s weights and report their empirical distributions at epochs 
0, 1, 10, and 25, as shown in Figure~\ref{fig:minorshift}.

\section{~~G-Net Implementation Details} \label{sec:GNetImp}
In this section, we discuss key implementation details of G-Nets, including the handling of fully connected layers, convolutional layers, general linear layers, and classification versus regression output layers. We also present alternative strategies for incorporating bias in linear layers to enhance the concentration properties of the EHD G-Net.

\paragraph{Fully-Connected Layers.} As discussed in the paper, when $W\in\R^{n\times p}$ maintains $\ell_2$-normalized rows, $x\in\eS^{p-1}$ and $\mathrnd{G}\in\R^{N\times p}$ is a Gaussian matrix, then by Grothendieck's identity and the law of large numbers:
\begin{equation}\label{EHDInf:eq}
\frac{1}{N} \texttt{sign}\left(W \mathrnd{G}^\top\right) \texttt{sign}\left(\mathrnd{G} x\right) 
\xlongrightarrow[\text{ $N\to \infty$ }]{}
\frac{2}{\pi} \texttt{arcsin} (W x).
\end{equation}
For finite $N$, we refer to the left-hand side of \eqref{EHDInf:eq} as the EHD representation of the right-hand side expression. When $W$ and $x$ do not maintain the normalization property, the right-hand side argument needs to be normalized properly. More specifically, consider general $W\in\R^{n\times p}$, $x\in\R^p$, and a Gaussian matrix $\mathrnd{G}\in\R^{N\times p}$, structured as follows:
\[
W = \begin{pmatrix}
    w_1^\top \\ \vdots \\w_n^\top
\end{pmatrix}, \qquad \mathrnd{G} = \begin{pmatrix}
    \mathrnd{g}_1^\top \\ \vdots \\ \mathrnd{g}_N^\top
\end{pmatrix}.
\]
Then we can construct a diagonal matrix to perform the normalization on $Wx$ as 
\begin{equation}\label{EHD:canonic}
   \frac{2}{\pi} \texttt{arcsin}\left( D_{W,x}^{-\frac{1}{2}}Wx\right) \xlongrightarrow[\text{    }]{\text{EHD}} \frac{1}{N}\sum_{i=1}^N\sign\left( \left\langle \mathrnd{g}_i,x\right\rangle \right)\texttt{sign}\left( W\mathrnd{g}_i\right),
\end{equation}
where 
\begin{equation}\label{dwx:eq}
D_{W,x} = \|x\|_2^2~\text{diag}\left(\|w_1\|_2^2,\ldots, \|w_n\|_2^2 \right),
\end{equation}
and $\text{diag}(\alpha_1,\ldots,\alpha_n)$ is a diagonal matrix with elements $\alpha_1,\ldots,\alpha_n$. Basically, the left-hand side expression of \eqref{EHD:canonic} is what is implemented in a fully-connected layer of G-Net.

\paragraph{Smooth Approximation of the Norm.} When training a G-Net, the rows of $W$ need to be normalized by their $\ell_2$ norm. Basically, the $i$-th row of $W$ takes the form $w_i^\top/\|w_i\|_2$. To maintain a bounded gradient flow during the training, we use the smooth approximation 
\[
\|w_i\|_2 \approx \sqrt{\|w_i\|_2^2 + \varepsilon},
\]
where $\varepsilon>0$ is a small constant. 
\paragraph{A Different Handling of the Bias for Faster Concentration.}
For an input $x\in\R^p$, the output of a fully-connected layer is normally in the form 
$Wx + b$ where $W\in\R^{n\times p}$ and $b\in\R^n$ are learnable weight and bias parameters. One could think of absorbing the bias in the weight matrix as
\[
Wx+b = \tilde W \tilde x,\quad \mbox{where}: \tilde W = \begin{bmatrix}
    W& b
\end{bmatrix},~~ \tilde x=\begin{bmatrix}
    x\\ 1
\end{bmatrix},
\]
and view the problem as an instance of the aforementioned fully-connected case. However, as discussed in the theoretical results, especially for Rademacher embedding, having the weights evenly spread out helps with a better concentration of EHD G-Net. Since the scales of $b$ and $W$ after training can differ significantly, an alternative approach to introducing a bias term in linear layers---aimed at promoting a more balanced distribution of weights---is to use a fully connected layer of the form
\[
W(x + c 1_p),
\]
where $W \in \mathbb{R}^{n \times p}$ and $c \in \mathbb{R}$ is a learnable scalar. By trading off a negligible degree of freedom, this formulation eliminates the issue of uneven scaling between weight and bias parameters. This idea naturally extends to other linear layers and simply involves adding a learnable constant to the input.

\paragraph{General Linear Layers.} The implementation scheme described above for fully connected layers can be extended to general linear layers, which may include layers such as batch normalization, average pooling, or cascades of multiple linear transformations. Consider a more general linear operator \(\mathcal{T}_W : \mathcal{V} \rightarrow \mathbb{R}^n\), where \(\mathcal{V} \subseteq \mathbb{R}^{p_1 \times p_2 \times \ldots \times p_v}\), and \(W\) denotes the parameters associated with the operator \(\mathcal{T}_W\). Let \(e_j\) denote the \(j\)-th canonical basis vector in \(\mathbb{R}^n\). Then the \(j\)-th component of the output of \(\mathcal{T}_W\) can be expressed as
\begin{equation}\label{eq:adjGen}
\left\langle \mathcal{T}_W(X), e_j \right\rangle = \left\langle X, \mathcal{T}_W^*(e_j) \right\rangle,
\end{equation}
where \(\mathcal{T}_W^*\) is the Hermitian adjoint of \(\mathcal{T}_W\). According to \eqref{eq:adjGen}, one can interpret \(\|\mathcal{T}_W^*(e_j)\|_{HS}\), the Hilbert–Schmidt norm of \(\mathcal{T}_W^*(e_j)\), as the norm of the \(j\)-th ``row'' of the operator \(\mathcal{T}_W\). Letting \(\mathrnd{G}_i\), for \(i = 1, \ldots, N\), be i.i.d. Gaussian tensors of the same shape as \(X\), a generalization of \eqref{EHD:canonic} takes the form
\begin{equation}\label{tensorForm:eq}
   \frac{2}{\pi} \texttt{arcsin}\left( d_{W,X}^{\odot-1}\odot \mathcal{T}_W(X)\right) \xlongrightarrow[\text{    }]{\text{EHD}} \frac{1}{N}\sum_{i=1}^N\sign\left( \left\langle \mathrnd{G}_i,X\right\rangle \right)\texttt{sign}\left( \mathcal{T}_W\left(\mathrnd{G}_i\right)\right),
\end{equation}
where $\odot$ denotes a Hadamard product,  $d_{W,X}^{\odot-1}$ denotes the Hadamard inverse\footnote{The Hadamard inverse of a vector $d$ is acquired by inverting each element of $d$, i.e., $[d^{\odot-1}]_j = 1/d_j$.} of the vector $d_{W,X}$, and  \[
d_{W,X} = \|x\|_{HS}\begin{pmatrix}
    \left\|\mathcal{T}_W^*(e_1)\right\|_{HS}\\ \vdots \\ \left\|\mathcal{T}_W^*(e_n)\right\|_{HS}
\end{pmatrix}.
\]
Equation~\eqref{tensorForm:eq} shows that, once the Hermitian adjoint of a linear operator is available, normalizing the operator and constructing the hyperdimensional embedding become straightforward tasks. For a simpler notation of the canonical basis, the range of $\mathcal{T}_W$ is considered to be $\R^n$.
However, this framework readily generalizes to multi-dimensional output arrays, in which case \(d_{W,X}\) is replaced by \(D_{W,X}\), a tensor with the same dimensions as \(\mathcal{T}_W(X)\).

\paragraph{Convolutional Layers.} Although convolutional layers are linear and can, in principle, be addressed using the formulation from the previous section, we introduce a specialized approach tailored for a more compact representation and improved concentration properties. For notational simplicity, we focus on one-dimensional convolution with a single output channel; the generalization to higher dimensions and multiple channels is straightforward. Consider $x\in\R^p$ as an input array, $w\in\R^n$ a convolutional filter, and
let $*$ denote a vector-wise convolution (or cross-correlation). The convolution $w*x$ is a vector of length $m$, where $m$ depends on input parameters such as $p$, $n$, padding, and stride. Each vector-wise convolution \( w * x \) can be expressed as a matrix-vector product:
\begin{equation}\label{wsx:eq}
w*x =\begin{pmatrix}
    w^\top  P_1^\top \\ \vdots\\ w^\top P_m^\top  
\end{pmatrix}x,
\end{equation}
where $P_k\in\{0,1\}^{p\times n}$, $k\in\{1,\ldots,m\}$, circularly shift the elements of $w$. While the explicit construction of \( P_k \) is not required, the row norms of the matrix in \eqref{wsx:eq} can be efficiently obtained via a simple convolution operation as follows:
\[
\begin{pmatrix}
      \|P_1 w\|_2^2 \\ \vdots\\ \|P_m w\|_2^2 
\end{pmatrix} = (w\odot w)*1_p,
\]
where $1_p$ denotes a length-\( p \) vector of all ones. Following the base equation in \eqref{EHD:canonic}, for $\mathrnd{g}_i\sim\mathcal{N}(0,I_p)$, the embedding takes the following form
\begin{equation}\label{convwx:eq}
   \frac{2}{\pi} \texttt{arcsin}\left( D_{w,x}^{-\frac{1}{2}}(w*x)\right) \xlongrightarrow[\text{    }]{\text{EHD}} \frac{1}{N}\sum_{i=1}^N\sign\left( \left\langle \mathrnd{g}_i,x\right\rangle \right)\texttt{sign}\left( w*\mathrnd{g}_i\right),
\end{equation}
where 
\[
D_{w,x} = \|x\|_2^2~\text{diag}\left((w\odot w)*1_p \right).
\]
One can use the commutative property of the convolution and write 
\begin{equation*}
w * x = x*w  = \begin{pmatrix}
    x^\top  P_1\\ \vdots\\ x^\top  P_m 
\end{pmatrix}w,
\end{equation*}
which for $\mathrnd{g}_i\sim\mathcal{N}(0,I_n)$ offers the following alternative embedding scheme
\begin{equation}\label{convxw:eq}
   \frac{2}{\pi} \texttt{arcsin}\left( D_{w,x}^{-\frac{1}{2}}(w*x)\right) \xlongrightarrow[\text{    }]{\text{EHD}} \frac{1}{N}\sum_{i=1}^N\sign\left( \left\langle \mathrnd{g}_i,w\right\rangle \right)\texttt{sign}\left( x*\mathrnd{g}_i\right),
\end{equation}
where 
\[
D_{w,x} = \|w\|^2_2~\text{diag}\left((x\odot x)*1_n \right).
\]
In convolutional layers, it is typically the case that \( n < p \), which makes \eqref{convxw:eq} preferable over \eqref{convwx:eq} due to its faster concentration with increasing \( N \).

For multi-channel input signals, consider an input vector of length $p$ with $n_c$ channels, and learnable weights of length $n$ corresponding to each input channel. We use the notation 
\[
X = \begin{bmatrix}
    x_1,\ldots, x_{n_c} 
\end{bmatrix}\in\R^{p\times n_c}, \qquad W = \begin{bmatrix}
    w_1,\ldots, w_{n_c} 
\end{bmatrix}\in\R^{n\times n_c}.
\]
In this case, the output of the convolutional layer is a vector of length $m$ which is computed via 
\[
W\circledast X = \sum_{j = 1}^{n_c}  w_j * x_j.
\]
For independent Gaussian matrices \(\mathrnd{G}_i\in\R^{n\times n_c}\), a generalization of \eqref{convxw:eq} to the multi-channel input case is offered through
\begin{equation}
   \frac{2}{\pi} \texttt{arcsin}\left( D_{W,X}^{-\frac{1}{2}}\left(W\circledast X \right)\right) \xlongrightarrow[\text{    }]{\text{EHD}} \frac{1}{N}\sum_{i=1}^N\sign\left( \left\langle \mathrnd{G}_i,W\right\rangle \right)\texttt{sign}\left( X\circledast\mathrnd{G}_i\right),
\end{equation}
where 
\begin{align*}
D_{W,X} &= \left(\sum_{j=1}^{n_c}\|w_j\|^2_2\right)~\text{diag}\left(\left(\sum_{j=1}^{n_c}x_j\odot x_j\right)*1_n \right)\\& = \|W\|_F^2~\text{diag}\left(\left(X\odot X\right)\circledast 1_{n\times n_c} \right).
\end{align*}

\paragraph{Network Output Layer.} In this section, we describe how to configure the network’s output layer to function either as a regressor or a classifier. 

For regression, we can model the final layer as
\[
y_L \;=\; \texttt{ASU}\left( D_{W_L,y_{L-1}}^{-\tfrac12}\, W_L y_{L-1}\right),
\]
where \(D_{W_L,y_{L-1}}\) is given by \eqref{dwx:eq}, and the subsequent hyper-dimensional embedding follows the standard procedure.  
To employ this structure, the only requirement is to rescale the response variable to the interval \(\bigl[-\tfrac{2}{\pi},\, \tfrac{2}{\pi}\bigr]\) so that its range is consistent with \(y_L\) above.

For classification G-Nets, we can still use a standard soft-max activation as
\begin{equation}\label{softmax:eq}
y_L \;=\; \texttt{SoftMax}\left( D_{W_L,y_{L-1}}^{-\tfrac12}\, W_L y_{L-1}\right).
\end{equation}
Since the \texttt{ASU} function is monotonic and preserves the position of the dominant component in an array, it is omitted from \eqref{softmax:eq}, yet a standard hyperdimensional embedding as the right-hand side of \eqref{dwx:eq} can be used. The EHD G-Net predicted label is determined by identifying the index of the maximum absolute component in the output vector.

\section{~~Extended Numerical Experiments}\label{sec:NumericalExp}
In this section, we conduct numerical experiments on several datasets commonly used as benchmarks in HDC classification tasks \cite{verges2025classification}, focusing on more challenging vision datasets such as MNIST, FashionMNIST, and CIFAR10. We also include two human-activity recognition datasets, HAR-WSS (walking, sitting, standing)~\cite{WSSDataLink} and Epilepsy~\cite{EpilepsyDataLink}; an automotive dataset, Ford-A~\cite{FordADataLink}; a natural-language dataset, AG News \cite{AGNewsDataLink}; and a time-series dataset, Fault Detection-A \cite{FaultDetectionADataLink}.
A key characteristic of all datasets selected for evaluation is that, unlike simpler HDC datasets such as European Languages~\cite{joshi2016language} and ISOLET~\cite{cole1990isolet}, which achieve high accuracy with linear classifiers like support vector machines (SVMs), the chosen benchmarks are not easily addressed by such models. In fact, many HDC methods struggle to perform well on these datasets. To assess the effectiveness of the proposed framework, we compare its performance against several established and recent methods, including classic HDC~\cite{Ge2020Classification}, HoloGN~\cite{ExHoloGN}, Laplace HDC~\cite{LaplaceHDC}, OnlineHD~\cite{onlineHD2021}, and RFF-HDC~\cite{yu2022understanding}.

Table~\ref{tab:summary} summarizes the datasets used, including the number of input features (\(n_0\)), output classes (\(n_L\)), and the corresponding G-Net architectures. We aimed to vary the layer configurations across datasets while maintaining simplicity and reproducibility. For instance, on the FashionMNIST dataset, we employed a single convolutional layer to stress-test G-Net under a shallower architecture. A demo of G-Net training/EHD conversion, along with all experiments reported in this section, is available at \url{https://github.com/GNet2025/GNet}.

\begin{table}[ht!]
\caption{Dataset and G-Net architecture specifications. In layer descriptions: CN denotes a convolutional layer (input channels, filters, output channels), EM an embedding layer (vocabulary size, number of outputs), FC a fully connected layer (number of outputs), and CL a classification layer (number of classes)}\label{tab:summary}
\setlength{\tabcolsep}{2pt}   
\renewcommand{\arraystretch}{0.9}   
\begin{center}
\tiny
\begin{NiceTabular}[corners=NW]{|c|c|c|c|c|c|c|c|}\toprule 
           \scriptsize \textbf{MNIST} &   \scriptsize \textbf{FashionMNIST} &\scriptsize \textbf{Ford-A} &  \scriptsize \textbf{WSS} & \scriptsize \textbf{Epilepsy} &  \scriptsize \textbf{CIFAR10} & \scriptsize \textbf{AG News} & \scriptsize \textbf{Fault Detection-A}\\\midrule 
           \Block{1-8}{\centering \scriptsize Data Specifications}\\\midrule 
             $n_0 = 28\!\times\! 28$ & $n_0 = 28\!\times\! 28$ & $n_0 = 500$ & $n_0 = 206\!\times\! 3$ & $n_0 = 206\!\times\! 3$ & $n_0 = 32\!\times\! 32\times\! 3$ & $n_0 = (400, 512)$ & $n_0 = 5120$ \\
             $n_L=10$ & $n_L=10$ & $n_L=2$ & $n_L=6$ & $n_L=4$ & $n_L=10$ &  $n_L=4$ &  $n_L=3$  \\\midrule 
            \Block{1-8}{\centering \scriptsize G-Net Specifications}\\\midrule 
             CN(1,32,3) & CN(1,32,5) & CN(1,16,15) & CN(3,64,11) & CN(3,64,11) & CN(3,32,5) & EM(44120,512)& CN(1, 3, 64, 9) 
            \\
             CN(32,64,5) & FC(512) & CN(16,16,15) & CN(64,48,7) & CN(64,48,7) & CN(32,64,3) & CN(512,64,\{5,6,7\})$\times$ 2 & CN(64, 48, 9)
            \\
             FC(512) & CL(10) & CN(16,25,13) & CL(6) & CL(4) & FC(512) & CL(4) & CL(3)
            \\
             CL(10) &  & CL(2) &  &  & CL(10)\\\bottomrule
\end{NiceTabular}\normalsize
\end{center}
\end{table}

The first set of experiments involves fitting a G-Net to the original training data, followed by evaluation in the binary hyperspace by applying the corresponding EHD G-Net to the hyperdimensional embedding of test data. The G-Net training is quick---about ten epochs on MNIST require roughly 48 seconds on a desktop computer equipped with a GeForce RTX 4090 GPU. The EHD conversion of the same G-Net takes less than 0.5 seconds, and the inference time for each embedded sample is in the order of milliseconds. 
\begin{figure}[!htbp]
\begin{center}
\begin{overpic}[trim={0.17cm -.25cm  0.25cm 0},clip,height=1.77in]{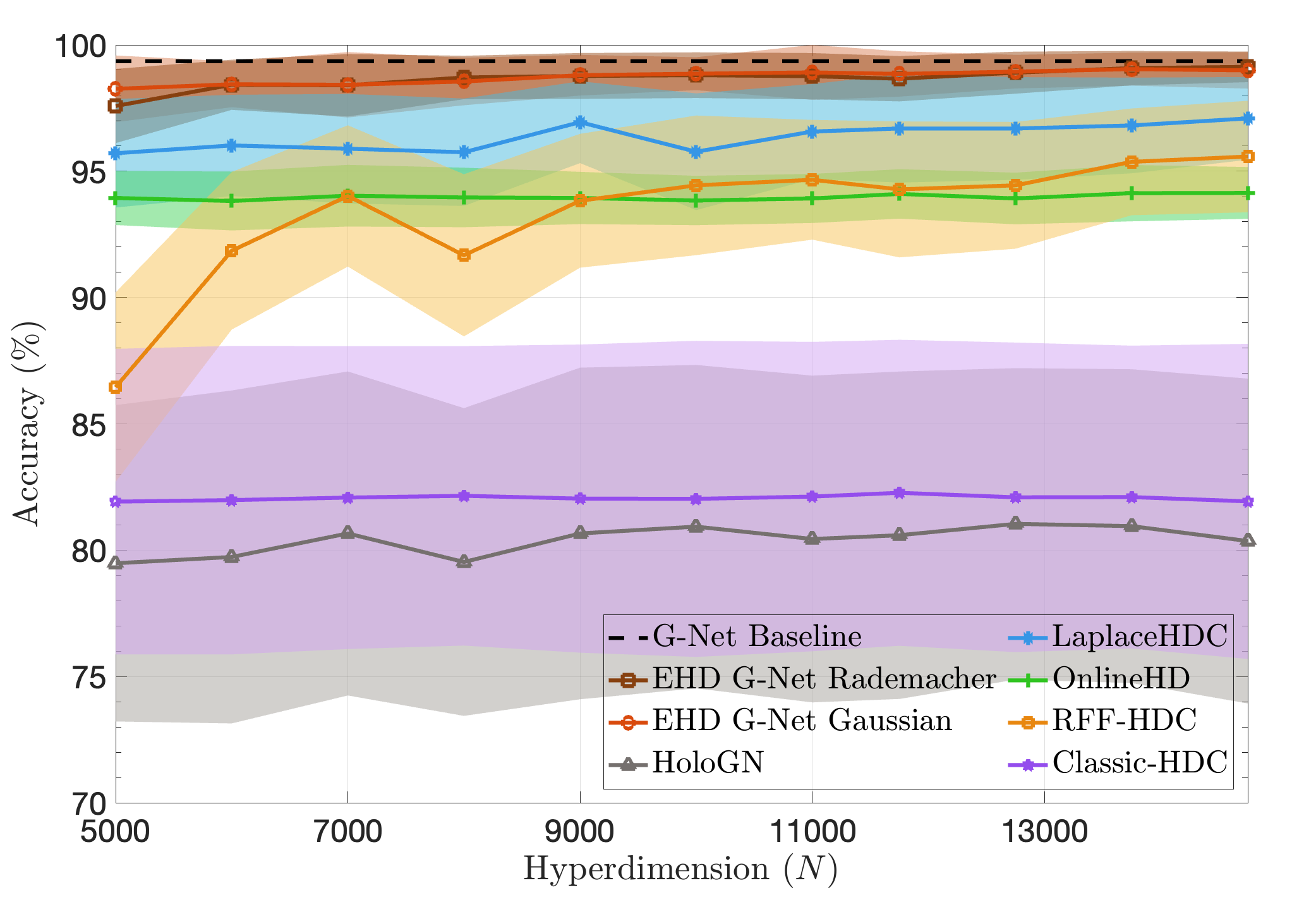}
\put (50,-3) {\scalebox{.7}{\rotatebox{0}{(a)}}}
\end{overpic}
\begin{overpic}[trim={0.17cm -.25cm  0.25cm 0},clip,height=1.77in]{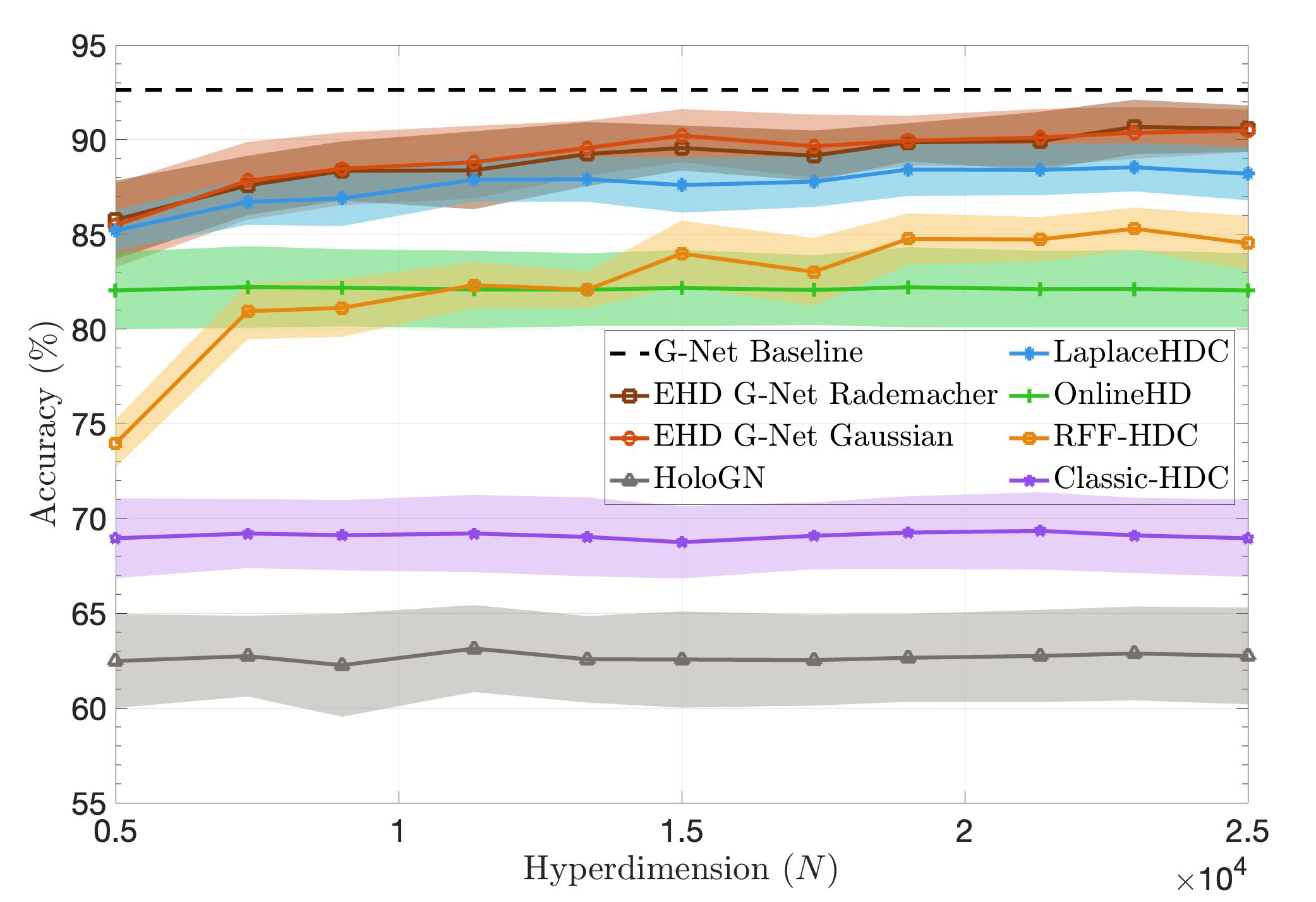}
\put (50,-3) {\scalebox{.7}{\rotatebox{0}{(b)}}}
\end{overpic}\vspace{.2cm}
\begin{overpic}[trim={0.17cm -.25cm  0.25cm 0},clip,height=1.77in]{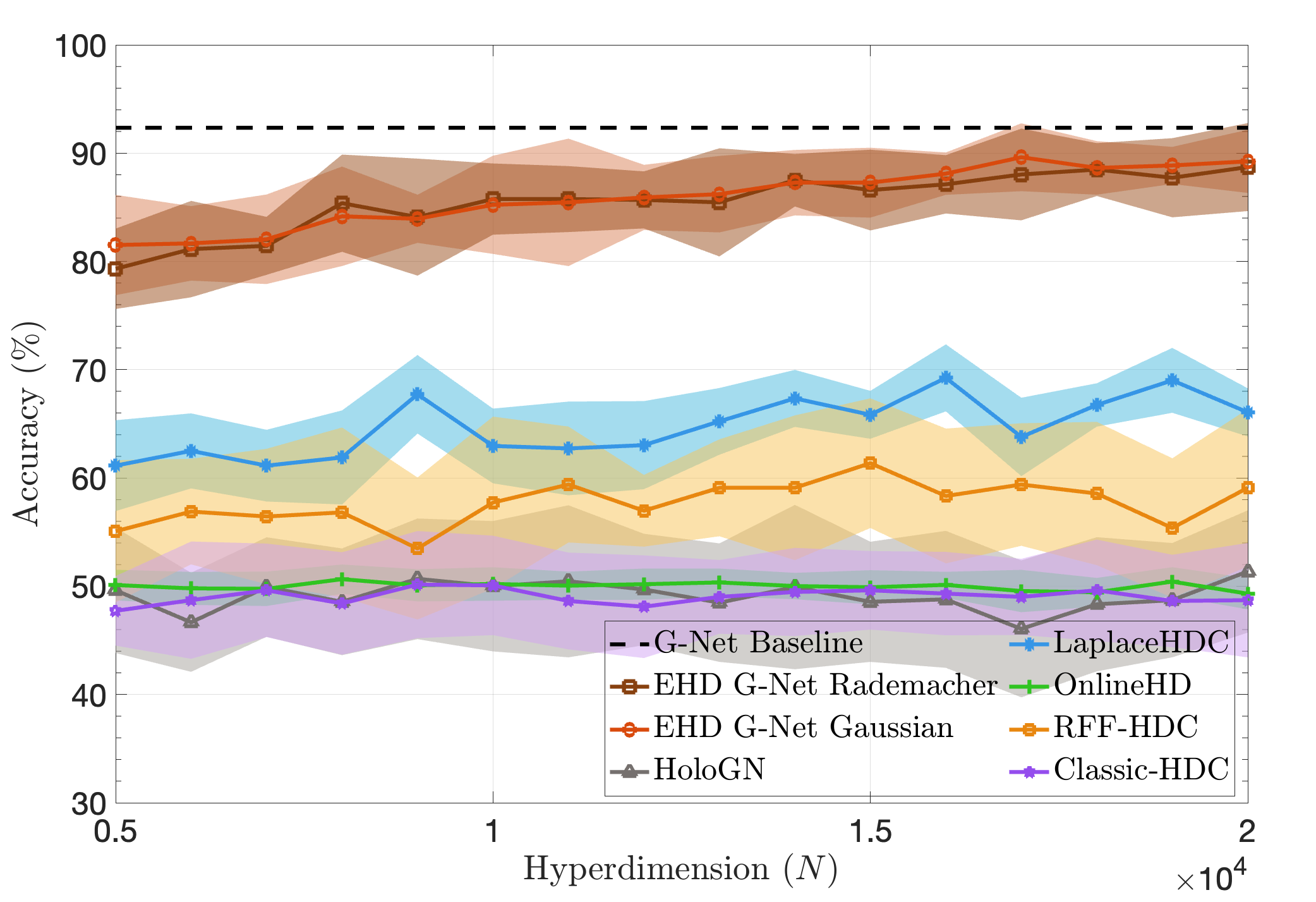}
\put (50,-3) {\scalebox{.7}{\rotatebox{0}{(c)}}}
\end{overpic}
\begin{overpic}[trim={0.17cm -.25cm  0.25cm 0},clip,height=1.77in]{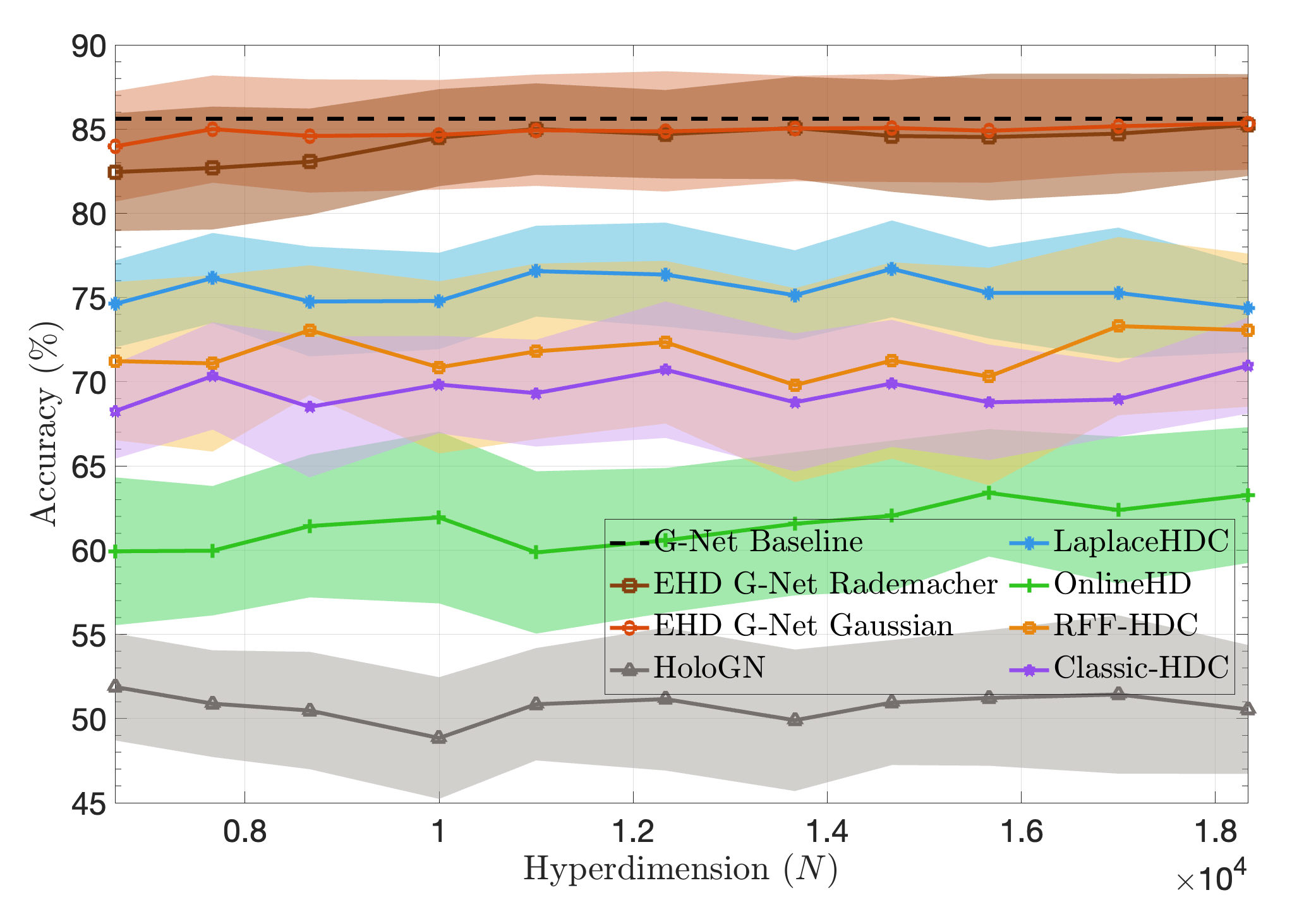}
\put (50,-3) {\scalebox{.7}{\rotatebox{0}{(d)}}}
\end{overpic}\vspace{.2cm}
\begin{overpic}[trim={0.17cm -.25cm  0.25cm 0},clip,height=1.77in]{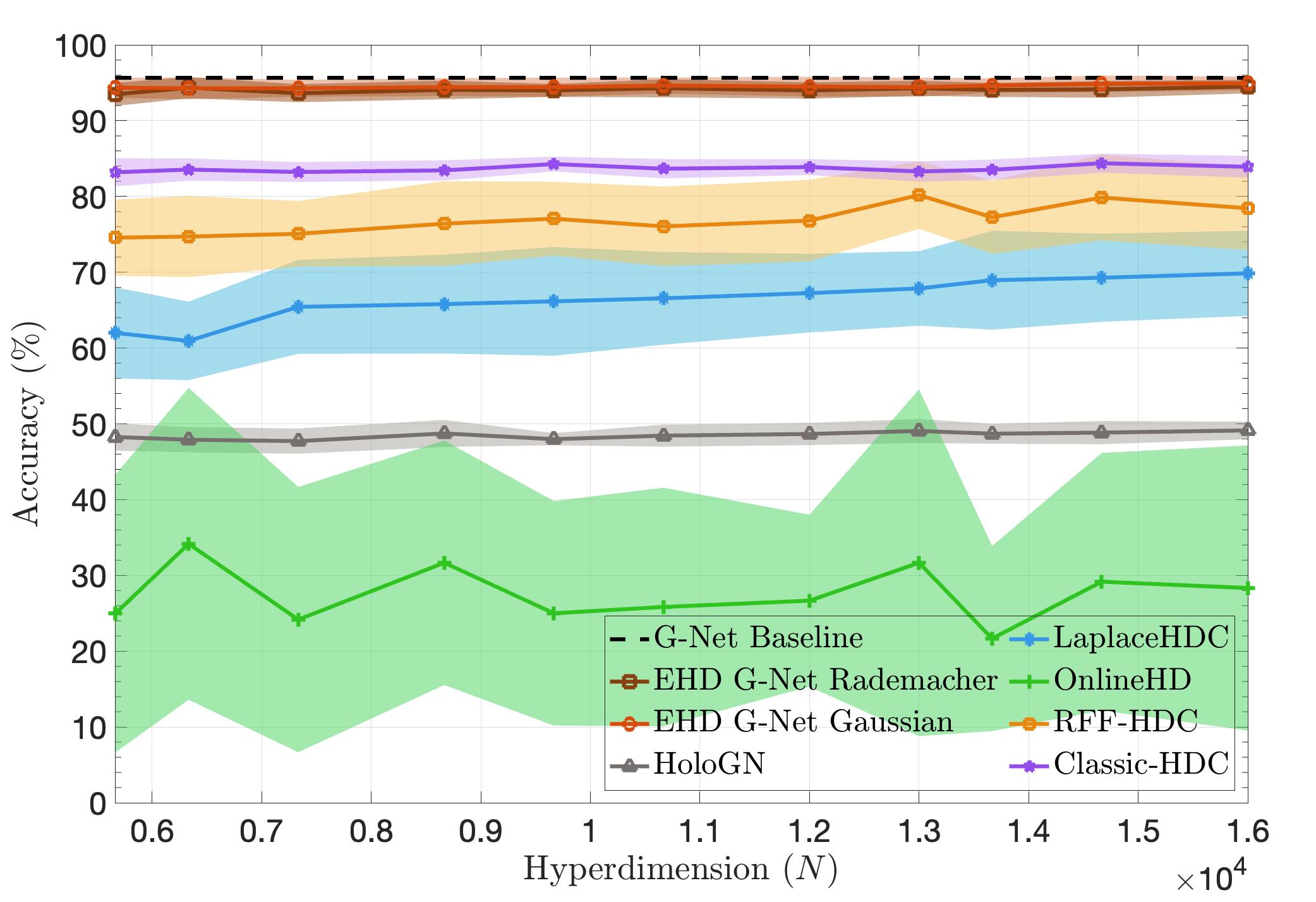}
\put (50,-3) {\scalebox{.7}{\rotatebox{0}{(e)}}}
\end{overpic}
\begin{overpic}[trim={0.17cm -.25cm  0.25cm 0},clip,height=1.77in]{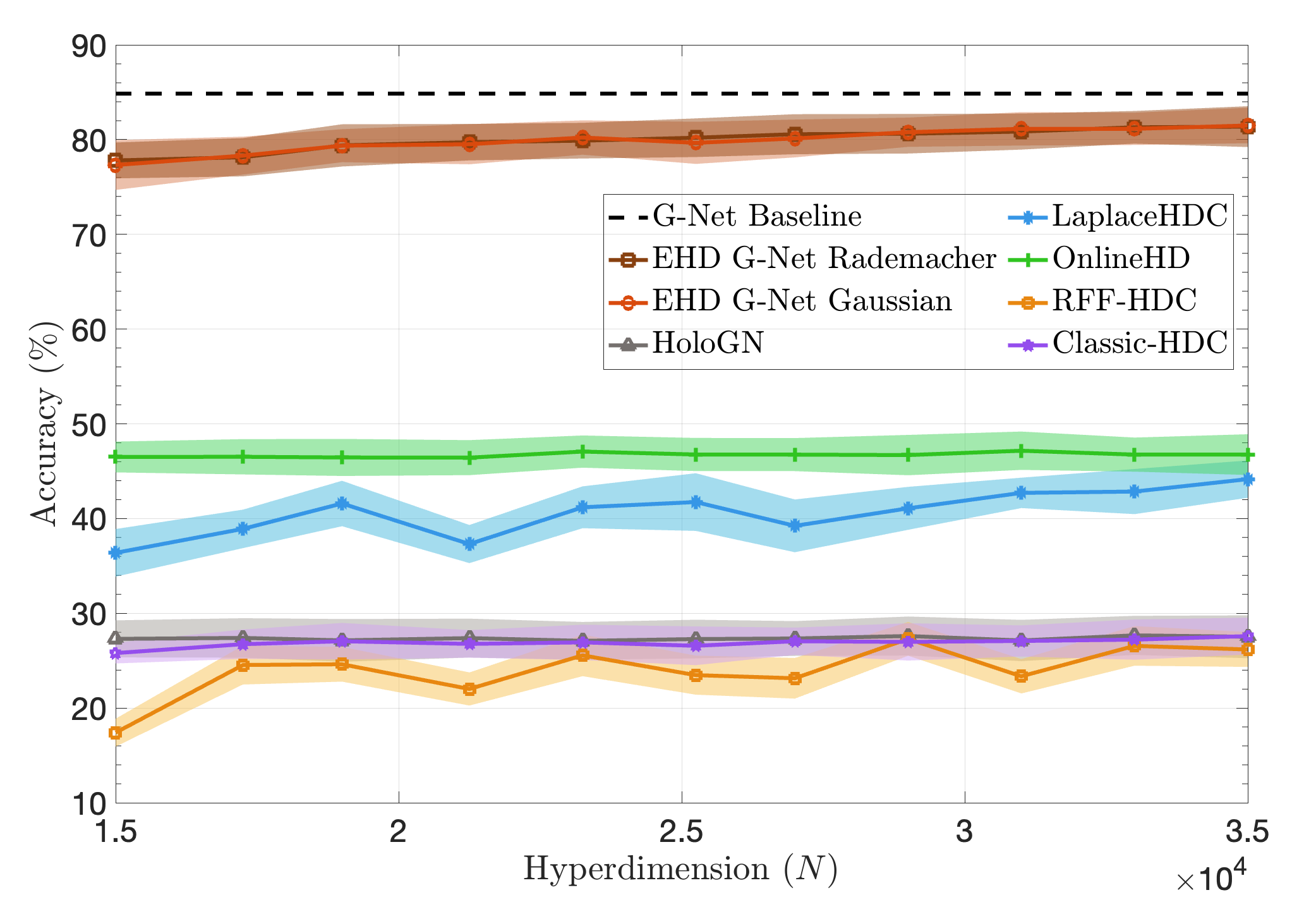}
\put (50,-3) {\scalebox{.7}{\rotatebox{0}{(f)}}}
\end{overpic}
\\[.2cm]
\begin{overpic}[trim={0.17cm -.25cm  0.25cm 0},clip,height=1.77in]{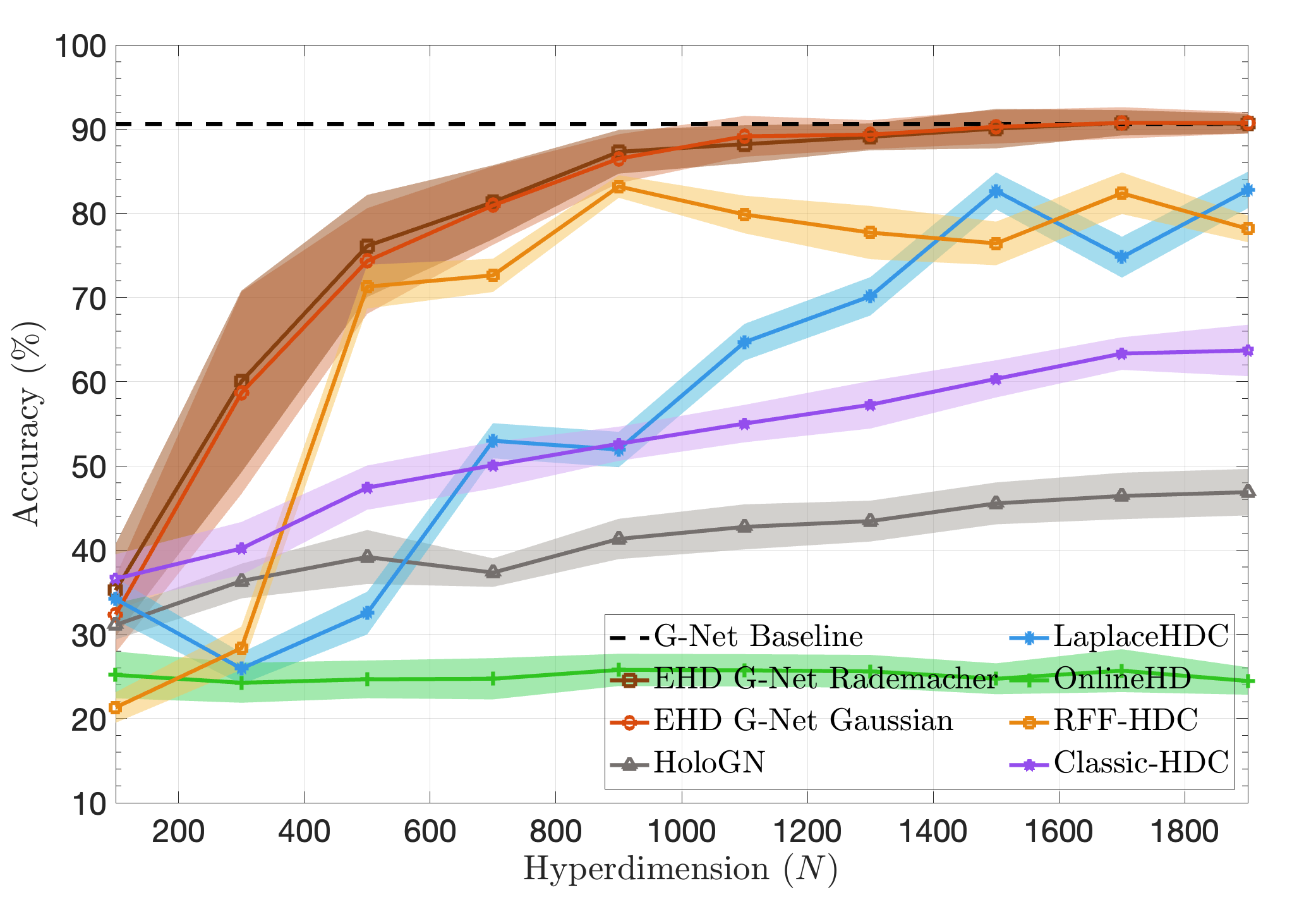}
\put (50,-3) {\scalebox{.7}{\rotatebox{0}{(g)}}}
\end{overpic}
\begin{overpic}[trim={0.17cm -.25cm  0.25cm 0},clip,height=1.77in]{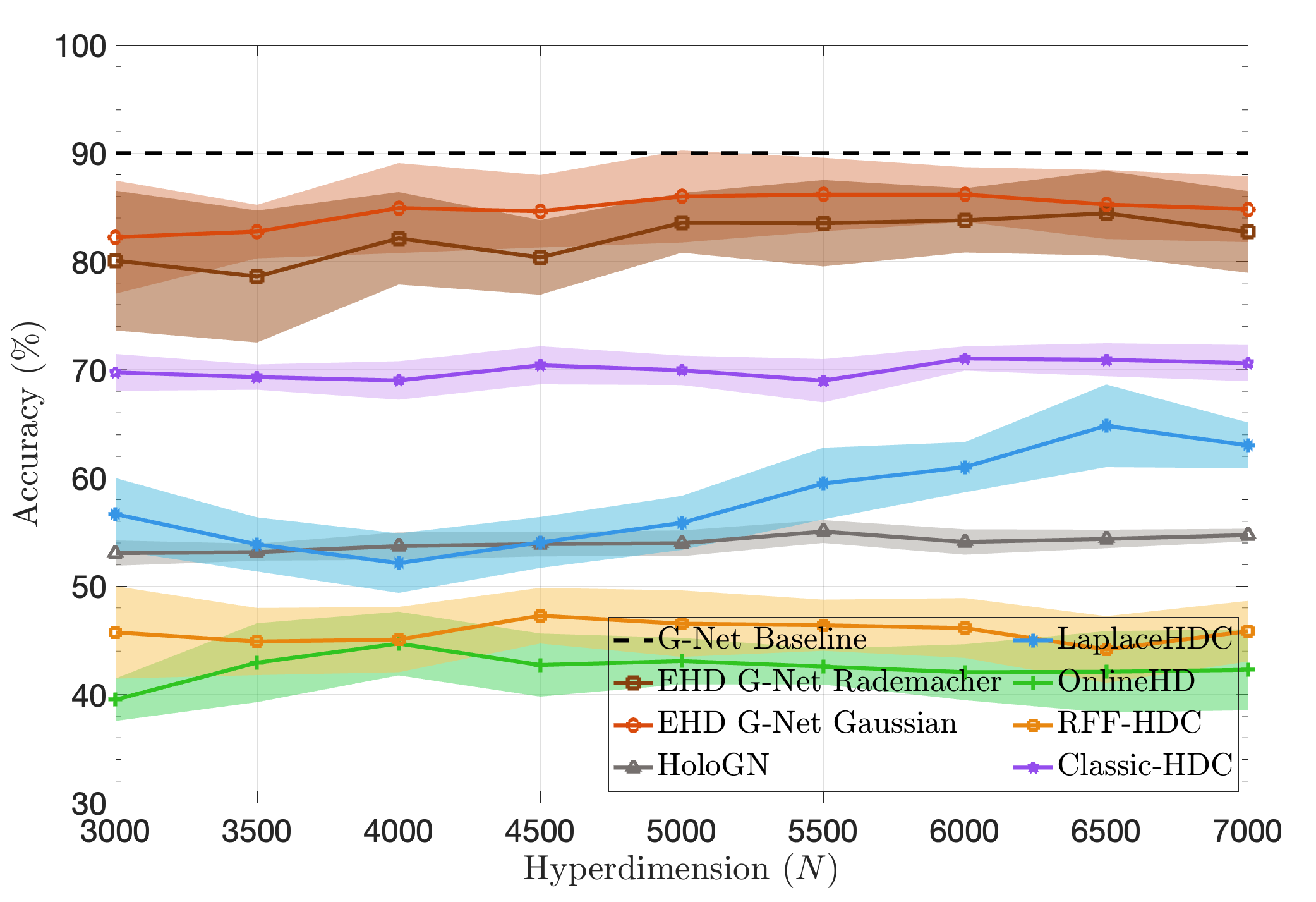}
\put (50,-3) {\scalebox{.7}{\rotatebox{0}{(h)}}}
\end{overpic}
\end{center}\vspace{-.2cm}
\caption{Comparison of Gaussian/Rademacher RASU G-Net with other HDC methods on different datasets (a) MNIST, (b) FashionMNIST, (c) Ford-A, (d) WSS, (e) Epilepsy, (f) CIFAR10, (g) AG News, (h) Fault Detection-A }\label{figS1} \vspace{-.6cm}
\end{figure}
Figure \ref{figS1} reports the average test accuracies of EHD G-Net and other HDC methods. The reported hyperdimension $N$ represents the average $N_\ell$ across G-Net layers and the dimension used in other methods. For each $N$, the conversion of the reference G-Net to an EHD G-Net was repeated multiple times with different random matrices; the same number of repetitions was applied to other HDC techniques. The plots show the resulting mean accuracies along with $\pm 1$ standard deviation.

The G-Net architecture used in Figure \ref{figS1} is a RASU network, and the results, both for Gaussian and Rademacher embeddings are reported. While Gaussian embedding yields slightly better results, the accuracies of the two embedding methods are very close. One observes that without training a large binary model---and by training only a compact network in the primal space followed by inexpensive binary encoding---we achieve classifiers that outperform state-of-the-art HDC models by a significant margin. For example, HDC accuracies exceed 99.2\% on MNIST, 81\% on CIFAR-10, and 91\% on FashionMNIST, rivaling real-valued convolutional neural networks. The reported accuracies can be even further improved by employing larger G-Nets and higher hyperdimensions, as EHD G-Net performance asymptotically approaches that of the original G-Net with increasing $N$.

\begin{figure}[!htbp]
\begin{center}
\begin{overpic}[trim={0.17cm -.25cm  0.25cm 0},clip,height=1.77in]{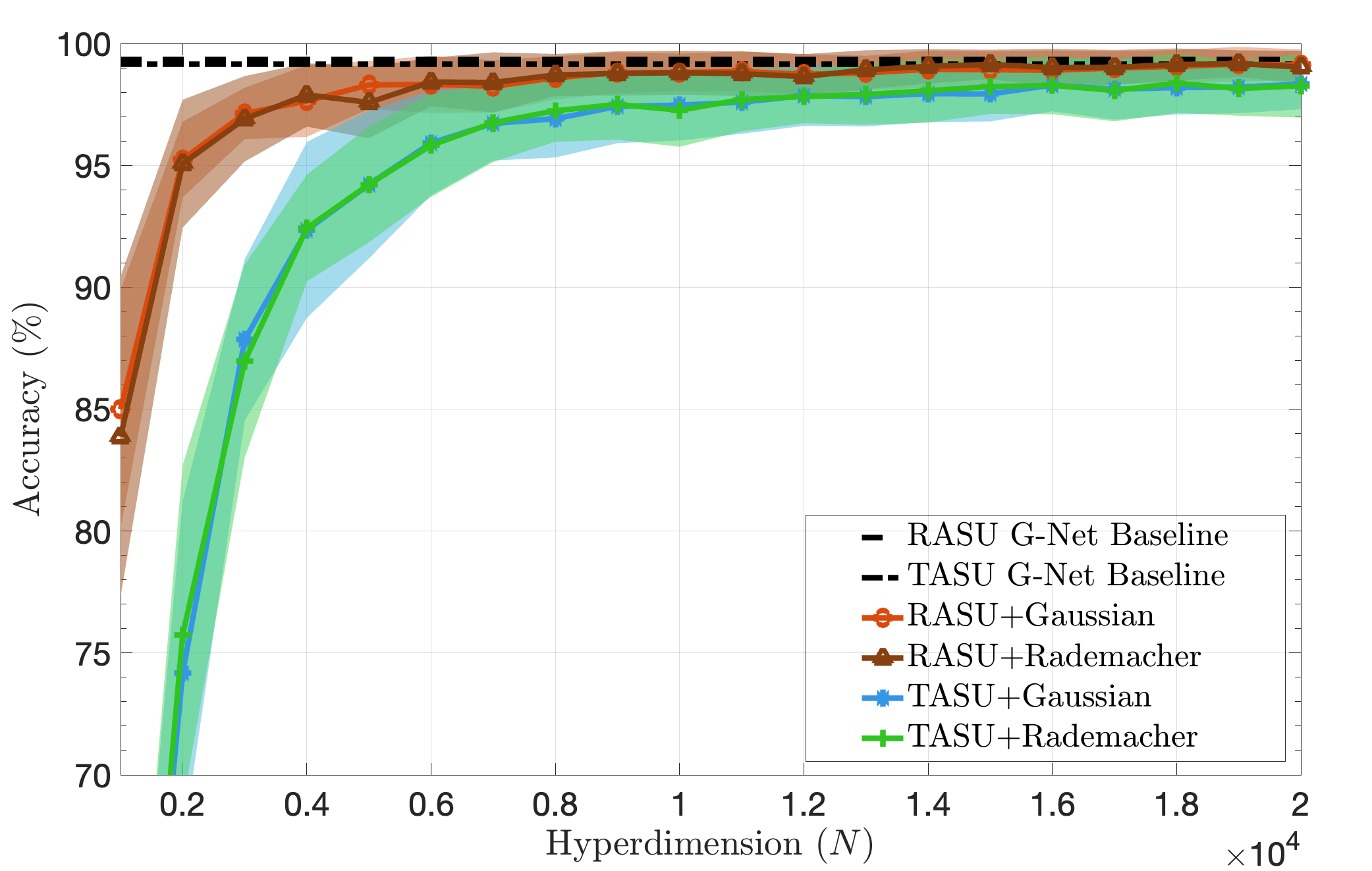}
\put (50,-3) {\scalebox{.7}{\rotatebox{0}{(a)}}}
\end{overpic}
\begin{overpic}[trim={0.17cm -.25cm  0.25cm 0},clip,height=1.77in]{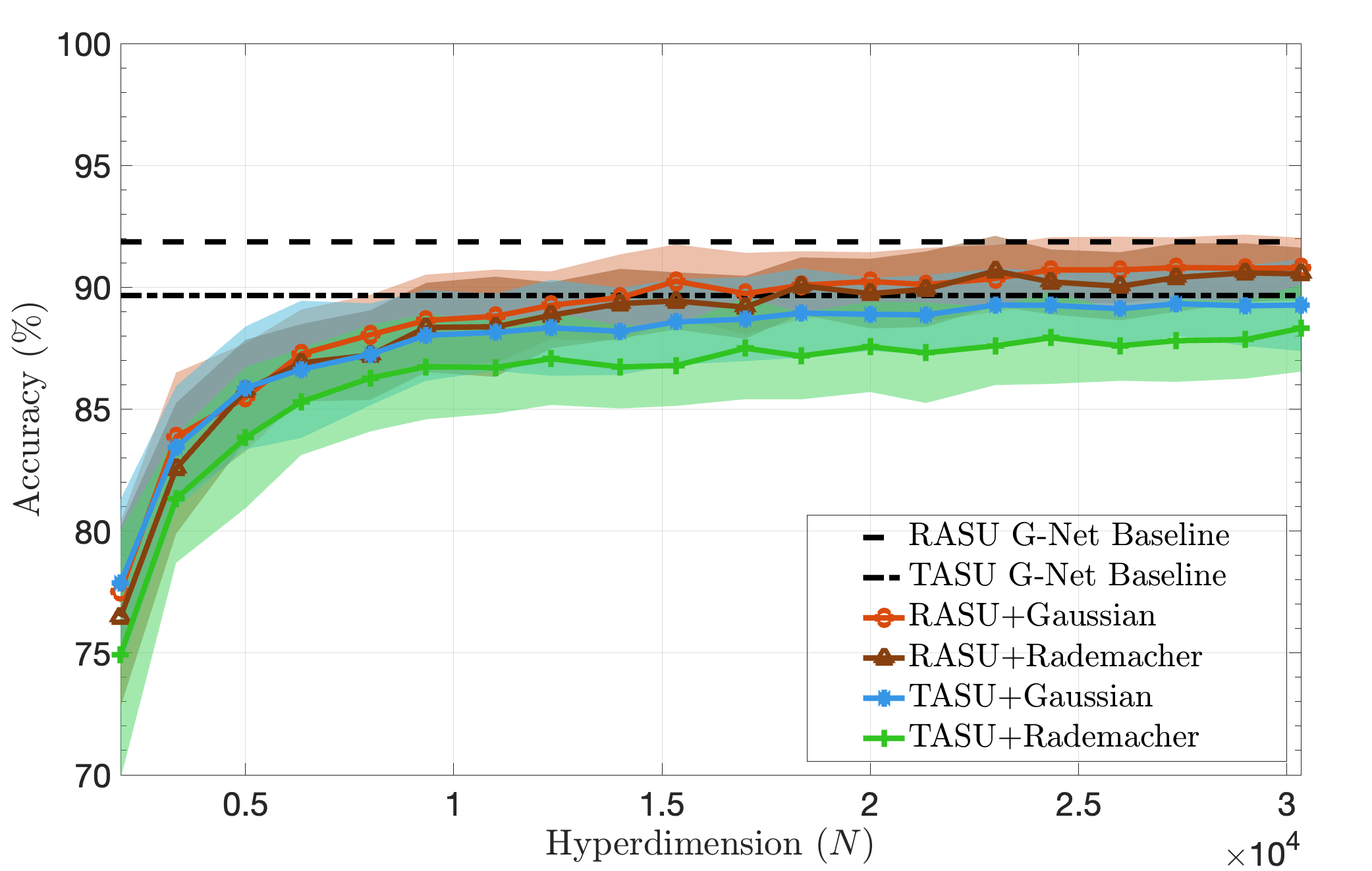}
\put (50,-3) {\scalebox{.7}{\rotatebox{0}{(b)}}}
\end{overpic}\vspace{.42cm}
\begin{overpic}[trim={0.17cm -.25cm  0.25cm 0},clip,height=1.77in]{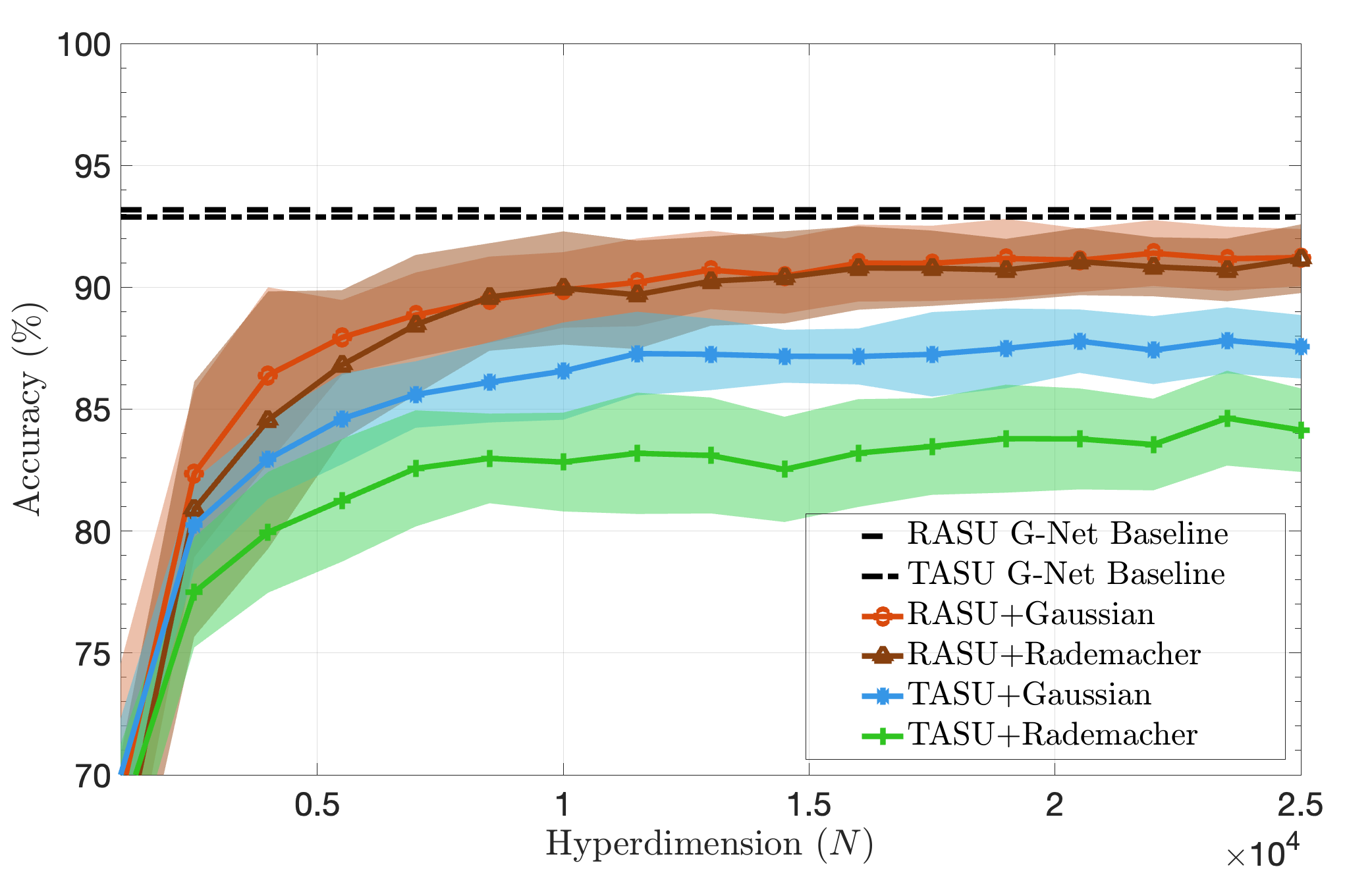}
\put (50,-3) {\scalebox{.7}{\rotatebox{0}{(c)}}}
\end{overpic}
\begin{overpic}[trim={0.17cm -.25cm  0.25cm 0},clip,height=1.77in]{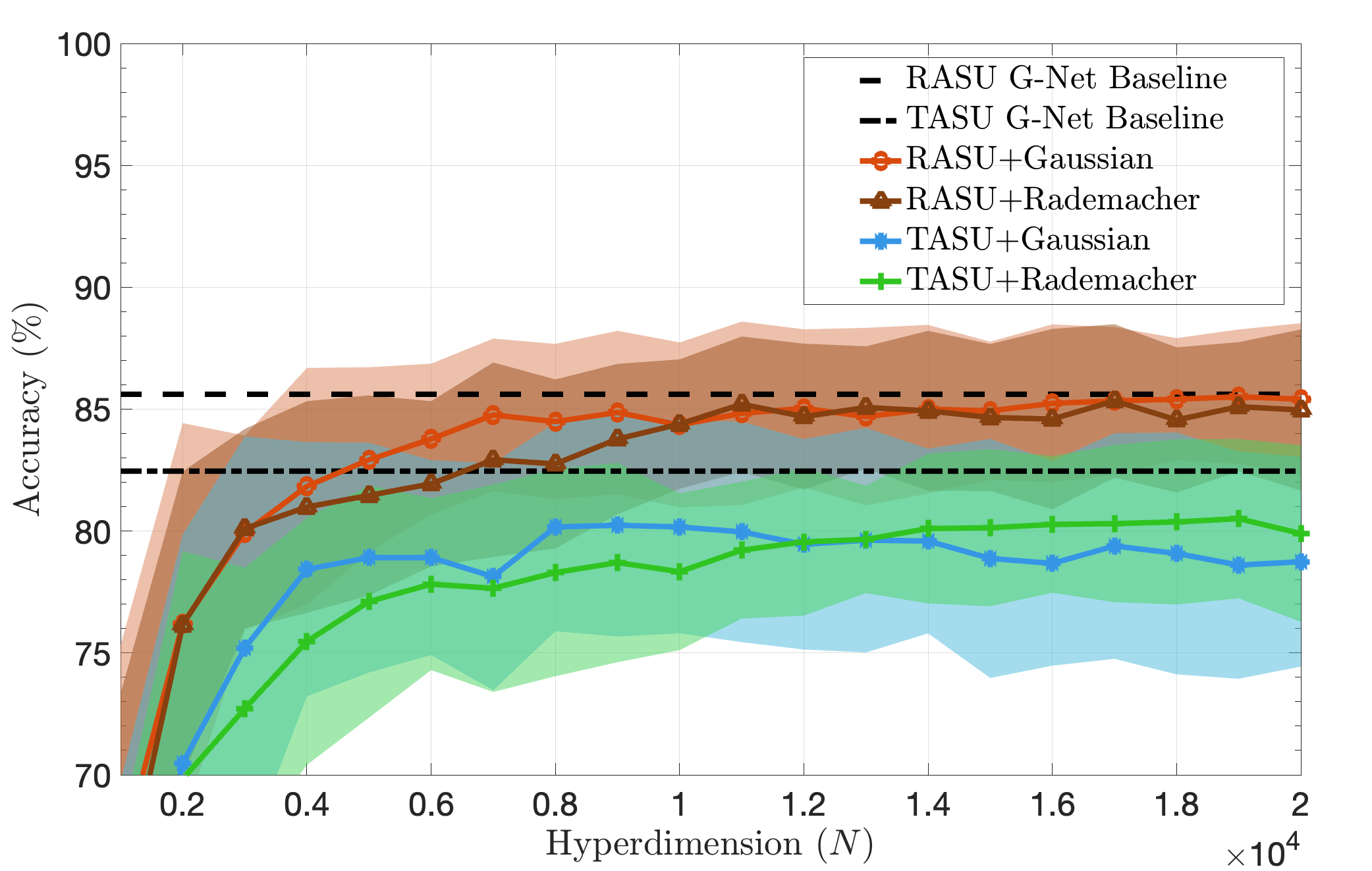}
\put (50,-3) {\scalebox{.7}{\rotatebox{0}{(d)}}}
\end{overpic}
\end{center}\vspace{-.2cm}
\caption{Performance of TASU versus RASU networks for Gaussian and Rademacher Embedding: (a) MNIST, (b) FashionMNIST, (c) Ford-A, (d) WSS }\label{figS2} \vspace{-.6cm}
\end{figure}

We extend our study of the first four datasets under the same experimental setup described above. While Figure~\ref{figS1} presents the performance of RASU G-Nets with both Rademacher and Gaussian embeddings, Figure~\ref{figS2} provides a more comprehensive comparison across a broader range of hyperdimensions for both activation types. Specifically, it presents G-Net and EHD G-Net accuracies of RASU and TASU networks, evaluated with Rademacher and Gaussian embeddings. As expected, RASU networks generally attain higher G-Net accuracies, with faster EHD G-Net concentration. However, TASU still remains competitive---achieving, for instance, 98.4\% accuracy on MNIST---although, as theory predicts, TASU-based EHD G-Nets converge more slowly to their G-Net baselines. This modest accuracy gap is the trade-off for a fully binary inference pipeline. Notably, the same constraint also gives TASU networks greater robustness to model perturbations, as will be discussed below. 

\begin{wrapfigure}[14]{r}{2.8in}
  \centering\vspace{-.3cm}
  \includegraphics[trim=0.17cm 0.25cm 0.25cm 0cm,clip,height=1.77in]{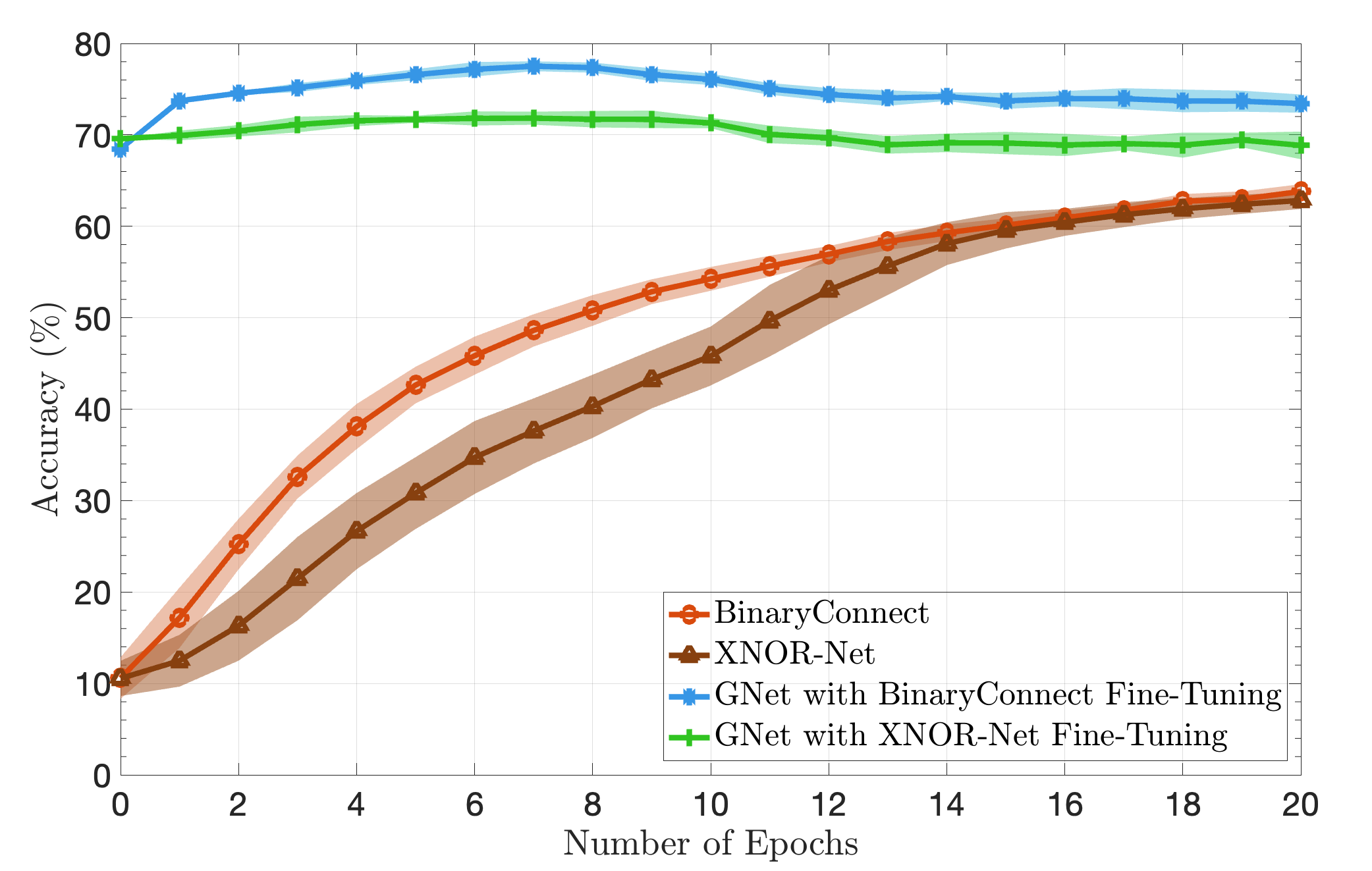}\vspace{-.7cm}
  \caption{Comparison with BNN Methods}
  \label{fig:BNNComp}
\end{wrapfigure}

As discussed earlier, the proposed framework develops a base G-Net model, from which combinatorially many BNNs can be instantiated. Although the frameworks and objectives differ, it is natural to ask how our approach relates to existing BNN methods. To this end, Figure~\ref{fig:BNNComp} compares G-Net with XNOR-Net~\cite{RastegariMohammad2016XICU} and BinaryConnect~\cite{courbariaux2015binaryconnect}. A direct, fair comparison with hybrid designs such as Bi-Real Net~\cite{LiuZechun2018BNEt} is not feasible, as those models interleave binary and floating-point operations.
To conduct a fair comparison, we first train a fully connected G-Net with two hidden layers (each of width 256) on FashionMNIST and then form a binary EHD G-Net using a hyperdimensional embedding of size 
$N=1000$. We subsequently fine-tune this model with either BNN technique and benchmark it against a similar BNN initialized with random weights. Figure~\ref{fig:BNNComp} shows that EHD-initialized BNNs start with substantially higher accuracy and attain their peak within a few epochs, whereas standard BNNs typically require many more epochs to reach comparable accuracy. From an optimization standpoint, G-Net allows performing a major part of the optimization in a smaller, continuous space, without the challenge of working with binary decision variables. G-Net accuracy serves as a baseline for the binary model, indicating how far its performance can be pushed in the course of training.

\begin{figure}[!htbp]
\begin{center}
\begin{overpic}[trim={0.17cm -.25cm  0.25cm 0},clip,height=1.77in]{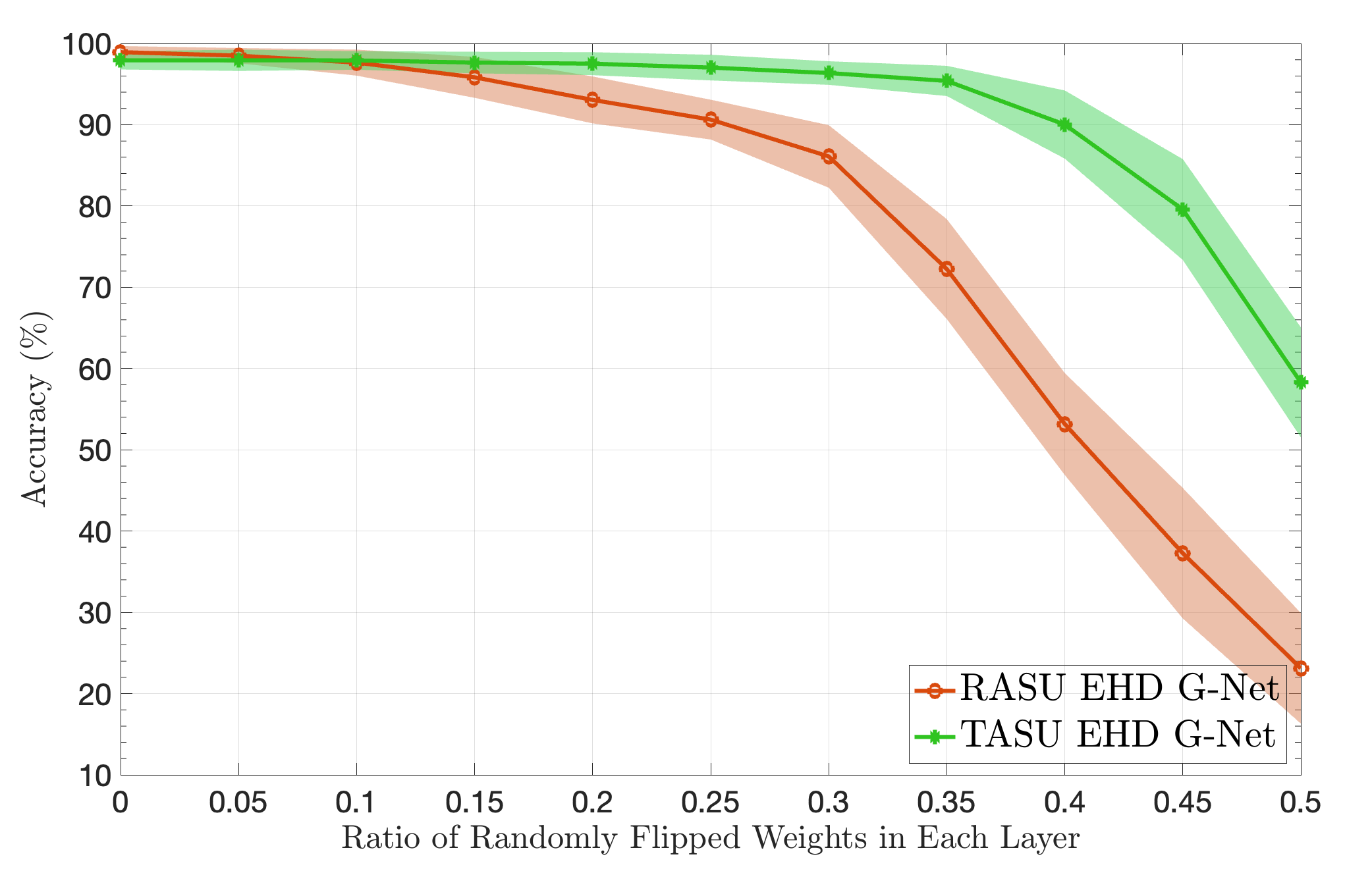}
\put (50,-3) {\scalebox{.7}{\rotatebox{0}{(a)}}}
\end{overpic}
\begin{overpic}[trim={0.17cm -.25cm  0.25cm 0},clip,height=1.77in]{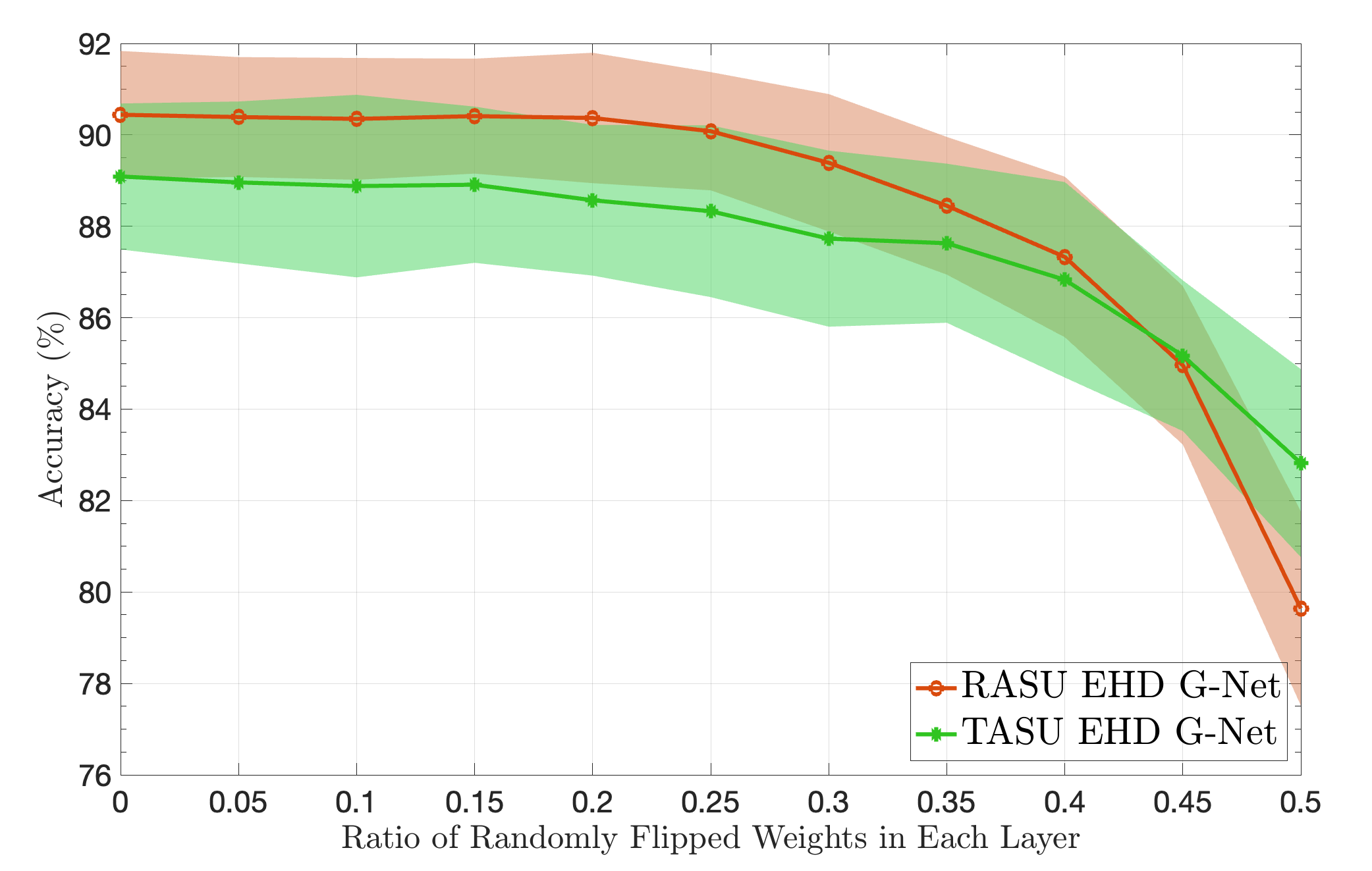}
\put (50,-3) {\scalebox{.7}{\rotatebox{0}{(b)}}}
\end{overpic}\vspace{.43cm}
\begin{overpic}[trim={0.17cm -.25cm  0.25cm 0},clip,height=1.77in]{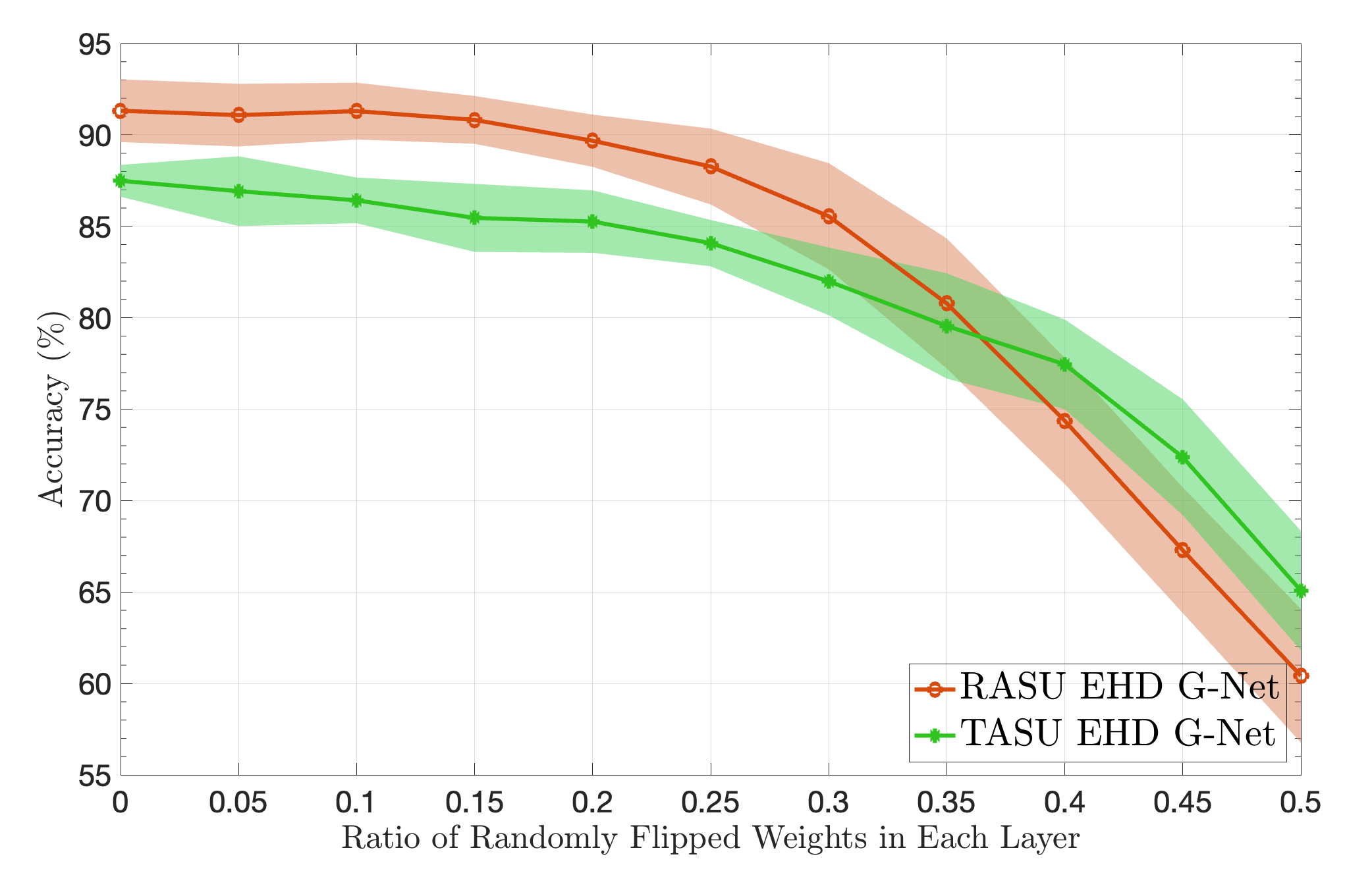}
\put (50,-3) {\scalebox{.7}{\rotatebox{0}{(c)}}}
\end{overpic}
\begin{overpic}[trim={0.17cm -.25cm  0.25cm 0},clip,height=1.77in]{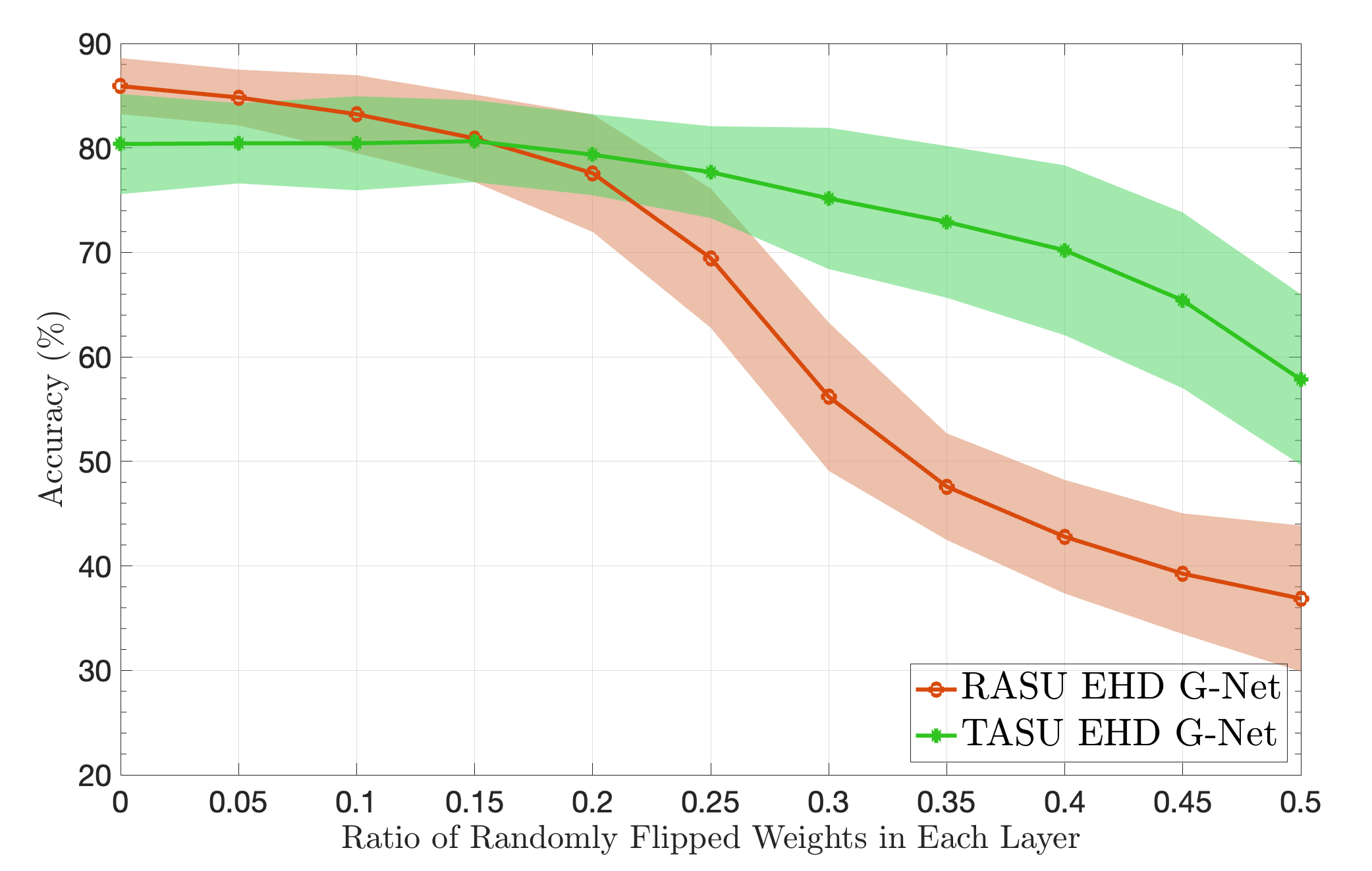}
\put (50,-3) {\scalebox{.7}{\rotatebox{0}{(d)}}}
\end{overpic}
\end{center}\vspace{-.2cm}
\caption{Robustness of TASU and RASU networks against random bit flips across all layers: (a) MNIST, (b) FashionMNIST, (c) Ford-A, (d) WSS}\label{figS3} \vspace{-.2cm}
\end{figure}

Hyperdimensional models are known for their resilience to model corruption, and our G-Net variants are no exception. To quantify this robustness, we stress-test RASU and TASU EHD G-Nets by randomly flipping a fraction of binary weights in every layer. For instance, applying a 10\% flip rate to the MNIST model randomly inverts 10\% of the weights in each of its four layers. Figure \ref{figS3} plots the resulting accuracy degradation versus corruption level, with confidence regions created by repeating the experiments twenty times. Notably, TASU networks lose accuracy more gradually than their RASU counterparts. This stems from TASU’s architecture: each TASU output is the sign of an inner product between two binary vectors, a quantity less sensitive to individual bit flips than the zero-thresholded binary inner products used in RASU layers. As a result, even after flipping 35\% of the EHD G-Net weights, the TASU model on MNIST still maintains over 
95\% accuracy.

\begin{figure}[!htbp]
\begin{center}
\begin{overpic}[trim={0.17cm -.25cm  0.25cm 0},clip,height=1.91in]{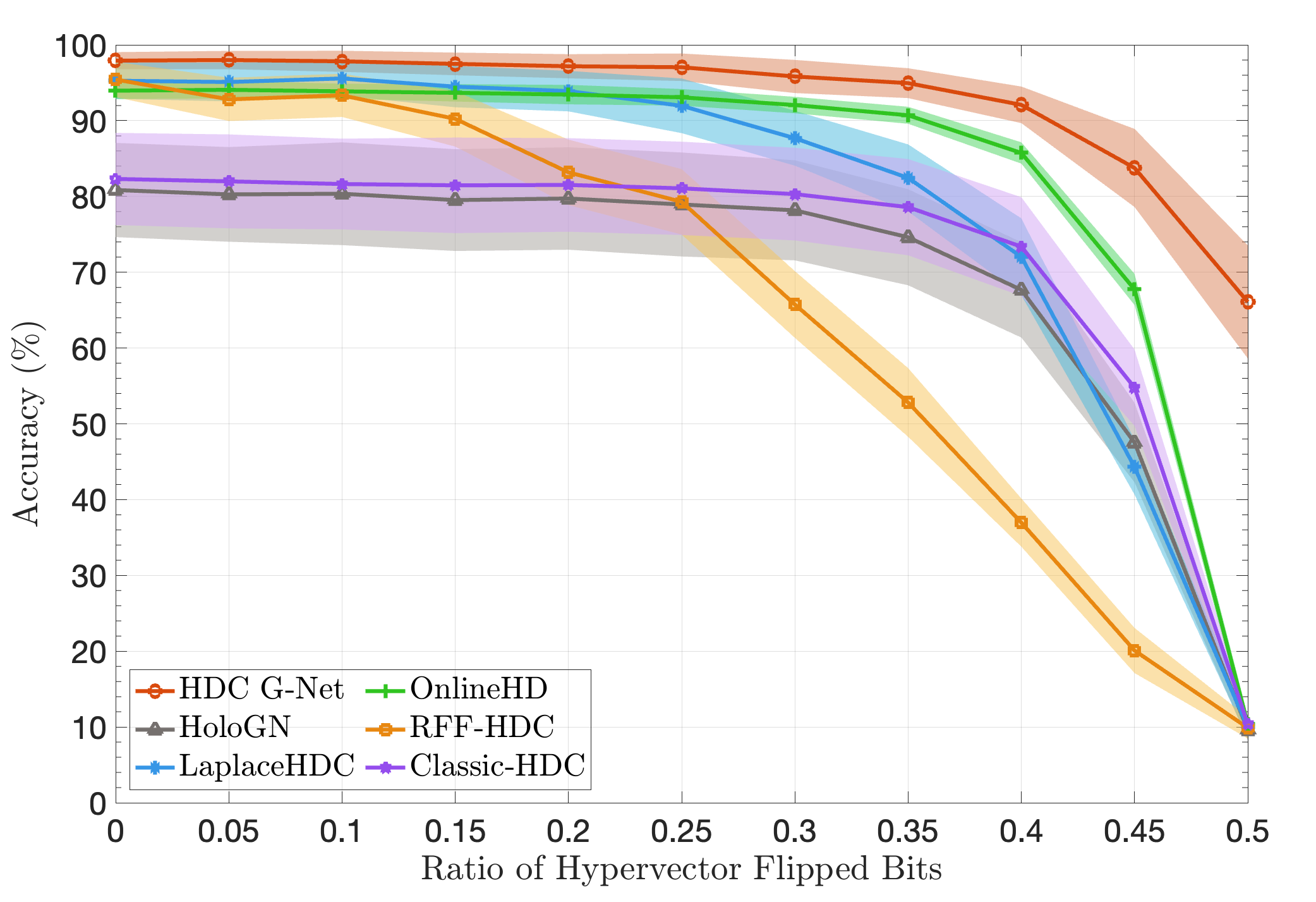}
\put (50,-3) {\scalebox{.7}{\rotatebox{0}{(a)}}}
\end{overpic}
\begin{overpic}[trim={0.17cm -.25cm  0.25cm 0},clip,height=1.91in]{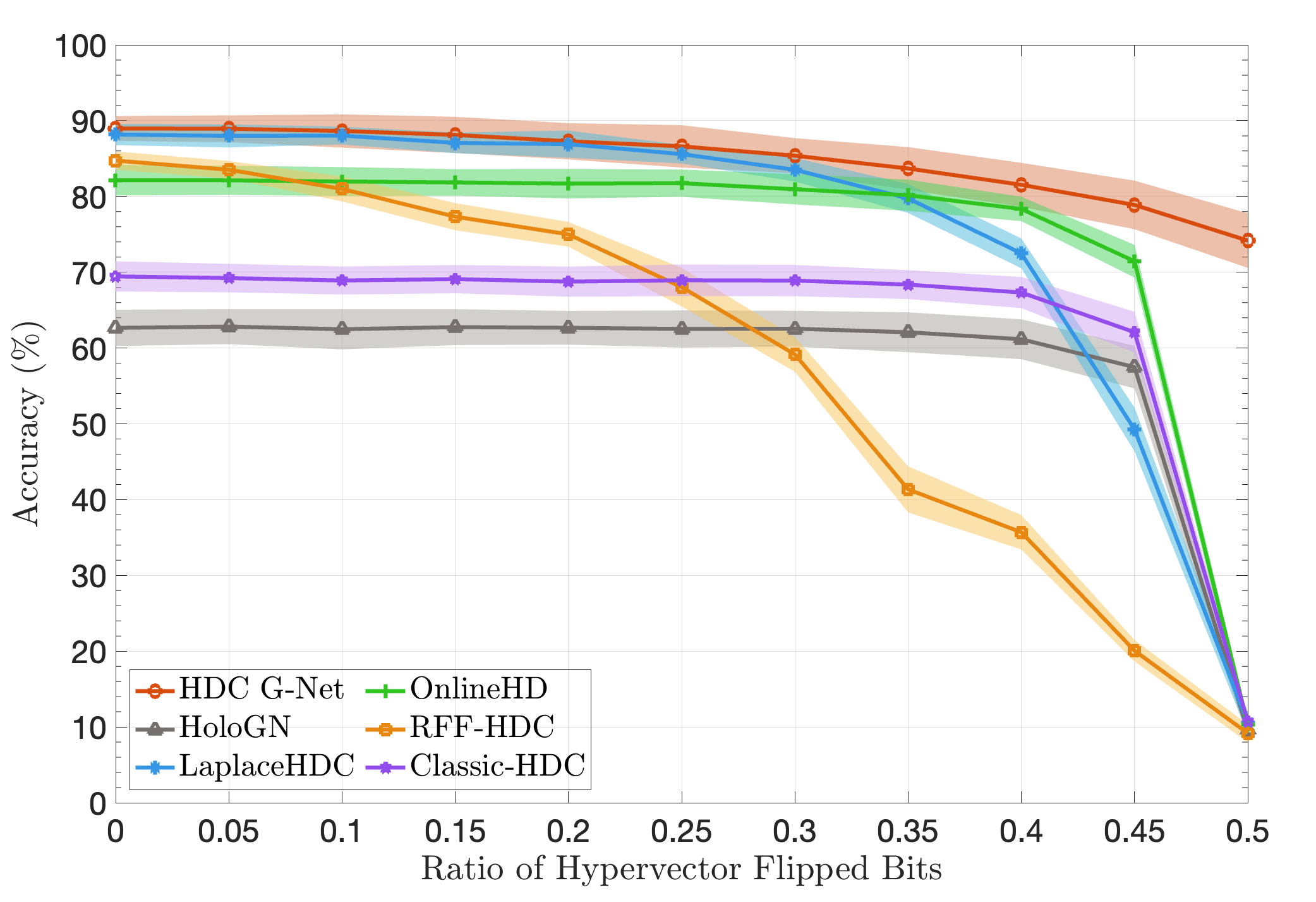}
\put (50,-3) {\scalebox{.7}{\rotatebox{0}{(b)}}}
\end{overpic}\vspace{.43cm}
\begin{overpic}[trim={0.17cm -.25cm  0.25cm 0},clip,height=1.91in]{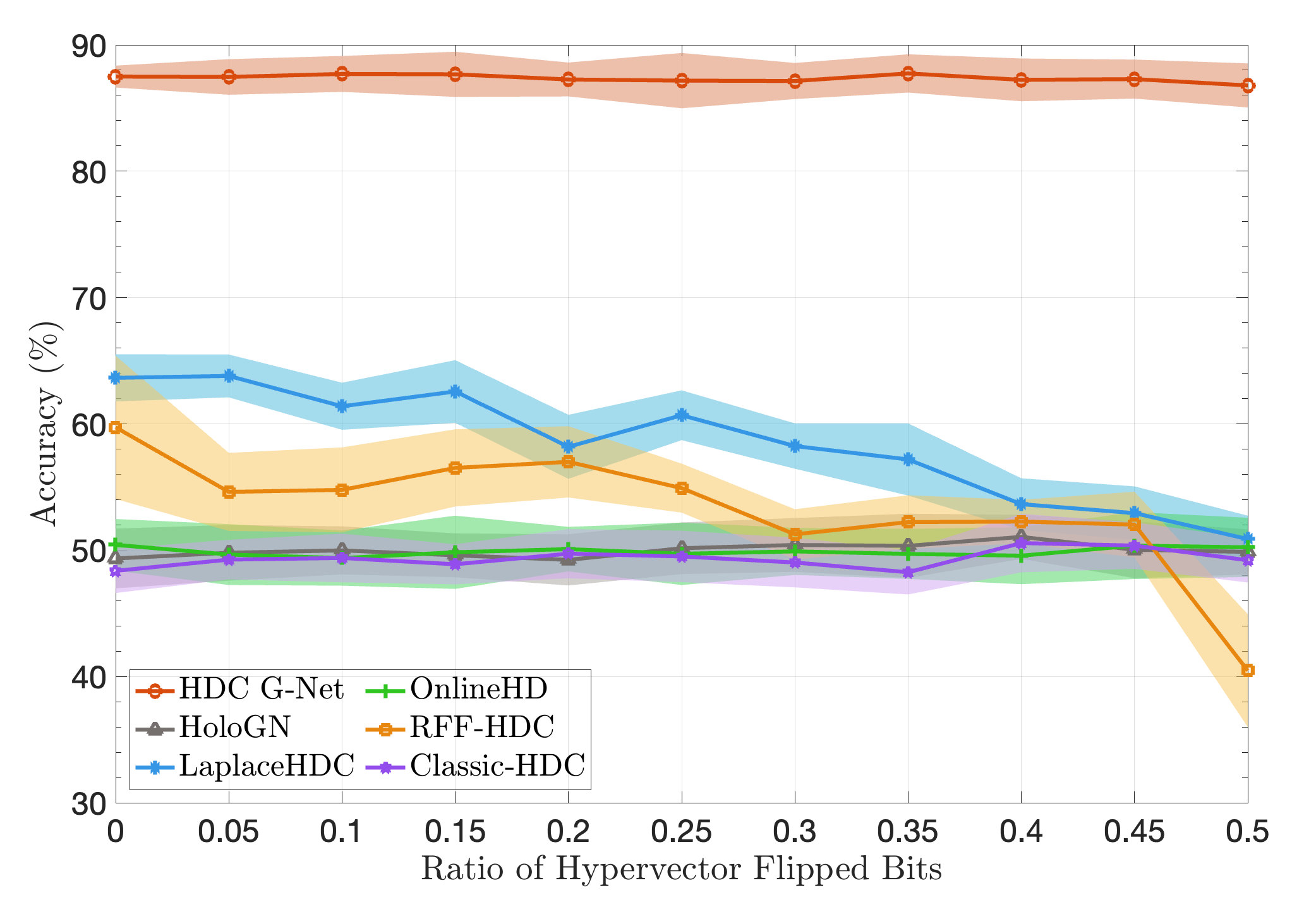}
\put (50,-3) {\scalebox{.7}{\rotatebox{0}{(c)}}}
\end{overpic}
\begin{overpic}[trim={0.17cm -.25cm  0.25cm 0},clip,height=1.91in]{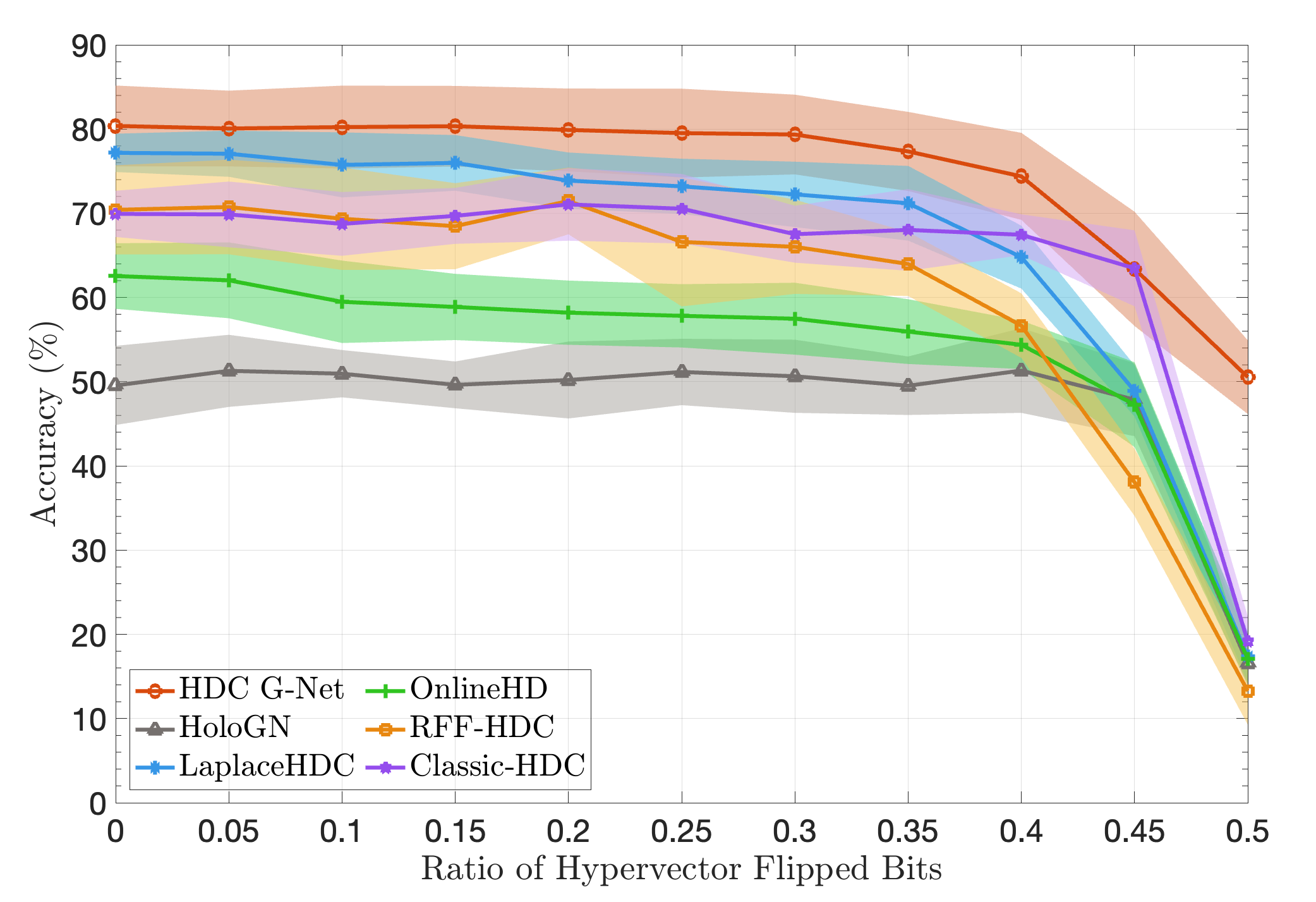}
\put (50,-3) {\scalebox{.7}{\rotatebox{0}{(d)}}}
\end{overpic}
\end{center}\vspace{-.2cm}
\caption{Comparing the robustness of an EHD G-Net  with other HDC methods: (a) MNIST, (b) FashionMNIST, (c) Ford-A, (d) WSS}\label{figS4} \vspace{-.6cm}
\end{figure}

Because each HDC method employs its own hyperdimensional embedding and inference pipeline, the weight-flipping corruption described above cannot be applied uniformly across other HDC frameworks. Instead, we observe a shared step across all models: each generates a hypervector that is subsequently fed into an inference block for prediction. To apply a common corruption pattern across all the HDC models in our comparison study, we fix the hyperdimension and corrupt the hypervectors by flipping a specified fraction of bits immediately after the embedding stage of each method. The noisy hypervector is then fed to the inference block for prediction, and the mean accuracy across multiple experiments is reported. For EHD G-Nets, bit flips are injected right after the first random-sign embedding, and the corrupted embedding is propagated through the network. Figure \ref{figS4} plots accuracy degradation for the various HDC models under this hypervector corruption. The TASU and RASU networks had a less noticeable performance gap under this test, and only TASU accuracies are reported in these plots. We attribute the superior robustness of EHD G-Nets to their tightly integrated, binary, multi-stage inference pipeline---a feature largely absent from existing HDC architectures.

\section{~~Computational Cost of EHD G-Net versus a Floating Point G-Net}\label{sec:compcost}

In this section, we mathematically compare the computational cost, in terms of time and memory, of EHD G-Net versus the corresponding floating-point G-Net. In a sense, any floating-point network can be considered as a binary neural network by viewing the floating-point weights as bit sequences, which are operated on by floating-point arithmetic operators. However, unlike the EHD G-Net, these bit sequences would have significant bits, and the operations on these bit sequences are complicated. In contrast, the EHD G-Net uses simple operations and has no significant bits; consequently, the EHD G-Net architecture is robust to random bit flips, see Figure \ref{figS3}. On the other hand,
floating-point numbers have significant bits (such as those encoding the sign and exponent of a floating-point number, which significantly change the output when changed). Thus, the EHD G-Net architecture is more suitable for noisy computing environments, such as low-power computing.

It is still instructive to study the different computational costs, in terms of memory and time, of each method. For simplicity, we restrict our attention to a single fully-connected EHD G-Net layer using a Rademacher matrix and the corresponding floating G-Net layer. 

\begin{figure}[!htbp]
\begin{center}
\begin{overpic}[trim={0.17cm -.25cm  0.25cm 0},clip,height=1.77in]{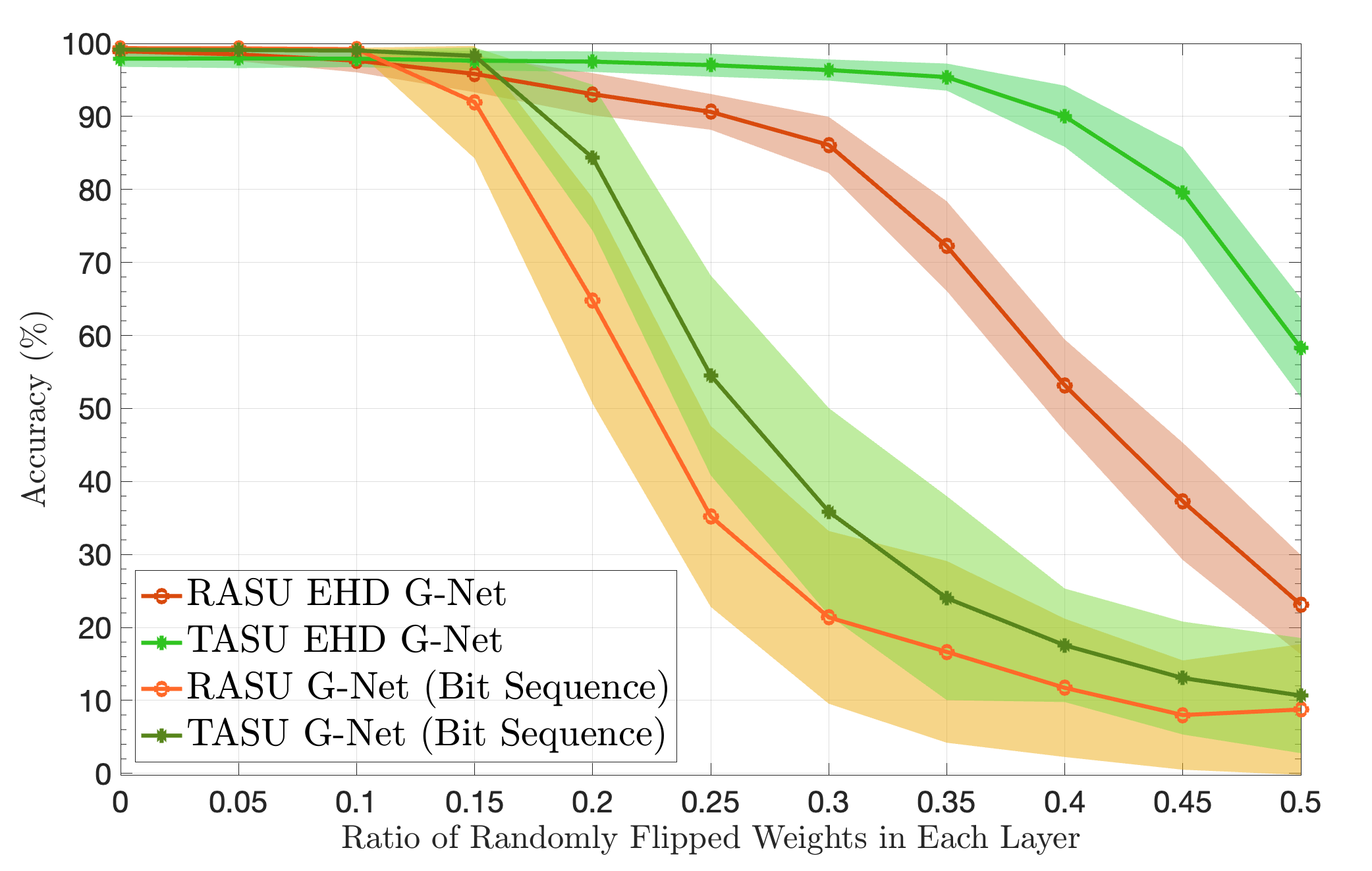}
\put (50,-3) {\scalebox{.7}{\rotatebox{0}{(a)}}}
\end{overpic}
\begin{overpic}[trim={0.17cm -.25cm  0.25cm 0},clip,height=1.77in]{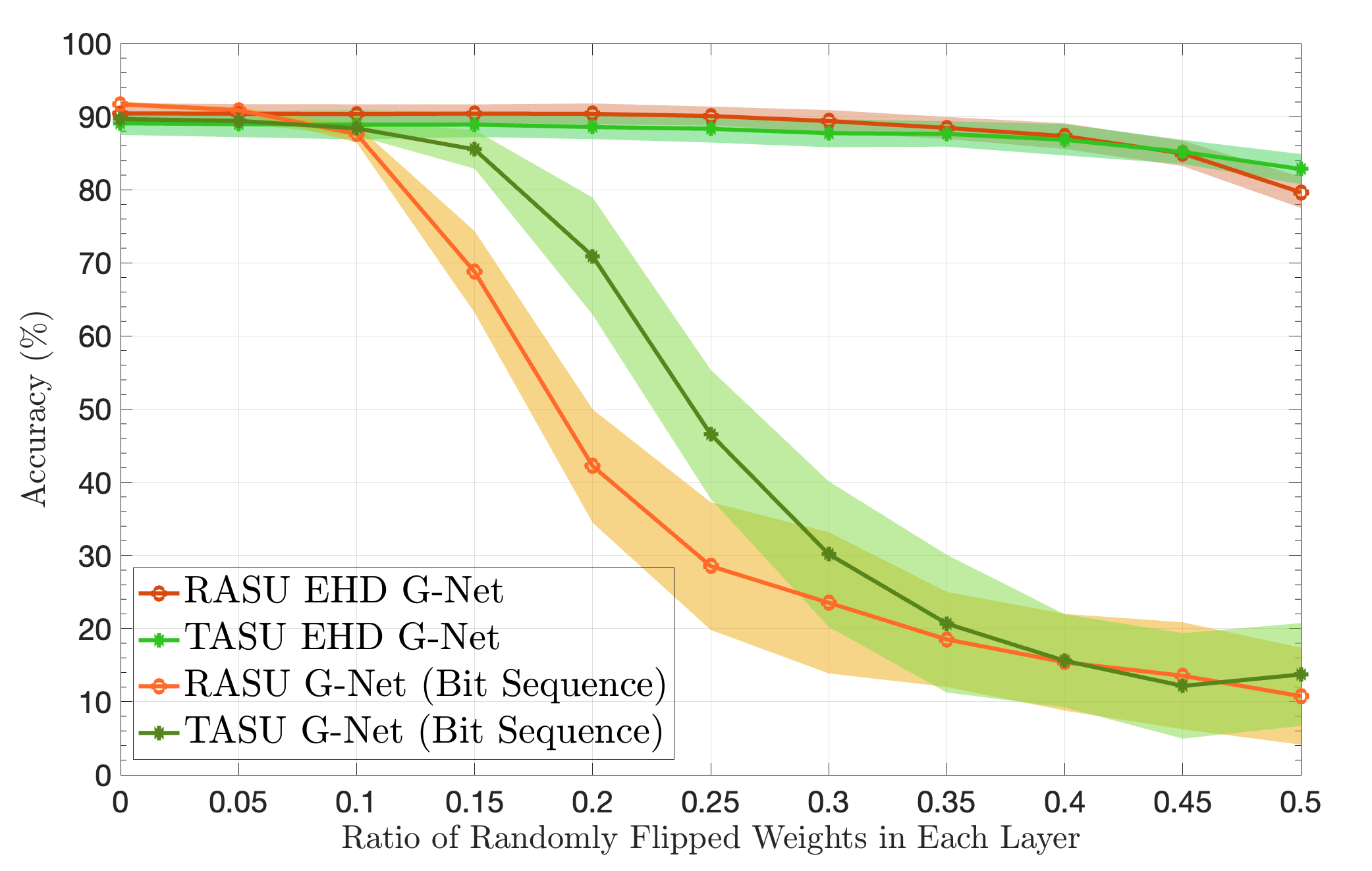}
\put (50,-3) {\scalebox{.7}{\rotatebox{0}{(b)}}}
\end{overpic}\vspace{-.2cm}
\end{center}
\caption{Robustness of TASU/RASU networks against random flips of EHD G-Net weights, and bit sequences of the corresponding G-Net: (a) MNIST, (b) FashionMNIST}\label{figS_R} \vspace{-.2cm}
\end{figure}

Fix a hyperdimension $N$, and the number of bits used in the floating-point representation of numbers $F$ (for example, $F = 64$ for double precision floating point numbers). Let $n$ and $m$ be the input and output dimensions, respectively.
Assume that $x \in \mathbb{R}^n$, $W \in \mathbb{R}^{m \times n}$, and $\mathrnd{R} \in \{-1,1\}^{N \times n}$ the hyper-dimension is $N$. 
Consider the EHD G-Net Layer that performs the map
$$
x \mapsto \texttt{sign}(W \mathrnd{R}^\top) \texttt{sign}(\mathrnd{R} x)
$$
and the corresponding floating-point G-Net layer 
$$
x \mapsto \frac{2}{\pi} \texttt{arcsin}(W x).
$$
 The EDH G-Net must store 
$\texttt{sign}(W \mathrnd{R}^\top)$ which has dimensions $m \times N$ and $\mathrnd{R}$, which has dimensions $N \times n$. The entries of these matrices can be encoded in binary, so the total memory required to store the EHD G-Net layer is
$$
\text{memory}_\text{EHD G-Net} = (m+n) N \quad \text{bits}.
$$
The floating point (FP) $G$-Net involves storing the $m \times n$ matrix $W$ whose entries each require $F$ bits to encode, for a total of 
$$
\text{memory}_\text{FP G-Net} = (m \cdot n) F \quad \text{bits}.
$$
Next, we consider the computational cost in terms of time.
The EHD G-Net layer starts by applying $\mathrnd{R}$ to $x$ and then applying the sign function. This involves an XOR of sign bits, adding the resulting integers, and taking the sign of the result. Let $t_\text{XOR}$ and $t_{\mathbb{Z},+}$ denote the time to compute the XOR and add integers, respectively. Finally, let $t_\text{sign}$ be the time to take the sign.
After that, we need to compute $t_\text{XOR}$ of $N m$ entries, and then sum $N m$ binary value, which can be done with popcount. Let $t_\text{popcount}$ denote the time for popcount (summing binary values). In summary, we have
$$
\text{time}_\text{EHD G-Net} = \mathcal{O} \Big( (t_\text{xor} + t_\text{popcount}) (N m) + (t_\text{xor} + t_{\mathbb{Z},+}) (N n) + t_\text{sign} (N) \Big).
$$
Next, consider the floating-point G-Net layer. Let $t_{\arcsin}$ be the cost to apply $2/\pi \arcsin$, $t_{\mathbb{R},+}$ be the time to add floating point numbers, and $t_{\mathbb{R},\times}$ be the time to multiply floating point numbers. 
Then, we have
$$
\text{time}_\text{FP G-Net} = \mathcal{O} \Big( 
(m) t_{\arcsin}  + 
(m \cdot n) (t_{\mathbb{R},+} + t_{\mathbb{R},\times}) \Big).
$$
If $N$ and $F$ are fixed, along with the time cost of all operations, then the computational complexity of EHD G-Net is $\mathcal{O}(m + n)$ in terms of memory and time complexity compared to $\mathcal{O}(m \cdot n)$ for floating point G-Net. However, even when $m$ and $n$ are small, we emphasize that EHD G-Net offers an advantage in terms of robustness to random bit flips. Figure \ref{figS_R} replicates the first row of Figure \ref{figS3} where we previously showed that randomly flipping the binary weights of an EHD G-Net causes a slow drop of accuracy. However, if instead one considers randomly flipping bit sequences on a floating point G-Net, the accuracy  drops much faster, as demonstrated in Figure \ref{figS_R}. 

\end{document}